\PassOptionsToPackage{usenames}{color}
\pdfoutput=1 
\documentclass[11pt,letterpaper]{article}
\usepackage{comment}
\usepackage{relsize} 

\usepackage[round]{natbib}
\usepackage{imakeidx} 
\makeindex 
\makeindex[name=construals,title={Index of Construals by Scene Role}]
\makeindex[name=revconstruals,title={Index of Construals by Function}]

\usepackage[boxed]{algorithm2e}

\usepackage[small,bf,skip=5pt]{caption}
\usepackage{sidecap} 
\usepackage{rotating}	

\makeatletter
\renewcommand{\paragraph}{%
  \@startsection{paragraph}{4}%
  {\z@}{3.25ex \@plus 1ex \@minus .2ex}{-1em}
  {\normalfont\normalsize\bfseries}%
}
\makeatother

\usepackage[nobottomtitles*]{titlesec}
\titleformat*{\subparagraph}{\itshape}
\titlespacing{\subparagraph}{%
  1em}{
  0pt}{
  1em}

\titleformat{\section}[block]
  {\Large\bfseries}
  {}
  {0pt}
  {\hspace{-1.2cm}
   \makebox[1cm][r]{\normalfont\thesection}\hspace{.2cm}}
\titleformat{name=\section,numberless}[block] 
  {\Large\bfseries}
  {}
  {0pt}
  {\hspace{-1.2cm}
   \makebox[1cm][r]{\normalfont}\hspace{.2cm}}
\titleformat{\subsection}[block]
  {\large\bfseries}
  {}
  {0pt}
  {\hspace{-1.2cm}
   \makebox[1cm][r]{\normalfont\thesubsection}\hspace{.2cm}}
\titleformat{\subsubsection}[block]
  {\normalsize\bfseries}
  {}
  {0pt}
  {\hspace{-1.2cm}
   \makebox[1cm][r]{\normalfont\thesubsubsection}\hspace{.2cm}}

\usepackage{enumitem} 
\setitemize{noitemsep,topsep=0em} 
\setenumerate{noitemsep,leftmargin=0em,itemindent=13pt,topsep=0em}

\usepackage{adjustbox}
\newcommand{\choices}[1]{\adjustbox{stack=ct}{#1}}  

\usepackage{textcomp}

\usepackage[procnames]{listings}

\usepackage{amssymb}	
\usepackage{amsmath}

\usepackage{pifont}
\newcommand{\chk}{\ding{51}}
\newcommand{\xxx}{\textcolor{mdred}{\ding{55}}}

\usepackage{fourier}
\usepackage[scaled=.87]{helvet}
\usepackage[scaled=.8]{beramono}
\usepackage[T1]{fontenc}
\usepackage[utf8x]{inputenc}

\usepackage{MnSymbol}	

\usepackage{latexsym}

\usepackage{array}
\usepackage{multirow}
\usepackage{booktabs} 
\usepackage{multicol}
\usepackage{footnote}

\usepackage{hyperref}
\usepackage{url}
\usepackage[usenames]{color}
\usepackage{xcolor}

\definecolor{darkblue}{rgb}{0, 0, 0.5}
\hypersetup{colorlinks=true,citecolor=darkblue, linkcolor=., urlcolor=darkblue}

\usepackage[normalem]{ulem} 
\usepackage{colortbl}
\usepackage{graphicx}
\usepackage{subcaption}
\usepackage{mdframed}

\usepackage{tikz}
\usepackage[edges]{forest}
\usetikzlibrary{arrows,positioning,calc}

\usepackage{color}
\usepackage{bm}
\definecolor{orange}{rgb}{1,0.5,0}
\definecolor{mdred}{rgb}{0.8,0,0}
\definecolor{mdgreen}{rgb}{0,0.6,0}
\definecolor{mdblue}{rgb}{0,0,0.7}
\definecolor{dkblue}{rgb}{0,0,0.5}
\definecolor{dkgray}{rgb}{0.3,0.3,0.3}
\definecolor{slate}{rgb}{0.25,0.25,0.4}
\definecolor{gray}{rgb}{0.5,0.5,0.5}
\definecolor{ltgray}{rgb}{0.7,0.7,0.7}
\definecolor{ltltgray}{rgb}{0.9,0.9,0.9}
\definecolor{purple}{rgb}{0.7,0,1.0}
\definecolor{lavender}{rgb}{0.65,0.55,1.0}

\makeatletter
\lst@AddToHook{EveryPar}{%
  \label{lst:\thelstnumber}
}
\makeatother
\usepackage{cleveref}

\crefformat{part}{\S#2#1#3}
\crefformat{chapter}{\S#2#1#3}
\crefformat{section}{\S#2#1#3}
\crefformat{subsection}{\S#2#1#3}
\crefformat{subsubsection}{\S#2#1#3}
\crefformat{paragraph}{\P#2#1#3}
\crefformat{subparagraph}{\P#2#1#3}
\crefmultiformat{section}{\S#2#1#3}{ and~\S#2#1#3}{, \S#2#1#3}{, and~\S#2#1#3}
\crefmultiformat{subsection}{\S#2#1#3}{ and~\S#2#1#3}{, \S#2#1#3}{, and~\S#2#1#3}
\crefmultiformat{subsubsection}{\S#2#1#3}{ and~\S#2#1#3}{, \S#2#1#3}{, and~\S#2#1#3}
\crefmultiformat{paragraph}{\P\P#2#1#3}{ and~#2#1#3}{, #2#1#3}{, and~#2#1#3}
\crefmultiformat{subparagraph}{\P\P#2#1#3}{ and~#2#1#3}{, #2#1#3}{, and~#2#1#3}
\crefrangeformat{section}{\mbox{\S\S#3#1#4--#5#2#6}}
\crefrangeformat{subsection}{\mbox{\S\S#3#1#4--#5#2#6}}
\crefrangeformat{subsubsection}{\mbox{\S\S#3#1#4--#5#2#6}}
\crefrangeformat{paragraph}{\mbox{\P\P#3#1#4--#5#2#6}}
\crefrangeformat{subparagraph}{\mbox{\P\P#3#1#4--#5#2#6}}
\crefname{part}{Part}{Parts}
\Crefname{part}{Part}{Parts}
\crefname{chapter}{ch.}{ch.}
\Crefname{chapter}{Ch.}{Ch.}
\crefname{figure}{figure}{figures}
\crefname{subfigure}{figure}{figures}
\Crefname{subfigure}{Figure}{Figures}
\crefname{appsec}{appendix}{appendices}
\Crefname{appsec}{Appendix}{Appendices}
\crefname{algocf}{algorithm}{algorithms}
\Crefname{algocf}{Algorithm}{Algorithms}
\crefname{enums}{example}{examples}
\Crefname{enums}{Example}{Examples}
\crefname{enumsi}{example}{examples}
\Crefname{enumsi}{Example}{Examples}
\crefname{}{example}{examples} 
\Crefname{}{Example}{Examples}
\crefformat{enums}{(#2#1#3)}
\crefformat{enumsi}{(#2#1#3)}
\crefformat{}{(#2#1#3)}
\crefname{xnumi}{example}{examples} 
\crefname{xnumi}{example}{examples} 
\Crefname{xnumii}{Example}{Examples} 
\Crefname{xnumii}{Example}{Examples} 
\crefformat{xnumi}{(#2#1#3)} 
\crefformat{xnumii}{(#2#1#3)} 
\crefrangeformat{enums}{\mbox{(#3#1#4--#5#2#6)}}
\crefrangeformat{enumsi}{\mbox{(#3#1#4--#5#2#6)}}
\crefrangeformat{xnumi}{\mbox{(#3#1#4--#5#2#6)}} 
\crefrangeformat{xnumii}{\mbox{(#3#1#4--#5#2#6)}} 
\crefmultiformat{enumsi}{(#2#1#3}{, #2#1#3)}{, #2#1#3}{, #2#1#3)}
\crefmultiformat{xnumi}{(#2#1#3}{, #2#1#3)}{, #2#1#3}{, #2#1#3)} 
\crefmultiformat{xnumii}{(#2#1#3}{, #2#1#3)}{, #2#1#3}{, #2#1#3)} 
\crefrangemultiformat{enumsi}{(#3#1#4--#5#2#6}{, #3#1#4--#5#2#6)}{, #3#1#4--#5#2#6}{, #3#1#4--#5#2#6)}
\crefrangemultiformat{xnumi}{(#3#1#4--#5#2#6}{, #3#1#4--#5#2#6)}{, #3#1#4--#5#2#6}{, #3#1#4--#5#2#6)} 
\crefrangemultiformat{xnumii}{(#3#1#4--#5#2#6}{, #3#1#4--#5#2#6)}{, #3#1#4--#5#2#6}{, #3#1#4--#5#2#6)} 

\ifx\creflastconjunction\undefined%
\newcommand{\creflastconjunction}{, and\nobreakspace} 
\else%
\renewcommand{\creflastconjunction}{, and\nobreakspace} 
\fi%

\newcommand*{\Fullref}[1]{\hyperref[{#1}]{\Cref*{#1}: \nameref*{#1}}}
\newcommand*{\fullref}[1]{\hyperref[{#1}]{\cref*{#1}: \nameref{#1}}}

\usepackage{suffix} 

\usepackage{gb4e} 
\newenvironment{xexe}{\begin{exe}\ex\begin{xlist}}{\end{xlist}\end{exe}} 

\newcommand{\backtick}[0]{\textasciigrave}

\newcommand{\w}[1]{\textit{#1}}	
\newcommand{\p}[1]{\textbf{\textsf{#1}}\index{#1@\textbf{\textsf{#1}}}} 
\WithSuffix\newcommand\p*[2]{\textbf{\textsf{#1}}\index{#2@\textbf{\textsf{#2}}}} 
\newcommand{\lbl}[1]{\textsc{#1}} 
\newcommand{\sst}[1]{\lbl{#1}\index{#1@\lbl{#1}}} 
\newcommand{\psst}[1]{\psstX{#1}{#1}} 
\newcommand{\psstX}[2]{\textcolor{mdgreen}{\hyperref[sec:#1]{\lbl{#2}}}\index{#1@\textcolor{mdgreen}{\textbf{\textsc{#1}}}}} 
\newcommand{\psstdef}[1]{\textcolor{mdgreen}{\hyperref[sec:#1]{\lbl{#1}}}\index{#1@\textcolor{mdgreen}{\textbf{\textsc{#1}}}|textbf}} 
\newcommand{\olbl}[1]{\textcolor{purple}{\textrm{#1}}} 

\newcommand{\rf}[2]{\psst{#1}$\leadsto$\psst{#2}\index[construals]{\protect\psst{#1}$\leadsto$\protect\psst{#2}}\index[revconstruals]{#2 #1@\protect\psst{#1}$\leadsto$\protect\psst{#2}}}
\newcommand{\Rf}[2]{\sst{#1}$\leadsto$\psst{#2}}

\newcommand{\backc}[0]{\hyperref[sec:coord]{\olbl{\backtick c}}\index{\backtick c@\olbl{\backtick c}}\xspace}
\newcommand{\backd}[0]{\hyperref[sec:discourse]{\olbl{\backtick d}}\index{\backtick d@\olbl{\backtick d}}\xspace}
\newcommand{\backi}[0]{\hyperref[sec:specialinf]{\olbl{\backtick i}}\index{\backtick i@\olbl{\backtick i}}\xspace}
\newcommand{\backposs}[0]{\hyperref[sec:possidiom]{\olbl{\backtick \$}}\index{\backtick \$@\olbl{\backtick \$}}\xspace}

\newcommand{\lex}[2][XXXX]{#2\index{#1@#2}} 
\WithSuffix\newcommand\lex*[2][XXXX]{\index{#1@#2}} 

\newcommand{\pex}[1]{\textit{#1}} 

\newcommand{\ghi}[1]{\href{https://github.com/carmls/snacs-guidelines/issues/#1}{\##1}} 

\newcommand{\finalversion}[1]{}
\newcommand{\futureversion}[1]{}

\newcommand{\longversion}[1]{} 
\newcommand{\draftnotice}[1]{} 

\newenvironment{ggroup}{{}}{{}}

\newcommand{\shortdef}[1]{\begin{mdframed}\noindent\textlarger{#1}\end{mdframed}}

\newenvironment{history}{\begin{mdframed}[linecolor=ltltgray,backgroundcolor=ltltgray]\small\noindent\textit{History.}}{\end{mdframed}}

\newcommand{\hierA}[1]{\textcolor{red}{\hyperref[sec:#1]{#1}}}
\newcommand{\hierB}[1]{\textcolor{blue}{\hyperref[sec:#1]{#1}}}
\newcommand{\hierC}[1]{\textcolor{mdgreen}{\hyperref[sec:#1]{#1}}}
\newcommand{\hierD}[1]{\textcolor{orange}{\hyperref[sec:#1]{#1}}}

\newcommand{\hierAdef}[1]{\section{\psstdef{#1}}\label{sec:#1}}
\newcommand{\hierBdef}[1]{\subsection{\psstdef{#1}}\label{sec:#1}}
\newcommand{\hierCdef}[1]{\subsubsection{\psstdef{#1}}\label{sec:#1}}
\newcommand{\hierDdef}[1]{\paragraph{\psstdef{#1}}\label{sec:#1}}

\title{Adposition and Case Supersenses v2.6:\\ Guidelines for English}

\newcommand{\emldisplay}[2]{\texttt{\href{mailto:#1}{#2}}}
\newcommand{\eml}[1]{\textsmaller[1.5]{\emldisplay{#1}{#1}}}

\newcommand{\affil}[1]{\textsuperscript{\,\textsf{\textsmaller{#1}}}}

\author{\textbf{Nathan Schneider}\affil{G} \quad \textbf{Jena D.~Hwang}\affil{A} \quad \textbf{Vivek Srikumar}\affil{U,A} \\
\textbf{Archna Bhatia}\affil{I} \quad \textbf{Na-Rae Han}\affil{P} \quad \textbf{Tim O'Gorman}\affil{C} \quad \textbf{Sarah R.~Moeller}\affil{C} \\
\textbf{Omri Abend}\affil{H} \quad \textbf{Adi Shalev}\affil{H} \quad \textbf{Austin Blodgett}\affil{G} \quad \textbf{Jakob Prange}\affil{G} \\
  \affil{G} Georgetown University \quad 
  \affil{A} Allen Institute for AI \quad
  \affil{U} University of Utah \\
  \affil{I} IHMC \quad
  \affil{P} University of Pittsburgh \quad
  \affil{C} University of Colorado Boulder \\
  \affil{H} Hebrew University of Jerusalem \\
\eml{nathan.schneider@georgetown.edu} \quad \eml{jenah@allenai.org} \quad \eml{svivek@cs.utah.edu}
}

\date{June 18, 2022}

\begin{document}
\maketitle
\begin{abstract}
\noindent 
This document offers a detailed linguistic description of SNACS \citep[Semantic Network of Adposition and Case Supersenses;][]{schneider-18}, 
an inventory of 52~semantic labels (``supersenses'')
that characterize the use of adpositions and case markers 
at a somewhat coarse level of granularity, 
as demonstrated in the STREUSLE corpus (\url{https://github.com/nert-nlp/streusle}; version~4.5 tracks guidelines version~2.6).
Though the SNACS inventory aspires to be universal, this document is specific to English; 
documentation for other languages will be published separately.

Version~2 is a revision of the supersense inventory proposed for English by 
\citet{schneider-15,schneider-16} 
(henceforth ``v1''), which in turn was based on previous schemes.
The present inventory was developed after extensive review of the 
v1 corpus annotations for English, 
plus previously unanalyzed genitive case possessives \citep{blodgett-18},
as well as consideration of adposition 
and case phenomena in Hebrew, Hindi, Korean, and German. 
\Citet{hwang-17} present the theoretical underpinnings of the v2 scheme.
\Citet{schneider-18} summarize the scheme, its application to English corpus data, 
and an automatic disambiguation task. 
\Citet{liu-21} offer an English Lexical Semantic Recognition tagger that includes SNACS labels in its output. 

This documentation can also be browsed alongside corpus data on the Xposition website \citep{xposition}: \url{http://www.xposition.org/}

\vfill
\end{abstract}

\tableofcontents

\section{Overview}


This document details version~2 of a scheme for annotating English 
prepositions and related grammatical markers with semantic class categories 
called \emph{supersenses}. 
The motivation and general principles for this scheme are laid out in 
publications cited in the abstract. 
This document focuses on the technical details, giving definitions, 
descriptions, and examples for each supersense and a variety of 
prepositions and constructions that occasion its use.

\subsection{What counts as an adposition?}

``Adposition'' is the cover term for prepositions and postpositions. 
Briefly, we consider an affix, word, or multiword expression to be adpositional if it:
\begin{itemize}
  \item mediates a semantically asymmetric figure--ground relation between two concepts, and
  \item is a grammatical item that can mark an NP. 
  We annotate \emph{tokens} of these items even where they mark clauses (as a subordinator) 
  or are intransitive.\footnote{Usually a coordinating conjunction, \p{but}  
  only receives a supersense when it is prepositional, as described 
  under \psst{PartPortion}.}
  We also include always-intransitive grammatical items whose core meaning is spatial and highly schematic, 
  like \p{together}, \p{apart}, and \p{away}.
\end{itemize}
%

Inspired by \citet{cgel}, the above criteria are broad enough to include 
a use of a word like \p{before} whether it takes an NP complement, 
takes a clausal complement (traditionally considered a subordinating conjunction), 
or is intransitive (traditionally considered an adverb):
\begin{xexe}
  \ex It rained \p{before} the party. [NP complement]
  \ex It rained \p{before} the party started. [clausal complement]
  \ex It rained \p{before}. [intransitive]
\end{xexe}

Even though they are not technically adpositions, 
we also apply adposition supersenses to possessive case marking 
(the clitic \p{'s} and possessive pronouns),
and some uses of the infinitive marker \p{to}, as detailed in \cref{sec:cxns}.

\subsection{Inventory}\label{sec:hier}

The v2.6 hierarchy is a tree with 52~supersense labels.
They are organized into three major subhierarchies: 
\psst{Circumstance} (18~labels), \psst{Participant} (15~labels),
and \psst{Configuration} (19~labels).

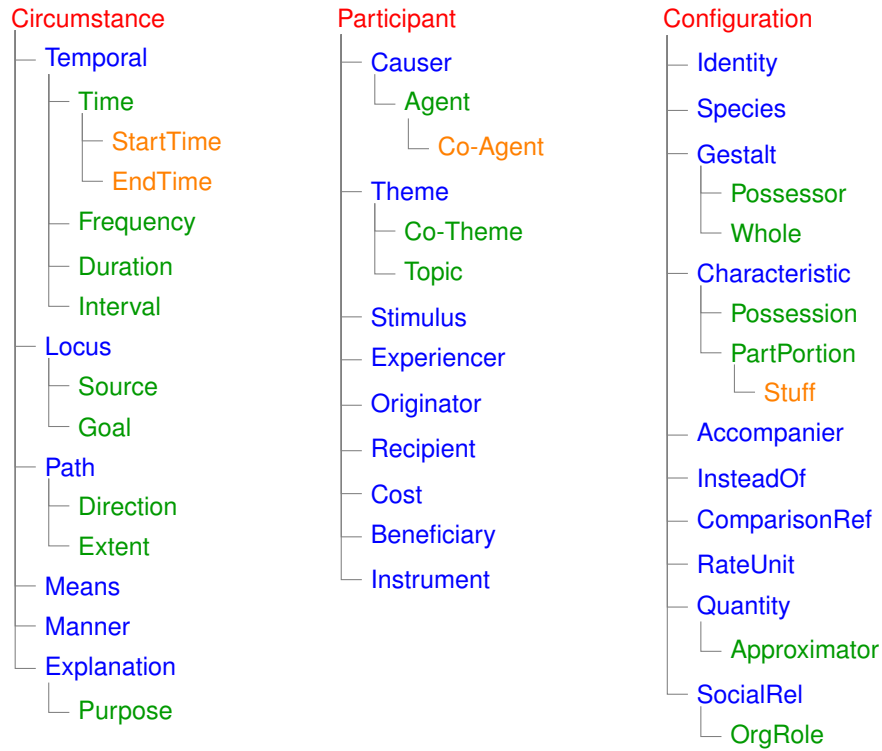
\begin{minipage}{\textwidth}
\vspace{.2cm}
\begin{multicols}{3}
\begin{ggroup}
  \sffamily\color{gray}
\begin{forest}
  for tree={%
    folder,
    grow'=0,
    fit=band,
    inner ysep=.75,
  }
  [{\hierA{Circumstance}}
    [{\hierB{Temporal}}
      [{\hierC{Time}}
        [{\hierD{StartTime}}]
        [{\hierD{EndTime}}]
      ]
      [{\hierC{Frequency}}]
      [{\hierC{Duration}}]
      [{\hierC{Interval}}]
    ]
    [{\hierB{Locus}}
      [{\hierC{Source}}]
      [{\hierC{Goal}}]
    ]
    [{\hierB{Path}}
      [{\hierC{Direction}}]
      [{\hierC{Extent}}]
    ]
    [{\hierB{Means}}]
    [{\hierB{Manner}}]
    [{\hierB{Explanation}}
      [{\hierC{Purpose}}]
    ]
  ]
\end{forest}
\columnbreak

\begin{forest}
  for tree={%
    folder,
    grow'=0,
    fit=band,
    inner ysep=.75,
  }
  [{\hierA{Participant}}
    [{\hierB{Causer}}]
    [{\hierB{Force}}
      [{\hierC{Agent}}
      ]
    ]
    [{\hierB{Theme}}
      [{\hierC{Topic}}]
      [{\hierC{Content}}]
    ]
    [{\hierB{Ancillary}}]
    [{\hierB{Stimulus}}]
    [{\hierB{Experiencer}}]
    [{\hierB{Originator}}]
    [{\hierB{Recipient}}]
    [{\hierB{Cost}}]
    [{\hierB{Beneficiary}}]
    [{\hierB{Instrument}}]
  ]
\end{forest}
\columnbreak

\begin{forest}
  for tree={%
    folder,
    grow'=0,
    fit=band,
    inner ysep=.75,
  }
  [{\hierA{Configuration}}
    [{\hierB{Identity}}]
    [{\hierB{Species}}]
    [{\hierB{Gestalt}}
      [{\hierC{Possessor}}]
      [{\hierC{Whole}}]
      [{\hierC{Org}}]
      [{\hierC{QuantityItem}}]
    ]
    [{\hierB{Characteristic}}
      [{\hierC{Possession}}]
      [{\hierC{PartPortion}}
        [{\hierD{Stuff}}]
      ]
      [{\hierC{OrgMember}}]
      [{\hierC{QuantityValue}}
        [{\hierD{Approximator}}]
      ]
    ]
    [{\hierB{Ensemble}}]
    [{\hierB{ComparisonRef}}]
    [{\hierB{SetIteration}}]
    [{\hierB{SocialRel}}]
  ]
\end{forest}
\end{ggroup}
\end{multicols}
\end{minipage}

\begin{itemize}
\item Items in the \psst{Circumstance} subhierarchy are prototypically 
expressed as adjuncts of time, place, manner, purpose, etc.\ 
elaborating an event or entity.
\item Items in the \psst{Participant} subhierarchy are prototypically 
entities functioning as arguments to an event.
\item Items in the \psst{Configuration} subhierarchy are prototypically
entities or properties in a static relationship to some entity.
\end{itemize}

\begin{history}
  v2.0--2.4 had 50~labels. In v2.5, the inventory was modified slightly:
  \sst{Co-Agent}, \sst{Co-Theme}, and \sst{InsteadOf}
  were removed (a mostly deterministic change as each was merged with another label),
  \sst{OrgRole} was split into \psst{Org} and \psst{OrgMember},
  \sst{Quantity} was split into \psst{QuantityItem} and \psst{QuantityValue},
  and \sst{Accompanier} was split into \psst{Ancillary} and \psst{Ensemble}.
\end{history}

\subsection{Limitations}

This inventory is only designed to capture semantic relations 
with a figure--ground asymmetry. This excludes:
\begin{itemize}
  \item The semantics of coordination, where the two sides of the relation 
are on equal footing (see \cref{sec:coord}).
  \item Aspects of meaning that pertain to information structure, discourse, 
or pragmatics (see \cref{sec:discourse}).
\end{itemize}
Moreover, this inventory only captures semantic distinctions 
that tend to correlate with major differences in syntactic distribution. 
Thus, while there are supersense labels for locative (\psst{Locus}), ablative (\psst{Source}),
allative (\psst{Goal}), and \psst{Path} semantics---and analogous temporal categories---%
finer-grained details of spatiotemporal meaning are for the most part lexical 
(viz.: the difference between \pex{\p{in} the box} and \pex{\p{on} the box}, 
or temporal \p{at}, \p{before}, \p{during}, and \p{after}) and are not represented here.\footnote{This is not to claim
that all members of a category can be grammatical in all the same contexts: 
\pex{\p{on} Saturday} and \pex{\p{at} 5:00} are both labeled \psst{Time}, 
though the prepositions are by no means interchangeable in American English. 
We are simply asserting that the different constructions specific to days of the week 
versus times of the day are minor aspects of the grammar of English.}

\subsection{Construal}\label{sec:construal}

In some cases, following \citet{hwang-17}, an adposition usage will be analyzed with
\emph{two} of the supersenses from the inventory.
This is done when the choice of adposition is analyzed as inviting a construal
that might not otherwise be the default for the semantic relation it marks
(perhaps due to semantic extension beyond the adposition's more prototypical meanings).

This is illustrated in \cref{SNACS-adpositions-table}.
The two semantic dimensions of an adposition usage are:

\textbf{Scene Role}: What is the basic semantic relation between the preposition-linked elements (e.g., governor and object)?
With a PP argument to a semantic predicate such as a content verb,
this will often correspond to a semantic role like \psst{Agent}, \psst{Theme}, or \psst{Recipient}.

\textbf{Function}: What semantic relation literally or metaphorically present in the scene
is higlighted by the choice of adposition?
Often this is a spatial meaning like \psst{Locus}, \psst{Source}, or \psst{Goal},
even if the situation is not literally spatial.

\textbf{In this document, construal is notated by \rf{Scene Role}{Function}
when the supersenses differ.}\footnote{This can be read as ``\psst{Scene Role} realized as \psst{Function}'' (or, ``an underlying \psst{Scene Role}
relation that is realized via an adposition coding for \psst{Function}'').}
Token annotations given as \psst{Supersense} are shorthand for a \textbf{congruent}
construal whose full form would be \rf{Supersense}{Supersense}.

\begin{table*}[t!]
\centering\small
\begin{tabular}{|l|l|l|l|c|}
\hline \textbf{Phrase} & \textbf{Scene Role} & \textbf{Coding} & \textbf{Function} & \textbf{Congruent?} \\ \hline
The ball was hit \p{by} the batter & \psst{Agent} & \p{by} & \psst{Agent} & \chk  \\[3pt]
Put the book \p{on} the shelf & \psst{Goal} & \p{on} & \psst{Locus} & \xxx  \\
Put the book \p{onto} the shelf & \psst{Goal} & \p{onto} & \psst{Goal} & \chk  \\[3pt]
I talked \p{to} her & \psst{Recipient} & \p{to} & \psst{Goal} & \xxx  \\[3pt]
I heard it \p{in} my bedroom & \psst{Locus} & \p{in} & \psst{Locus} & \chk  \\
I heard it \p{from} my bedroom & \psst{Locus}  & \p{from} & \psst{Source} & \xxx  \\[3pt]
John\p{'s} death & \psst{Theme} & \p{'s} & \psst{Gestalt} & \xxx \\
the windshield \p{of} the car & \psst{Whole} & \p{of} & \psst{Whole} & \chk \\
\hline
\end{tabular}
\caption{\label{SNACS-adpositions-table} Examples from \citet{shalev-19} illustrating construal analysis in terms of scene role, morphosyntactic coding, and function. The scene role and function annotations are labels from \cref{sec:hier} and are often but not always congruent for a particular token. The function annotation reflects the semantics of the morphosyntactic coding (i.e.~the choice of adposition).}
\end{table*}

Constraints on the supersenses that can serve as roles or as functions for English
adpositions are discussed in \cref{sec:constraints}.

\hierAdef{Circumstance}

\shortdef{Macrolabel for labels pertaining to space and time,
and other relations that are usually semantically non-core properties of events.}

\psst{Circumstance} is used directly for:
\begin{itemize}
  \item \textbf{Contextualization}
\begin{exe}
    \ex \p*{In}{in} arguing for tax reform, the president claimed that loopholes allow 
    big corporations to profit from moving their headquarters overseas.
    \ex\label{ex:in-activity-Circumstance} 
      You crossed the line \p{in} sharing confidential information.\\{} 
      [but see \cref{ex:in-activity-Topic} under \psst{Topic}, which is syntactically parallel]
    \ex I found out \p{in} our conversation that she speaks 5~languages.
    \ex\rf{Circumstance}{Locus}:\begin{xlist}
      \ex I haven't seen them \p{in} that setting.
      \ex \p*{In}{in} that case, I wouldn't worry about it.
    \end{xlist}
    \ex We have to keep going \p{through} all these challenges. [metaphoric motion] (\rf{Circumstance}{Path})
    \ex Bipartisan compromise is unlikely \p{with} the election just around the corner.
    \ex\label{ex:as-while} \p*{As}{as} we watched, she transformed into a cat. 
    [`while', `unfolding at the same time as'; not simply providing a `when'---contrast~\cref{ex:as-when} under \psst{Time}]
  \end{exe}
  For these cases, the preposition helps situate 
  the background context in which the main event takes place. 
  The background context is often realized as a subordinate clause 
  preceding the main clause. 
  It may also be realized as an adjective complement:
  \begin{exe}
    \ex\begin{xlist}
      \ex My tutor was helpful \p{in} giving concrete examples and exercises.
      \ex You were correct \p{in} \choices{answering the question\\your answer}.
    \end{xlist}
  \end{exe}
  Relatedly, we use \psst{Circumstance} to analyze \pex{involved \p{in}}:
  \begin{exe}
    \ex\begin{xlist}
      \ex I was involved \p{in} a car accident. (\psst{Circumstance})
      \ex Many steps are involved \p{in} the process of buying a home. \\(\rf{Whole}{Circumstance})
    \end{xlist}
  \end{exe}
  \item \textbf{Setting events}
  \begin{exe}
    \ex\label{ex:settingevt} We are having fun \choices{\p{at} the party\\\p{on} vacation}. (\rf{Circumstance}{Locus})
  \end{exe}
  The object of the preposition is a noun denoting a containing event; 
  it thus may help establish the place, time, and/or reason for the governing scene, 
  but is not specifically providing any one of these, despite the locative preposition.
  These can be questioned (at least in some contexts) with \emph{Where?} or \emph{When?}. 
  \Cref{ex:settingevt} entails \cref{ex:settingevtpred}:
  \begin{exe}
    \ex\label{ex:settingevtpred} We are \choices{\p{at} the party\\\p{on} vacation}. (\rf{Circumstance}{Locus})
  \end{exe}
  which may be responsive to the questions \emph{Where are you?} and \emph{What are you doing?}.\footnote{When 
  the object of the preposition is not a (dynamic) event, as with \pex{We are \p{at} odds/\p{on} medication}, 
  \rf{Characteristic}{Locus} usually applies: see discussion of state PPs at \psst{Characteristic}.}
  Journey-type PPs are treated similarly:
  \begin{exe}
    \ex\label{ex:on-way} They are \choices{\p{on} a journey\\\lex[on\_the\_way]{\p{on}\_~~the~~\_way}\\\lex*[on\_their\_way]{\p{on}\_~~their~~\_way}\p{on}\_~~their$_{\text{\backposs}}$~~\_way}  (\rf{Circumstance}{Locus})
  \end{exe}
  
  \item \textbf{Occasions}
  \begin{exe}
    \ex I bought her a bike \p{for} Christmas.
    \ex I had peanut butter \p{for} lunch.
  \end{exe}
  These simultaneously express a \psst{Time} and some element of causality 
  similar to \psst{Purpose}.
  But the PP is not exactly answering a \pex{Why?}\ or \pex{When?}\ question. 
  Instead, the sentence most naturally answers a question like \pex{On what occasion was X done?}
  or \pex{Under what circumstances did X happen?}.
  \item Any other descriptions of event/state properties that are \textbf{insufficiently specified} 
  to fall under spatial, temporal, causal, or other subtypes like \psst{Manner}. E.g.:
  \begin{exe}
    \ex\label{ex:over-lunch} Let's discuss the matter \p{over} lunch. [compare \cref{ex:at-lunch}]
  \end{exe}

\item \textbf{Conditions}
\begin{exe}
  \ex You can leave \choices{\p{as\_long\_as}\\provided} your work is done.
  \ex Whether you can leave \choices{depends \p{on}\\is subject \p{to}} whether your work is done.
\end{exe}
\end{itemize}

\hierBdef{Temporal}

\shortdef{Supercategory for temporal descriptions: 
\textbf{when}, \textbf{for how long}, \textbf{how often}, \textbf{how many times}, 
etc.\ something happened or will happen.}

Applies directly only to event descriptors with an aspectual quality that do not fit any of the subcategories:
\begin{exe}
  \ex The party is \p{over}. (= complete) (\psst{Temporal})
  \ex The plans are \choices{\p{in} progress\\\p{on} hold}. 
      (\rf{Temporal}{Locus})
  \ex\begin{xlist}
    \ex The party tomorrow is \p{on}. (= still scheduled to happen in the future) (\rf{Temporal}{Locus}) [see discussion at \psst{Characteristic}]
    \ex The party tomorrow is \p{off}. (= canceled) (\rf{Temporal}{Locus})
    \end{xlist}
\end{exe}

\begin{history}
  The v1 category \sst{Age} (e.g., \pex{a child \p{of} five}) 
  was a mutual subtype of \psst{Temporal} and \sst{Attribute}. 
  Being quite specific and rare, for v2 it was removed; see \cref{sec:age}. 
  Combined with the changes to \psst{Time} subcategories (see below), 
  this reduced by~3 the number of labels in the \psst{Temporal} subtree, 
  bringing it to 7.
\end{history}

\hierCdef{Time}

\shortdef{\textbf{When} something happened or will happen, in relation to an 
explicit or implicit reference time or event.}

\begin{exe}
  \ex We ate \choices{\p{in} the afternoon\\\p{during} the afternoon\\\p{at} 2:00\\\p{on} Friday}.
  \ex\label{ex:at-lunch} Let's talk \choices{\p{at}\\\p{during}} lunch. [compare \cref{ex:over-lunch}]
\end{exe}
For a containing time period or event, \p{during} can be used and is unambiguously \psst{Time}---%
unlike \p{in}, \p{at}, and \p{on}, which can also be locative.\footnote{See \cref{sec:temploc} regarding the 
use of locational metaphors for temporal relations.}
\begin{exe}
  \ex\begin{xlist}
    \ex They will greet us \choices{\p{on}\\\p{upon}} our arrival.
    \ex\label{ex:onXOccasion} I succeeded \p{on} \choices{the fourth attempt\\several occasions}. [contrast \emph{on occasion}, \cref{ex:onOccasion}]
  \end{xlist}
  \ex\label{ex:as-when}\p*{As}{as} meaning `when' (contrast \cref{ex:as-while} under \psst{Circumstance}):\begin{xlist}
    \ex The lights went out \p{as} I opened the door.
    \ex A bee stung me \p{as} I was eating lunch.
    \ex\label{ex:as-a-child} I played the piano \p{as} a child. (\rf{Time}{Identity}) [also~\cref{ex:as-a-child2}]
  \end{xlist}
  \ex I will finish \p{after} \choices{tomorrow\\lunch\\you (do)}.
  \ex I will finish \p{by} \choices{tomorrow\\lunch}.
  \ex I will contact you \choices{\p{as\_soon\_as}\\once} it's ready.
\end{exe}

The preposition \p{since} is ambiguous:
\begin{exe}
  \ex {} [`after'] I bought a new car---that was \p{since} the breakup. (\psst{Time})
  \ex {} [`ever since'] I have loved you \p{since} the party where we met. (\psst{StartTime}) 
  \ex {} [`because'] I'll try not to whistle \p{since} I know that gets on your nerves. (\psst{Explanation})
\end{exe}

Simple \psst{Time} is also used if the reference time is implicit and determined from 
the discourse:
\begin{exe}
  \ex We broke up last year, and I haven't seen her \p{since}. [since we broke up]
\end{exe}

However, \rf{Time}{Interval} is used for adpositions whose complement (object) 
is the amount of time between two reference points:
\begin{exe}
  \ex We left the party \p{after} an hour. [an hour after it started] (\rf{Time}{Interval})
  \ex We left the party an hour \p{ago}. [an hour before now] (\rf{Time}{Interval})
\end{exe}

The preposition \p{over} is also ambiguous:
\begin{exe}
  \ex The deal was negotiated \p{over} (the course of) a year. (\psst{Duration})
  \ex He arrived in town \p{over} the weekend. (\rf{Time}{Duration})
\end{exe}
See discussion under \psst{Duration}.

If the scene role is \psst{Time}, the PP can usually be questioned with \emph{When?}.

\psst{Time} is also used for special constructions for expressing clock times, e.g.~identifying 
a time via an offset:
\begin{exe}
  \ex\begin{xlist}
    \ex The alarm rang at \choices{a quarter \p{after}\\half \p{past}} 8. (\psst{Time})
    \ex\label{ex:quarterTo} The alarm rang at a quarter \p{to} 8. (\rf{Time}{Goal})
    \ex The alarm rang at a quarter \p{of} 8.\footnote{In some dialects, this is an alternate way to express the same meaning as \cref{ex:quarterTo}. 
    It seems that \p{to} and \p{of} construe the same time interval from opposite directions.} (\rf{Time}{Source})
  \end{xlist}
  \ex The alarm rang 15~minutes \p{before} 8. (\psst{Time}) [``15~minutes'' modifies the PP]
\end{exe}

\begin{history}
  In v1, point-like temporal prepositions (\p{at}, \p{on}, \p{in}, \p{as}) 
  were distinguished from displaced temporal prepositions (\p{before}, \p{after}, etc.)\ 
  which present the two times in the relation as unequal. 
  \sst{RelativeTime} inherited from \psst{Time} and was reserved for the 
  displaced temporal prepositions, as well as subclasses \psst{StartTime}, 
  \psst{EndTime}, \sst{DeicticTime}, and \sst{ClockTimeCxn}. 
  
  For v2, \sst{RelativeTime} was merged into \psst{Time}: the distinction 
  was found to be entirely lexical and lacked parallelism with the spatial hierarchy. 
  \sst{ClockTimeCxn} was also merged with \psst{Time}, the usages covered by the former 
  (expressions of clock time like \pex{ten \p{to} seven})
  being exceedingly rare and not very different semantically from 
  prepositions like \p{before}.
  \sst{DeicticTime} became \psst{Interval}.
\end{history}

\hierDdef{StartTime}

\shortdef{When the event denoted by the governor begins.}

Prototypical prepositions are \p{from} and \p{since} (but see note under \psst{Time} 
about the ambiguity of \p{since}):
\begin{exe}\ex\begin{xlist}
  \ex The show will run \p{from} 10 a.m. to 2 p.m.
  \ex a document dating \p{from} the thirteenth century
\end{xlist}\end{exe}

Note that simple \psst{Time} is used with verbs like \w{start} and \w{begin}: 
the event directly described by the PP is the starting, not the thing that started.
\begin{exe}
  \ex The show will start \p{at} 10 a.m. (\psst{Time})
\end{exe}

\hierDdef{EndTime}

\shortdef{When the event denoted by the governor finishes.}

Prototypical prepositions are \p{to}, \p{until}, \p{till}, \p{up\_to}, and \p{through}:
\begin{exe}
  \ex The show will run from 10 a.m. \p{to} 2 p.m.
  \ex Add the cider and boil \p{until} the liquid has reduced by half.
  \ex If we have survived \p{up\_to} now what is stopping us from surviving in the future?
  \ex They will be in London from March 24 \p{through} May 7.
\end{exe}

Note that simple \psst{Time} is used with verbs like \w{end} and \w{finish}: 
the event directly described by the PP is the ending, not the thing that ended.
\begin{exe}
  \ex The show will end \p{at} 2~p.m. (\psst{Time})
\end{exe}

\hierCdef{Frequency}

\shortdef{\textbf{At what rate} something happens or continues, 
or the instance of repetition that the event represents.}

\begin{exe}
  \ex Guests were arriving \p{at} a steady clip.
  \ex The risk becomes worse \p{by} the day.
  \ex\label{ex:onOccasion} I see them \choices{\lex[on\_occasion]{\p{on}\_occasion}\\\lex[from\_time\_to\_time]{\p{from}\_time\_to\_time}}. [contrast \emph{on \dots occasion}, \cref{ex:onXOccasion}]
  \ex\label{ex:dailyBasis} I see them \lex[on\_a\_daily\_basis]{\p{on}\_a\_~~daily~~\_basis}. (\rf{Frequency}{Manner}) [cf.~\cref{ex:bipartisanBasis}]
  \ex I keep getting the same message \p{over} and \p{over} again. 
\end{exe}
\psst{Frequency} is also used when an iteration is specified with an obligatory 
ordinal number modifier. 
If the ordinal number is optional, the preposition presumably receives another label:
\begin{exe}
  \ex\begin{xlist}
    \ex The camcorder failed \p{for} the third time. (\psst{Frequency})
    \ex\label{ex:freqrow} I skipped lunch \p{for} \choices{three days\\the third day} in$_{\text{\rf{Characteristic}{Locus}}}$ a row. (\psst{Frequency}) [see \cref{ex:inarow}]
    \ex We arrived \p{for} our (third) visit. (\psst{Purpose})
  \end{xlist}
\end{exe}

Contrast: \psst{SetIteration}

\hierCdef{Duration}

\shortdef{Indication of \textbf{how long} an event or state lasts
(with reference to an amount of time or 
time period\slash larger event that it spans).}

\begin{exe}
  \ex\label{ex:forDuration} I walked \choices{\p{for}\\\#\p{in}} 20~minutes.
  \ex\label{ex:GoalDuration} I walked to$_{\psst{Goal}}$ the store \choices{\p{in}/\p{within}\\\#\p{for}} 20~minutes. [see \cref{ex:inDuration}]
  \ex\label{ex:ExtentDuration} I walked a mile \choices{\p{in}/\p{within}\\\#\p{for}} 20~minutes.
  \ex\label{ex:AmbigDuration} I mowed the lawn \choices{\p{for}\\\p{in}/\p{within}} an hour.
\end{exe}
Note that the presence of a goal \cref{ex:GoalDuration} or 
extent of an event (\pex{a mile} in \cref{ex:ExtentDuration}) 
can affect the choice \psst{Duration} preposition, blocking \p{for}.
\Cref{ex:AmbigDuration} shows a direct object which can be interpreted 
either as something against which partial progress is made---licensing \p{for} 
and the inference that some of the lawn was not reached---or 
as defining the complete scope of progress, licensing \p{in}/\p{within} 
and the inference that the lawn was covered in its entirety.

The object of a \psst{Duration} preposition can also be a reference event 
or time period used as a yardstick for the extent of the main event:
\begin{exe}
  \ex\label{ex:EventDuration} I walked \p{for} the entire race. [the entire time of the race]
  \ex I walked \choices{\p{throughout}\\\p{through}\\well \p{into}} the night.
  \ex\label{ex:overDuration} The deal was negotiated \p{over} (the course of) a year.
\end{exe}
But \p{over} can also mark a time period that \emph{contains} the main event 
and is larger than it. While the path preposition \p{over} highlights that the 
object of the preposition extends over a period of time, it does not require that 
the main event extend over a period of time:
\begin{exe}
  \ex\label{ex:overTimeDuration} He arrived in town \p{over} the weekend. (\rf{Time}{Duration})
\end{exe}
Note that \p{during} can be substituted for \p{over} in \cref{ex:overTimeDuration} but not \cref{ex:overDuration}.

Some \p{for}-\psst{Duration}s measure the length of the specified event's \emph{result}:
\begin{exe}\ex \begin{xlist}
  \ex John went to the store \p{for} an hour. [he spent an hour at the store, not an hour going there]\footnote{This stands 
in contrast with \pex{John walked to the store \p{for} an hour}, where the most natural reading is that it took an hour to get to the store \citep[p.~230]{chang-98}.}
  \ex John left the party \p{for} an hour. [he spent an hour away from the party before returning]
\end{xlist}\end{exe}

A \psst{Duration} may be a stretch of time in which a simple event is repeated 
iteratively or habitually:
\begin{exe}\ex\begin{xlist}
  \ex I lifted weights \p{for} an hour. [many individual lifting acts collectively lasting an hour]
  \ex I walked to the store \p{for} a year. [over the course of a year, habitually went to the store by walking]
\end{xlist}\end{exe}

See further discussion at \psst{Interval}.

\hierCdef{Interval}

\shortdef{A marker that points retrospectively or prospectively in time, 
and if transitive, marks the time elapsed between two points in time.}

The clearest example is \p{ago}, which only serves to locate the \psst{Time}
of some past event in terms of its distance from the present:
\begin{exe}
  \ex\label{ex:ago} I arrived a year \p{ago}. (\rf{Time}{Interval}) \\{}
  [points backwards from the present: before now]
\end{exe}
The most common use of \psst{Interval} is in the construal \rf{Time}{Interval}: 
the time of an event is described via a temporal offset from some other time.

Another retrospective marker, \p{back}, can be transitive \cref{ex:backTrans}, 
or can be an intransitive modifier 
of a \psst{Time} PP \cref{ex:backIntrans}. 
Plain \psst{Interval} is used in the latter case:
\begin{exe}
  \ex\label{ex:backTrans} I arrived a year \p{back}.\footnote{While 
  \pex{a while \p{back}} and \pex{a few generations \p{back}} are generally accepted, 
  the use of \p{back} rather than \p{ago} for nearer and more precise temporal references,
  e.g.~\pex{10~minutes \p{back}}, appears to be especially associated with Indian English \citep[p.~7]{yadurajan-01}.} (\rf{Time}{Interval})
  \ex\label{ex:backIntrans} I arrived \p{back} in$_{\psst{Time}}$ June. (\psst{Interval})
\end{exe}

(This category is unusual in primarily marking a construal for a different scene role. 
But this seems justified given the restrictive set of English temporal prepositions 
that can appear with a temporal offset, and the distinct ambiguity of \p{in}.
\psst{Interval} is designed as the temporal counterpart of \psst{Direction}, 
which can construe static distance measures; 
in fact, \sst{TimeDirection} was considered as a possible name, 
but \psst{Interval} seemed more straightforward for the most frequent class of usages.)

Other adpositions can also take an amount of intervening time as their \emph{complement} (object):
\begin{exe}
  \ex\label{ex:inAmbiguousTime} I will eat \p{in} 10~minutes.
    \begin{xlist}
      \ex\label{ex:inDuration} {} [`for no more than 10~minutes' reading]: \psst{Duration}\footnote{This usage of \p{in} has been classified under the terms \emph{frame adverbial} \citep{pustejovsky-91} and \emph{span adverbial} \citep{chang-98}.}
      \ex\label{ex:inInterval} {} [`10~minutes from now' reading]: \rf{Time}{Interval}\footnote{This usage of \p{in}, as well as \p{ago} \cref{ex:ago} and \p{back} \cref{ex:backTrans,ex:backIntrans},
      are \emph{deictic}, i.e., they are inherently relative to the speech time or deictic center. 
      (See also \citet[pp.~154--157]{klein-94}.)
      This was taken to be a criterion for the v1 category \sst{DeicticTime}, 
      but that was never well-defined in v1 and was broadened for this version.}
    \end{xlist}
  \ex\label{ex:AfterObj} The game started at 7:00, but I arrived \choices{\p{after}\\\p{within}} 20~minutes. (\rf{Time}{Interval})
\end{exe}
Some adpositions license a temporal difference measure in \emph{modifier} position, which does not qualify:
\begin{exe}
  \ex To beat the crowds, I will arrive \uline{a while} \choices{\p{before} (it starts)\\\p{beforehand}}. (\psst{Time})
  \ex\label{ex:AfterMod} The game started at 7:00, but I arrived \uline{20~minutes} \choices{\p{after} (it started)\\\p{afterward}}. (\psst{Time})
\end{exe}
The preposition \p{after} can be used either way---contrast \cref{ex:AfterMod} with \cref{ex:AfterObj}.

Note that having \psst{Interval} as a separate category allows us to distinguish the sense of \p{in} 
in \cref{ex:inInterval} from both the \psst{Duration} sense \cref{ex:inDuration} 
and the \psst{Time} sense (\pex{\p{in} the morning}).

\paragraph{Versus \psst{Duration}.} 
The prepositions \p{in} and \p{within} are ambiguous between \psst{Interval} and \psst{Duration}.\footnote{By contrast, 
\p{after} seems to strongly favor \rf{Time}{Interval}. 
\pex{\p*{After}{after} a week, I had climbed all the way to the summit} is possible, 
but the conclusion that the climbing took a week may be an inference 
rather than something that is directly expressed.}
The distinction can be subtle and context-dependent.
The key test is whether the phrase answers a \emph{When?}\ question. 
If so, its scene role is \psst{Time}; otherwise, it is a \psst{Duration}.
\begin{exe}
  \ex \rf{Time}{Interval}:
  \begin{xlist}
    \ex I reached the summit \p{in} 3~days. [= 3~days later, I reached the summit.]
    \ex I was at the summit \p{within} 3~days. [= 3~days later, I was at the summit.]
    \ex I finished climbing \p{in} 3~days. [= 3~days later, I finished climbing.]
    \ex They had the engine fixed \p{in} 3~days. [= 3~days later, they had the engine fixed.]
  \end{xlist}
\end{exe}
\begin{exe}
  \ex \psst{Duration}:
    \begin{xlist}
      \ex I reached the summit \p{in} 3~days. [it took not more than 3~days]
      \ex I had climbed 1000 feet \p{in} [a total of] 3~days.
      \ex I fixed the engine \p{in} 3~days. [it took not more than 3~days] 
    \end{xlist}
\end{exe}

With a negated event, we use \psst{Duration}:
\begin{exe}
  \ex I haven't eaten \choices{\p{in}\\\p{for}} hours. [hours have passed since the last time I ate]
  (\choices{\#When} haven't you eaten?)
\end{exe}


\begin{history}
  Version~1 featured a label called \sst{DeicticTime}, under \sst{RelativeTime}, 
  which was meant to cover \p{ago} and temporal usages of other adpositions 
  (such as \p{in}) whose reference point is the utterance time or deictic center. 
  This concept proved difficult to apply and was (without good justification) 
  used as a catch-all for intransitive usages of temporal prepositions. 
  For v2, the new concept of \psst{Interval} is broader in that it drops the deictic 
  requirement (also covering \p{within}), while \psst{Time} has been clarified to include 
  intransitive usages of prepositions like \p{before} where the reference time 
  can be recovered from discourse context.
\end{history}

\hierBdef{Locus}

\shortdef{Location, condition, or value. May be abstract.}

\begin{exe}
  \ex I like to sing \choices{\p{at} the gym\\\p{on} Main St.\\\p{in} the shower}.
  \ex The cat is \choices{\p{on\_top\_of}\\\p{off}\\\p{beside}\\\p{near}} the dog.
  \ex There are flowers \choices{\p{between}\\\p{among}} the trees.
  \ex\label{ex:onRight} When you drive north, the river is \p{on} the right.
  \ex I read it \choices{\p{in} a book\\\p{on} a website}.
  \ex the data \p{in} the study
  \ex The charge is \p{on} my credit card.
  \ex We met \p{on} a trip to Paris.
  \ex The Dow is \p{at} \choices{a new high\\20,000}.\\{}
  [absolute scalar point value: see discussion at \psst{ComparisonRef}]
  \ex That's \p{in} my price range.
\end{exe}
The \psst{Locus} may be a part of another scene argument:
part of a figure whose static orientation is described, 
or a focal part of a ground where contact with the figure occurs:\footnote{\psst{PartPortion} 
was considered but rejected for these cases. Instead we assume the verb 
semantics would stipulate that it licenses a \psst{Theme} as well as a (core) \psst{Locus} 
which must be a part of that \psst{Theme}.}
\begin{exe}
  \ex She was lying \p{on} her back.
  \ex\begin{xlist}
    \ex She kissed me \p{on} the cheek.
    \ex I want to punch you \p{in} the face.
  \end{xlist}
\end{exe}
Words that incorporate a kind of reference point are \psst{Locus} 
even without an overt object:
\begin{exe}
  \ex\begin{xlist}
    \ex The cat is \p{inside} the house.
    \ex The cat is \p{inside}.
  \end{xlist}
  \ex\begin{xlist}
    \ex All passengers are \p{aboard} the ship.
    \ex All passengers are \p{aboard}.
  \end{xlist}
\end{exe}
\psst{Locus} also applies to \p{in}, \p{out}, \p{off}, \p{away}, \p{back}, 
etc.\ when used to describe a location without an overt object:
\begin{exe}
  \ex\begin{xlist}
    \ex The doctor is \choices{\p{in}\\\p{out\_of}\\\p{away\_from}} the office.
    \ex The doctor is \choices{\p{in}\\\p{out}\\\p{away}}.
    \ex They are \p{out} to eat.
  \end{xlist}
\end{exe}
And to \p{around} meaning `nearby' or `in the area':
\begin{exe}
  \ex Will you be \p{around} in the afternoon?
  \ex She's the best doctor \p{around}!
\end{exe}

In a phenomenon called \textbf{fictive motion} \citep{talmy-96}, 
dynamic language may be used to describe static scenes. 
We use construal for these:
\begin{exe}
  \ex A road runs \p{through} my property. (\rf{Locus}{Path})
  \ex John saw Mary \choices{\p{through} the window\\\p{over} the fence}.\footnote{The scene establishes a static spatial arrangement of John, Mary, and the window\slash fence, 
  with only metaphorical motion. Yet this is a non-prototypical \psst{Locus}: it cannot be questioned with \emph{Where?}, for example. 
  Moreover, we understand from the scene that the object of the preposition is something with respect to which the viewer is navigating in order to see without obstruction.} (\rf{Locus}{Path})
  \ex The road extends \p{to} the river. (\rf{Locus}{Goal})
  \ex I saw him \p{from} the roof. (\rf{Locus}{Source})
  \ex\label{ex:protesters} Protesters were \choices{kept\\missing} \p{from} the area. (\rf{Locus}{Source})
  \ex\begin{xlist} 
    \ex We live \p{across\_from} you. (\rf{Locus}{Source})
    \ex We're just \p{across} the street from$_{\text{\rf{Locus}{Source}}}$ you. (\rf{Locus}{Path}) 
  \end{xlist}
\end{exe}
Construal is also used for prepositions licensed by scalar adjectives of distance, 
\cref{ex:adjDist}, and prepositions used with a cardinal direction, \cref{ex:cardinal}:
\begin{exe}
  \ex\label{ex:adjDist} \begin{xlist}
    \ex We are quite close \p{to} the river. (\rf{Locus}{Goal})
    \ex We are quite far \p{from} the river. (\rf{Locus}{Source})
  \end{xlist}
  \ex\label{ex:cardinal} \begin{xlist}
    \ex The river is \p{to} the north. (\rf{Locus}{Goal}) [cf.~\cref{ex:onRight}]
    \ex The river is north \p{of} Paris. (\rf{Locus}{Source})
  \end{xlist}
\end{exe}
See also \rf{Locus}{Direction} for static distance measurements, 
described under \psst{Direction}.

Qualitative states of entities are analyzed as \rf{Characteristic}{Locus}, as described under \psst{Characteristic}.

\hierCdef{Source}

\shortdef{Initial location, condition, or value. May be abstract.}

For motion events, the initial location is where the thing in motion 
(the figure) starts out.
\psst{Source} also applies to abstract or metaphoric initial locations, 
including initial states in a dynamic event.

In English, a prototypical \psst{Source} preposition is \p{from}:
\begin{exe}
  \ex\label{ex:catBox} The cat jumped \choices{\p{from}\\\p{out\_of}} the box.
  \ex\label{ex:catLedge} The cat jumped \choices{\p{from}\\\p{off\_of}\\\p{off}} the ledge.
  \ex\label{ex:internet} I got it \choices{\p{from}\\\p{off}} the internet.
  \ex people \p{from} France
  \ex The temperature is rising \p{from} a low of 30 degrees.
  \ex I have arrived \p{from} work.
  \ex We discovered he was French \p{from} his attire. [indication]
  \ex I made it \p{out\_of} clay. [material]
  \ex\label{ex:coma} She \choices{awoke \p{from}\\came \p{out\_of}} a coma.
  \ex We are moving \p{off\_of} that strategy.
\end{exe}
The \psst{Source} use of \p{from} can combine with a specific locative PP:
\begin{exe}
  \ex I took the cat \p{from} behind$_{\psst{Locus}}$ the couch.
\end{exe}
Note that \p{away\_from} is ambiguous between marking a starting point (\psst{Source}) 
and a separate orientational reference point (\psst{Direction}):
\begin{exe}
  \ex At the sound of the gun, the sprinters ran \choices{\p{away\_from}\\\p{from}} the starting line. (\psst{Source})
  \ex The bikers ride parallel to the river for several miles, then 
  head east, \choices{\p{away\_from}\\\#\p{from}} the river. 
  (\psst{Direction}: bikers are never at the river)
\end{exe}
%
%
%
Note, too, that \p{off(\_of)} and \p{out(\_of)} can also mark simple states:
\begin{exe}
  \ex I am \p{off} \choices{medications\\work}. (\rf{Characteristic}{Locus})
  \ex The lights are \choices{\p{off}\\\p{out}}. (\rf{Characteristic}{Locus})
  \ex Stay \p{out\_of} trouble. (\rf{Characteristic}{Locus})
\end{exe}
States are discussed at length under \psst{Characteristic}. 
There is also a (negated) possession sense of \p{out}/\p{out\_of}:
\begin{exe}
  \ex We are \p{out\_of} toilet paper. (\psst{Possession})
\end{exe}

Sometimes a specific \psst{Source} is implicit, and the preposition is intransitive. 
But if no specific referent is implied, another label may be more appropriate:
\begin{exe}
  \ex The cat was sitting on the ledge, then jumped \p{off}. (\psst{Source}: implicit `(of) it')
  \ex He was offered the deal, but walked \p{away}. (\psst{Source}: implicit `from it')
  \ex The bird flew \choices{\p{away}\\\p{off}}. (\psst{Direction}: vaguely away from the viewpoint)
\end{exe}

\psst{Source} is prototypically inanimate, 
though it can be used to construe animate \psst{Participant}s 
(especially \psst{Originator} and \psst{Force}).
Contrasts with \psst{Goal}.

\paragraph{Agency as giving.}
When an \psst{Agent}'s action to help somebody is conceptualized as 
giving, and the nominalized action as the thing given, 
then \p{from} can mark the \psst{Agent} (metaphorical giver).
If the \p{from}-PP is adnominal, \rf{Agent}{Source} is used \cref{ex:AgentSource}.
However, if the \p{from}-PP is adverbial, and the verb relates to the metaphoric 
transfer rather than the event described by the action nominal, 
then the argument linking becomes too complicated for this scheme to express; 
simple \psst{Source} is used by default \cref{ex:AgentiveSource}:
\begin{exe}
  \ex\label{ex:AgentSource} The attention \p{from} the staff made us feel welcome. (\rf{Agent}{Source})
  \ex\label{ex:AgentiveSource}\psst{Source}:\begin{xlist} 
    \ex I received great care \p{from} this doctor.
    \ex I got a second chance \p{from} her.
    \ex I need a favor \p{from} you.
  \end{xlist}
\end{exe}

\hierCdef{Goal}

\shortdef{Final location (destination), condition, or value. May be abstract.}

Prototypical prepositions include \p{to}, \p{into}, and \p{onto}:
\begin{exe}
  \ex I ran \p{to} the store.
  \ex The cat jumped \p{onto} the ledge.
  \ex I touched my ear \p{to} the floor.
  \ex She sank \p{to} her knees.
  \ex Add vanilla extract \p{to} the mix.
  \ex Everyone contributed \p{to} the meeting.
  \ex The temperature is rising \p{to} a high of 40 degrees.
  \ex We have access \p{to} the library's extensive collections.
  \ex She slipped \p{into} a coma.
  \ex The drugs put her \p{in} a coma. (\rf{Goal}{Locus})
  \ex\label{ex:result} \textbf{Result} \citep[p.~1224]{cgel}: \begin{xlist}
    \ex We arrived at the airport only \p{to} discover that our flight had been canceled.
    \ex May you live \p{to} be 100!
  \end{xlist}
\end{exe}
For motion events, a \psst{Goal} must have been reached if the event 
has progressed to completion (was not interrupted).
\psst{Direction} is used instead for \p{toward(s)} and \p{for}, 
which mark an intended destination that is not necessarily reached:
\begin{exe}
  \ex\begin{xlist}
    \ex I headed \p{to} work. (\psst{Goal})
    \ex I headed \choices{\p{towards}\\\p{for}\\\#\p{to}} work but never made it there. (\psst{Direction})
  \end{xlist}
\end{exe}


\paragraph{\emph{go to\lex*[go to]{go \p{to}}}.} A conventional way to express one's status as a student at some school is
with the expression \pex{go \p{to} (name or kind of school)}.
Construal is used when \pex{go \p{to}} indicates student status, rather than 
(or in addition to) physical attendance:
\begin{exe}
  \ex\label{ex:student} I went \p{to} (school at$_{\psst{Locus}}$) UC Berkeley. (\rf{Org}{Goal})
  \exp{ex:student} I went \p{to} UC Berkeley for the football game. (\psst{Goal})
\end{exe}
Going to a business as a customer, going to an attorney as a client, 
going to a doctor as a patient, etc.\ can also convey long-term status, 
but there is considerable gray area between habitual going and 
being in a professional relationship, so we simply use \psst{Goal}:
\begin{exe}
  \ex I go \p{to} Dr.~Smith for my allergies. (\psst{Goal})
\end{exe}

\paragraph{Locative as destination.}
English regularly allows canonically static locative prepositions to mark 
goals with motion verbs like \pex{put}.
We use the \rf{Goal}{Locus} construal to capture both the static and dynamic aspects of meaning:
\begin{exe}
  \ex\rf{Goal}{Locus}: \begin{xlist}
    \ex I put the lamp \p{next\_to} the chair.
    \ex I'll just hop \p{in} the shower.
    \ex I put my CV \p{on} the internet.
    \ex The cat jumped \p{on} my face.
    \ex The box fell \p{on} its side.
    \ex We arrived \p{at} the airport.
  \end{xlist}
\end{exe}

\paragraph{Application of a substance.}
\begin{xexe}
  \ex the paint that was applied \p{to} the wall (\psst{Goal})
  \ex the paint that was sprayed \p{onto} the wall (\psst{Goal})
  \ex the paint that was sprayed \p{on} the wall (\rf{Goal}{Locus}) 
\end{xexe}
The wall is the endpoint of the paint, hence \psst{Goal} is the scene role. 
(Though the wall can be said to be affected by the action, we prioritize 
the motion aspect of the scene in choosing \psst{Goal} rather than \psst{Theme}.)


\psst{Goal} is prototypically inanimate, though it can be used to construe animate \psst{Participant}s 
(especially \psst{Recipient}).
Contrasts with \psst{Source}.

\hierBdef{Path}

\shortdef{The ground that must be covered in order for the motion to be complete.}

The ground covered is often a linear extent with or without 
specific starting and ending points:
\begin{exe}
  \ex The bird flew \p{over} the building.
  \ex The sun traveled \p{across} the sky.
  \ex Hot water is running \p{through} the pipes.
  \ex\label{ex:pathmanner} They dance \p{in} a circle. (\rf{Path}{Locus})
\end{exe}

It can also be a waypoint\slash something that must be passed or encircled. 
\begin{exe}
  \ex We flew to Rome \p{via} Paris.
  \ex I go \p{by} that coffee shop every morning.
  \ex The earth has completed another orbit \p{around} the sun.
\end{exe}
If this is a portal in the boundary of a container, 
it is often construed as \psst{Source}, \psst{Goal}, or \psst{Locus}:
\begin{exe}
  \ex The bird flew \p{in} the window. (\rf{Path}{Locus})
  \ex The bird flew \p{out} the window. (\rf{Path}{Source})
  \ex A cool breeze blew \p{into} the window. (\rf{Path}{Goal})
\end{exe}
However, if the entirety of the motion event is simply located, \psst{Locus} applies:
\begin{exe}
  \ex The bird was flying \p{in} the house. [The flying took place in the house.] (\psst{Locus})
  \ex They were running \p{on} the street. [The running took place on the street.] (\psst{Locus})
\end{exe}

The prepositions \p{around} and \p{throughout} can mark a region in which motion 
that follows an aimless or complex trajectory is contained. 
Construal is used for these, whether or not the region is explicit:
\begin{exe}\ex \rf{Locus}{Path}:\begin{xlist}
  \ex The kids ran \p{around}.
  \ex The kids ran \choices{\p{around}\\\p{throughout}} the kitchen.
  \ex The kids ran \p{around} in the kitchen.
\end{xlist}\end{exe}

See also: \psst{Instrument}, \psst{Manner}

\begin{history}
  The v1 hierarchy distinguished many different subcategories of path descriptions. 
  The labels \sst{Traversed}, \sst{1DTrajectory}, \sst{2DArea}, \sst{3DMedium}, 
  \sst{Contour}, \sst{Via}, \sst{Transit}, and \sst{Course} have all been merged
  with \psst{Path} for v2.
\end{history}

\hierCdef{Direction}

\shortdef{How motion or an object is aimed\slash oriented.}

A \psst{Direction} expresses the orientation of a stationary figure or of a figure's motion.
Prototypical markers\footnote{Known variously as \emph{adverbs}, \emph{particles}, 
and \emph{intransitive prepositions}.}
are \p{away} and \p{back}; \p{up} and \p{down}; 
\p{off}; and \p{out},
provided that no specific \psst{Source} or \psst{Goal} is salient:
\begin{exe}
  \ex The bird flew \choices{\p{up}\\\p{out}\\\p{away}\\\p{off}}.
  \ex I walked \p{over} to where they were sitting.
  \ex The price shot \p{up}.
\end{exe}

In addition, transitive \p{toward(s)}, \p{for}, and \p{at} can 
indicate where something is aimed or directed (but see discussion at \psst{Goal}):
\begin{exe}
  \ex The camera is aimed \p{at} the subject.
  \ex The toddler kicked \p{at} the wall.
\end{exe}

See discussion of \p{away\_from} at \psst{Source}.

\paragraph{Distance.}
\rf{Locus}{Direction} is used for expressions of static distance between two points:
\begin{exe}
  \ex 
    \begin{xlist}
      \ex The mountains are 3~km \choices{\p{away}\\\p{apart}}. (\rf{Locus}{Direction})
      \ex The mountains are 3~km \p{away\_from} our house. (\rf{Locus}{Direction})
    \end{xlist}
\end{exe}
This also applies to distances measured by \emph{travel time} (the amount of time 
is taken to be metonymic for the physical distance):
\begin{exe}
  \ex The mountains are an hour \choices{\p{away}\\\p{apart}}. (\rf{Locus}{Direction})
\end{exe}
Compare \psst{Extent}, which is the length of a path of motion or the amount of change.

\paragraph{Informal direction modifier in location description.}
\begin{exe}
  \ex They live (way) \choices{\p{out} past$_{\text{\rf{Locus}{Path}}}$ the highway.\\
    \p{over} by$_{\psst{Locus}}$ the school} (\rf{Locus}{Direction})
\end{exe}
Cf.~\cref{ex:backIntrans} at \psst{Interval}.

\hierCdef{Extent}

\shortdef{The size of a path, amount of change, or degree.}

This can be the physical distance traversed or the amount of change on a scale:
\begin{exe}
  \ex We ran \p{for} miles.
  \ex The price shot up \p{by} 10\%.
  \ex an increase \p{of} 10\% (\rf{Extent}{Identity})
\end{exe}
For static distance measurements, see \psst{Direction}.

For scalar \p{as} (see \cref{sec:as-as}), \psst{Extent} serves as the function (and sometimes also the role):
\begin{exe}
  \ex\begin{xlist}
    \ex I helped \p{as} much as I could. (\psst{Extent})
    \ex Your face is \p{as} red as a rose. (\rf{Characteristic}{Extent})
    \ex I stayed \p{as} long as I could. (\rf{Duration}{Extent})
  \end{xlist}
\end{exe}
    
\psst{Extent} also covers degree expressions, such as the following PP idioms:
\begin{exe}\ex\begin{xlist}
  \ex I'm not tired \lex[at\_all]{\p{at}\_all}.
  \ex The food is mediocre \lex[at\_best]{\p{at}\_best}.
  \ex You should \lex[at\_least]{\p{at}\_least} try.
  \ex It is the worst \lex[by\_far]{\p{by}\_far}.
  \ex We've finished \lex[for\_the\_most\_part]{\p{for}\_the\_most\_part}.
  \ex It was a success \choices{\lex[in\_every\_respect]{\p{in}\_every\_respect}\\\lex[on\_all\_levels]{\p{on}\_all\_levels}}.
  \ex I hate it when they repeat a song \lex[to\_death]{\p{to}\_death}.
\end{xlist}\end{exe}
Typically these are licensed by a verb or adjective.

\hierBdef{Means}

\shortdef{Secondary action or event presented as playing 
an intermediate causal role in facilitating 
(but not instigating) the main event.}

Prototypically a volitional action, though not necessarily \cref{ex:chlorophyll}. 
A volitional \psst{Means} will often modify an intended result, 
though the outcome can be unintended as well \cref{ex:oops}.
\begin{exe}
  \ex Open the door \p{by} turning the knob.
  \ex They retaliated \choices{\p{by} shooting\\\p{with} shootings}.
  \ex\label{ex:oops} The owners destroyed the company \p{by} growing it too fast.
  \ex\label{ex:chlorophyll} Chlorophyll absorbs the light \p{by} transfer of electrons.
\end{exe}

\psst{Means} is similar to \psst{Instrument}, which is used for causally supporting entities 
and is a kind of \psst{Participant}.

Contrast with \psst{Explanation}, which characterizes \textbf{why} 
something happens. I.e., an \psst{Explanation} portrays the secondary event 
as the causal \emph{instigator} of the main event, whereas \psst{Means} 
portrays it merely as a \emph{facilitator}.

Contrast also with \psst{Manner}. Both \psst{Means} and \psst{Manner} elaborate on the \textbf{how} of an event; 
however, a \psst{Manner} describes a \emph{quality} of the main event, 
rather than specifying a facilitating event.

\begin{history}
  In v1, \psst{Means} was a subtype of \psst{Instrument}, 
  but with the removal of multiple inheritance for v2, 
  the former was moved directly under \psst{Circumstance} 
  and the latter directly under \psst{Participant}.
\end{history}

\hierBdef{Manner}

\shortdef{Qualitative description of a situation, adding color to the main scene.}


\psst{Manner} is used as the scene role for several kinds of descriptors 
which typically license some sort of \pex{How?} question:

\begin{itemize}
  \item The style in which an action is performed or an event unfolds, 
  expressed adverbially (canonical use of the term ``manner''):
  \begin{exe}
    \ex He reacted \choices{\p{with} anger\\\p{in} anger\\angrily}.\footnote{\pex{He reacted \p{out\_of} anger} is \rf{Explanation}{Source}.}
    \ex He reacted \p{with} nervous laughter. [contrast: \psst{Means}]
    \ex\label{ex:wroteinfrench} \choices{I wrote the book\\They chatted} \p{in} French. [contrast \cref{ex:bookinfrench}]
    \ex I made the decision \choices{\p{by} myself\\\p{without} anyone else\\\lex[on\_my\_own]{\p{on}\_~~my~~\_own}}. [see \cref{sec:refl}]
    \ex We talked \p{in} person.
    \ex \rf{Manner}{ComparisonRef}:\begin{xlist}
      \ex You eat \p{like} a pig (eats).
      \ex You smell \p{like} a pig.
    \end{xlist}
    \ex\label{ex:smellOf} \choices{Your father smells\\The soup tastes} \p{of} elderberries. (\rf{Manner}{Stuff}) [also~\cref{ex:smellOfNotCmp}]
    \ex She loves teaching, and it shows \p{in} her smile. (\rf{Manner}{Locus})
  \end{exe}
  
By contrast, depictives characterizing a participant of an event have a scene role of \psst{Characteristic}:
  \begin{exe}
    \ex She entered the room \choices{\p{in} a stupor\\drunk}. (= she was in a stupor when she entered) (\rf{Characteristic}{Locus}) [repeated: \psst{Characteristic}]
  \end{exe}

  \item \emph{What} + \p{like} (\emph{what he looks \p{like}}, etc.): see \cref{ex:whatlike} under \psst{ComparisonRef}.

  \item \textbf{\pex{On a(n)\dots basis}}:
  There seems to be an event-modifying construction \pex{\p{on} a(n) MODIFIER basis} where the 
  modifier phrase reflects the scene role being filled. 
  We use \psst{Manner} as the function:
  \begin{xexe}
    \ex\label{ex:bipartisanBasis} The legislation was passed \lex[on\_a\_bipartisan\_basis]{\p{on}\_a\_~~bipartisan~~\_basis}. (\psst{Manner})
    \ex I see them \lex[on\_a\_daily\_basis]{\p{on}\_a\_~~daily~~\_basis}. (\rf{Frequency}{Manner}) [also \cref{ex:dailyBasis}]
  \end{xexe}

  \end{itemize}

See also: \psst{Means}, \psst{Characteristic}, \psst{ComparisonRef}.

\begin{history}
  In v1, \psst{Manner} was positioned as an ancestor of all 
  categories that license a \emph{How?} question, including 
  \psst{Instrument}, \psst{Means}, and \sst{Contour}, as in \cref{ex:pathmanner}. 
  This criterion was deemed too broad, so \psst{Manner} has no 
  subtypes in v2.
\end{history}

\hierBdef{Explanation}

\shortdef{Assertion of \textbf{why} something happens or is the case.}

This marks a secondary event that is asserted as the reason for the main event or state. 

\begin{exe}
\ex  I went outside \p{because\_of} the smell. 
\ex  The rain is \p{due\_to} a cold front. 
\ex  He reacted \p{out\_of} anger. (\rf{Explanation}{Source})
\ex\begin{xlist}
  \ex He thanked her \p{for} the cookies.
  \ex Thank you \p{for} being so helpful.
\end{xlist}
\end{exe}

When a preposition like \p{after} is used and the relation is temporal as well as causal, 
construal captures the overlap. While \p{since} and \p{as} can also be temporal, 
there are tokens where they cannot be paraphrased respectively with \p{after} and \emph{when}:
\begin{exe}
  \ex I joined a protest \p{after} the shameful vote in Congress. (\rf{Explanation}{Time})
  \ex Her popularity has grown \p{since} she announced a bid for president. (\rf{Explanation}{Time})
  \ex I will appoint him \choices{\p{since}\\\p{as}\\\#\p{after}\\\#when} he is most qualified for the job. (\psst{Explanation})
\end{exe}

Question test: \psst{Explanation} and its subtype \psst{Purpose} license
\pex{Why?} questions.

\hierCdef{Purpose}

\shortdef{A desired outcome presented as contingent on some event, situation, entity, or resource.
The \psst{Purpose} may be specific (e.g., an outcome that somebody tries to achieve by performing an action)
or generic (e.g., an entity that was designed for or incidentally provides some affordance).}

Central usages of \psst{Purpose} explain the motivation behind
(hence subtype of \psst{Explanation}) an action;
the action serves as a means for achieving or facilitating the \psst{Purpose}.
Yet it is possible to complete the action without realizing the purpose.

Prototypical markers include \p{for} and infinitive marker \p{to}:
\begin{exe}
  \ex\begin{xlist}
      \ex He rose \p{to} make a grand speech.
      \ex He rose \p{for} a grand speech.
      \ex surgery \p{to} treat a leg injury
    \end{xlist}
\end{exe}
Something directly manipulated/affected can stand in metonymically 
for the desired event:
\begin{exe}\ex\begin{xlist}
  \ex I went to the store \p{for} eggs. [understood: `to acquire/buy eggs']
  \ex surgery \p{for} a leg injury [understood: `to treat a leg injury']
\end{xlist}\end{exe}

Less central usages present a potentially desirable outcome that could be
brought about thanks to the availability of an entity,
such as a tool, facility, or expendable resource:\footnote{In FrameNet
as of v1.7, these sorts of purposes are labeled as \sst{Inherent\_purpose}.
See, e.g., the example ``MONEY [to support yourself and your family]'' in the \textbf{Money} frame
(\url{https://framenet2.icsi.berkeley.edu/fnReports/data/lu/lu13361.xml?mode=annotation}).}

\begin{exe}
  \ex\label{ex:inherentPurp} \begin{xlist}
    \ex There is some wood \p{to} start a fire (with).
    \ex Do you have a couch \choices{\p{to} sleep on\\\p{for} sleeping on}?
    \ex This place is great \p{for} ping-pong.
    \end{xlist}
\end{exe}

However, this category \emph{excludes} infinitival complements of modal and
aspectual predicates \emph{that lack a direct object}:
\begin{exe}
  \ex\label{ex:NotPurpose}\begin{xlist}\ex He wants/needs \p{to} leave. (\backi)
    \ex He is ready \p{to} leave. (\backi)
    \ex He started/managed \p{to} leave. (\backi)
    \end{xlist}
\end{exe}

The following tests help to clarify the boundaries of \psst{Purpose}:
\begin{enumerate}
\item If a relation can be phrased as \textbf{\pex{\textsc{in order} to \emph{VP}}}
or \textbf{\pex{\textsc{in order} for \emph{NP} to \emph{VP}}}, it is a \psst{Purpose}.
\begin{exe}
  \ex\begin{xlist}
    \ex I arrived (\textsc{in order}) \p{to} see the movie.
    \ex I need \$10 (\textsc{in order}) \p{to} see the movie.
    \ex It takes \$10 (\textsc{in order}) \p{to} see the movie.
    \ex Bring the product to the store (\textsc{in order}) for$_{\text{\backi}}$ us \p{to} service it.
    \ex Bring the product to the store (\textsc{in order}) for$_{\text{\backi}}$ the part \p{to} be replaced.
    \end{xlist}
\end{exe}

\item If a relation can be phrased as \textbf{\pex{for \textsc{the purpose of} \emph{NP}}},
or \textbf{\pex{for \textsc{the purpose of} \emph{<inferred verb>} \emph{NP}}}
(provided that the meaning is not better captured by another label,
e.g.~\psst{Beneficiary}), or \textbf{\pex{that \emph{<someone>} \textsc{intends} to \emph{VP}}}, it is a \psst{Purpose}.
\begin{exe}
  \ex\begin{xlist}
    \ex I arrived \p{for} (\textsc{the purpose of}) the movie.
    \ex I need \$10 \p{for} (\textsc{the purpose of} seeing) the movie.
    \ex I went to the store \p{for} (\textsc{the purpose of} buying) eggs.
    \ex a couch \p{for} (\textsc{the purpose of}) sleeping on
    \ex a couch \p{to} sleep on $\Rightarrow$ a couch \p{for} \textsc{the purpose of} sleeping on
    \ex I found a party (that I \textsc{intend}) \p{to} attend
    \end{xlist}
\end{exe}
Be careful, however, with inserting an inferred verb, as sometimes it is better
captured by another label:
\begin{exe}
  \ex\begin{xlist}
    \ex I babysat \p{for} (\textsc{the purpose of} helping) my aunt and uncle (= as a favor) (\psst{Beneficiary})
    \ex I made a cake \p{for} (\textsc{the purpose of} celebrating) your birthday (= on the occasion of your birthday) (\psst{Circumstance})
    \end{xlist}
\end{exe}

\item If a relation can be phrased as \textbf{\pex{\emph{NP} is good/bad for \emph{V}-ing}}, it is a \psst{Purpose}.\footnote{The positive or negative evaluation is being delimited to a particular purpose:
\cref{ex:GoodGym} is not claiming the gym is good \emph{in general}, just with respect to lifting weights.}
\begin{exe}
  \ex\begin{xlist}
    \ex\label{ex:GoodGym} This is a good gym \p{to} lift weights at.\\
      $\Rightarrow$ This is a good gym \p{for} (lifting) weights.\\
      $\Rightarrow$ This gym is good \p{for} (lifting) weights.
    \ex This cleaner is good \p{for} (cleaning) hardwood floors.
  \end{xlist}
\end{exe}

\item An infinitival modifier of an indefinite pronoun (\w{anything}, \w{someone}) or vague noun (\w{thing}, \w{stuff}) is \psst{Purpose} if the pronoun or vague noun has an entity referent that is involved in the infinitival event.
\begin{exe}
  \ex\begin{xlist}
    \ex I can't think of anybody/a single person \p{to} ask.
    \ex I found something \p{to} eat.
    \end{xlist}
  \ex I found \choices{something\\stuff} \p{to} do. (\w{something}/\w{stuff} does not refer to an entity) (\backi)
\end{exe}

\item An infinitive clause not meeting the above criteria
may express the \textbf{result} of an event, in which case the appropriate
label is \psst{Goal}: see \cref{ex:result}.

\item Many uses of infinitives are not purposes, including complements of
modal\slash aspectual predicates that lack a direct object \cref{ex:NotPurpose},
and syntactic constructions like clausal subjects and certain clefts:
\begin{exe}
  \ex\begin{xlist}
    \ex \p*{To}{to} see the movie is a joy. (\backi)
    \ex It is fun \p{to} see the movie. (\backi)
    \end{xlist}
\end{exe}

\end{enumerate}

\paragraph{Commercial services.}
A special qualification to the above tests applies to commercial scenes
(\w{buying}, \w{paying}, \w{hiring}, \w{costing}, etc.):
for explicitly commercial scenes,\footnote{A more general predicate such as \w{give}, \w{need}, or \w{request} is not considered to evoke a commercial scene, even if it involves money exchanged for a service.}
if the \psst{Purpose} tests pass, the appropriate label is \rf{Theme}{Purpose}.
This expresses that the \psst{Purpose} is not merely a desired outcome,
but is actually promised and paid for in a transaction:
\begin{exe}
  \ex\begin{xlist}
    \ex It costs \$10 \p{to} see the movie. (\rf{Theme}{Purpose})
    \ex I hired John \p{to} fix the problem. (\rf{Theme}{Purpose})
    \end{xlist}
\end{exe}
See additional examples at \cref{ex:services} under \psst{Theme}.

\paragraph{Sufficiency and excess.}
Expressions of sufficiency/excess with an infinitival that passes the above tests
for \psst{Purpose} are labeled \rf{ComparisonRef}{Purpose}.
\begin{exe}
  \ex a bag large enough \choices{\p{for}\\\p{to} hold} the groceries (\rf{ComparisonRef}{Purpose})\\[3pt]
  $\Rightarrow$ a bag large enough \p{for} (\textsc{the purpose of} holding) the groceries
\end{exe}
Sufficiency/excess usages failing the tests are \rf{ComparisonRef}{Goal},
as this is similar to how an infinitival can express a result---cf.~\cref{ex:result} under \psst{Goal}:
\begin{exe}
  \ex a forest canopy too dense \p{to} admit sunlight (\rf{ComparisonRef}{Goal})
  \begin{xlist}
    \ex *a forest canopy too dense in order to admit sunlight
    \ex *a forest canopy too dense for the purpose of admitting sunlight [would imply that somebody was trying to admit sunlight]
  \end{xlist}
\end{exe}
There is additional discussion under \psst{ComparisonRef}.

\paragraph{Versus \psst{Circumstance} for ritualized occasions.}
\psst{Purpose} applies to \p{for} 
when it marks a ritualized activity such as a meal or holiday/commemoration for which the main event describes a \textbf{preparation} stage:
\begin{exe}
  \ex \psst{Purpose}:
    \begin{xlist}
      \ex I walked to this restaurant \p{for} dinner. [walking is not a part of dinner]
      \ex I bought food \p{for} dinner.
      \ex We saved money \p{for} our annual vacation.
    \end{xlist}
\end{exe}
However, if the activity marked by \p{for} is interpreted as \textbf{containing} the main event, 
then we use \psst{Circumstance}: 
\needspace{1em}
\begin{exe}
  \ex \psst{Circumstance}: 
    \begin{xlist}
      \ex We ate there \p{for} dinner.
      \ex I received a new bicycle \p{for} Christmas.
      \ex I always drink eggnog \p{for} Christmas. [at and in celebration of Christmastime]
      \ex We were wearing costumes \p{for} Halloween. 
    \end{xlist}
\end{exe}
If in doubt, \psst{Circumstance} is broader: e.g., \emph{We went there \p{for} dinner} 
if \emph{went} is ambiguous between journeying and attending.

\begin{history}
  In v1, the usages illustrated in \cref{ex:inherentPurp} were assigned a separate label,
  \sst{Function}, which inherited from both \sst{Attribute} and \psst{Purpose}.
  With the introduction of construal, v2.0--v2.3 labeled these \rf{Characteristic}{Purpose}.
  As of v2.4 this distinction has been abandoned, as it was inconsistent with the policies
  for spatial and temporal labels: now purposes of entities are simply \psst{Purpose}.
\end{history}

\hierAdef{Participant}

\shortdef{Thing, usually an entity, that plays a causal role in an event.}

Not used directly---see subtypes.

\hierBdef{Causer}

\shortdef{External instigator of an event.}

This label is for a role introduced by certain constructions (e.g., causative constructions)
in which an event that ordinarily has a certain set of core participants (possibly including a \psst{Force} or \psst{Agent}) is framed as caused by another participant via valency-augmentation. 

In English, this is rarely expressed with a preposition, 
but may be seen in the passive \p{by}-phrase of 
a Caused Motion Construction:
\begin{exe}
    \ex The horse was jumped over the fence \p{by} Claire.\\{} 
    [active paraphrase: Claire jumped the horse over the fence.]
\end{exe}
Note that in the example, the horse is the one actually jumping, and (if annotated) would thus be the \psst{Agent}; whereas Claire is the \psst{Causer} of that semi-volitional jumping event.

A \psst{Causer} may be animate or inanimate.

\begin{history}
  The label \psst{Causer} has a new interpretation in v2.6. 
  The new category is motivated by causative constructions in languages like Hindi, where the case marking of the external causer needs to be accounted for.
  See note at \psst{Force}.
\end{history}

\hierBdef{Force}

\shortdef{Instigator of, and a core participant in, an event.}

\psst{Force} is a generalization of \psst{Agent} with respect to animacy.
\psst{Force} is applied directly to \emph{inanimate} things or forces conceptualized as entities, 
such as in a passive \p{by}-phrase (\cref{sec:passives}):
\begin{exe}
  \ex the devastation of$_{\psst{Theme}}$ the town wreaked \p{by} the fire
  \ex dying \p{of} cancer 
  \ex\rf{Force}{Gestalt}:\begin{xlist} 
    \ex the devastation \p{of} the fire on$_{\psst{Theme}}$ the town
    \ex the fire\p{'s} devastation of$_{\psst{Theme}}$ the town
  \end{xlist}
\end{exe}
The \psst{Force} is sometimes construed as a \psst{Source}:
\begin{exe}
  \ex\rf{Force}{Source}: \begin{xlist}
    \ex the devastation \p{from} the fire
    \ex fatalities \p{from} cancer
    \ex FDR suffered \p{from} polio.
  \end{xlist}
\end{exe}

Contrast: \psst{Causer}

See also: \psst{Instrument}

\begin{history}
  Until v2.6, this was called \sst{Causer} (following VerbNet). 
  In v2.6 it was renamed following UMR \citep{umr} so that \psst{Causer} could be used for external causers.
\end{history}

\hierCdef{Agent}

\shortdef{Animate (and typically volitional) participant
instigating an action or acting in a complementary way to the instigator.}

\psst{Agent} is most directly associated with the passive \p{by}-phrase (\cref{sec:passives}),
but also permits other construals:
\begin{exe}
  \ex the decisive vote \p{by} the City Council
  \ex\rf{Agent}{Gestalt}:\begin{xlist}
    \ex the decisive vote \p{of} the City Council
    \ex the City Council\p{'s} decisive vote
    \ex the president\p{'s} achievements
    \ex they needed Joan\p{'s} help
  \end{xlist}
\end{exe}
When two symmetric \psst{Agent}s are collected in a single NP 
functioning as a set, it is marked as a \psst{Whole} construal:
\begin{exe}
  \ex There was a war \p{between} France and Spain. (\rf{Agent}{Whole})
  \ex a discussion \p{among} the board members (\rf{Agent}{Whole})

\end{exe}

\paragraph{Secondary agents.}
Many event predicates license multiple participants acting in complementary ways
and portrayed as having independent agency.
In such cases, an \psst{Agent} scene role is allowed for either participant,
even if one (typically realized in a syntactically more prominent position, such as subject)
is perhaps understood as slightly more agentive:
\begin{exe}
  \ex I fought in a war \p{against} the Germans. (\rf{Agent}{Beneficiary})
  \ex a match \p{versus} Serena Williams (\rf{Agent}{Beneficiary}) 
  \ex\label{ex:talkWith} I \choices{talked\\argued} \p{with} my roommate about cleaning duties. (\rf{Agent}{Ancillary})
\end{exe}

In other cases, a second ``extra'' participant may be portrayed as accompanying
the first in the event---possibly acting volitionally, and possibly implying some sort of
additional social interaction or alignment of goals, but not strictly necessary
for that kind of event to take place.
\p*{Together\_with}{together\_with} serves as a diagnostic for these participants,
which are labeled simply \psst{Ancillary}.
\begin{exe}
  \ex\label{ex:argueWith} (Together) \p{with} my roommate, I argued that it was unfair of the landlord to increase our rent. (\psst{Ancillary})
\end{exe}

See also: \psst{Org}, \psst{Originator}, \psst{Source}, \psst{Stimulus}, \psst{Ancillary}, \psst{SocialRel}

\begin{history}
In v1 and v2.0--2.4, \psst{Agent} had a subtype \sst{Co-Agent} for (core) secondary agents.
The two were merged in v2.5; the difference between primary and secondary agents
is expressed via construal.
\end{history}

\hierBdef{Theme}

\shortdef{Undergoer that is a semantically core participant in an event or state, 
and that does not meet the criteria for any other label.}

Prototypical \psst{Theme}s undergo (nonagentive\footnote{We distinguish agentivity at the token level, 
unlike VerbNet, where the subject of motion verbs like \emph{arrive} is \psst{Theme} because it need not be agentive.}) motion, are transferred, 
or undergo an internal change of state (sometimes called \emph{patients}).
Adpositional \psst{Theme}s are usually, but not always, construed as something else:
\begin{exe}
  \ex\begin{xlist}
    \ex Quit \p{with} the whining!
    \ex She helped me \p{with} my taxes.
    \ex Don't \choices{bother\\waste time} \p{with} an extra trip.
    \ex I managed to cope \p{with} \choices{the heavy load\\my fear of heights}.
  \end{xlist}
  \ex There's nothing wrong \p{with} the engine.
  \ex Fill the bowl \p{with} water. (\rf{Theme}{Instrument})
  \ex\begin{xlist}
      \ex The food was covered \p{with} grease. (\rf{Theme}{Instrument})
      \ex The food was covered \p{in} grease. (\rf{Theme}{Locus})
    \end{xlist}
  \ex\label{ex:confuse-associate} You shouldn't \choices{confuse\\associate} Mozart \p{with} Rossini. (\rf{Theme}{Ancillary})
  \ex My hovercraft is full \p{of} eels.
  \ex a copy \p{of} the key
  \ex \begin{xlist}
      \ex\label{ex:search-for} Sheldukher \choices{looked\\searched\\fumbled} \p{for} his laser pistol.\\{}
      [contrast with transitive verb plus \psst{Characteristic} in \cref{ex:search-obj-for}]
      \ex Sheldukher \choices{asked\\made a request} \p{for} his laser pistol.
      \ex There is a significant demand \p{for} new housing.
      \ex Let's wait \p{for} \choices{Steve\\more information\\the end of the party}.
    \end{xlist}
  \ex \begin{xlist}
      \ex What happened \p{to} you?
      \ex This species is specific\lex*[specific to]{specific \p{to}}/\lex[native to]{native \p{to}} North America.
      \ex Balancing of risk and reward is \lex[inherent to]{inherent \p{to}} the game.
      \ex The mechanic made a repair \p{to} the engine. (\rf{Theme}{Goal})
      \ex Due to my injury, I am \choices{limited\lex*[limited to]{limited \p{to}}\\
        constrained\lex*[constrained to]{constrained \p{to}}\\
        restricted\lex*[restricted to]{restricted \p{to}}} \p{to} working from home. (\rf{Theme}{Goal})
      \ex It is important to adhere\lex*[adhere to]{adhere \p{to}}/keep\lex*[keep to]{keep \p{to}}/\lex[stick to]{stick \p{to}} your convictions. (\rf{Theme}{Goal})
    \end{xlist}
  \ex
  {\setlength\multicolsep{0pt}%
  \begin{multicols}{2}
    \begin{xlist}
        \sn \psst{Theme}
        \ex the approach \p{of} the waves
        \ex the \choices{death\\murder} \p{of} a salesman
        
        \sn \rf{Theme}{Gestalt}
        \sn the wave\p{s'} approach
        \sn the salesman\p{'s} \choices{death\\murder}
    \end{xlist}
  \end{multicols}}
  \ex Someone in relation to a time period of their life:
    \begin{xlist}
        \ex \p{my} time in grad school (\rf{Theme}{Gestalt})
        \ex I've never seen that in \p{my} life  (\rf{Theme}{Gestalt})
        \end{xlist}
  \ex \begin{xlist}
      \ex The mechanic worked \p{on} the engine.
      \ex We noshed \p{on} snacks.  
      \ex They spent \$500 \p{on} the bicycle. (\rf{Possession}{Theme}) [see \psst{Possession}]
    \end{xlist}
  \ex \begin{xlist}
      \ex There was an increase \p{in} oil prices.
      \ex Bad weather may result \p{in} a delay.
      \ex I'm covered \p{in} bees! (\rf{Theme}{Locus})
      \ex I put a hole \p{in} the box. (= punctured the box) (\rf{Theme}{Locus})
    \end{xlist}
  \ex\label{ex:save-prevent} \begin{xlist}
      \ex The training saved us \p{from} almost certain death. (\rf{Theme}{Source})
      \ex They prevented us \p{from} boarding the plane. (\rf{Theme}{Source})
    \end{xlist}
\end{exe}

\paragraph{Transfer, goods, and services.}
In a commercial scene, goods, services, and money are distinguished. 
\psst{Possession} is used as the scene role 
for goods for sale. 
\psst{Possession} also applies to a piece of property transferred between parties, lost, acquired, or carried, even if no money changes hands.
\psst{Theme} is the scene role for commercial services. 
\psst{Cost} applies to the money asked, paid, or owed.

The construal \rf{Theme}{Purpose} is used for services marked by \p{to}, \p{for}, or similar:
\begin{exe}
  \ex\label{ex:services} Services: \begin{xlist}
    \ex They spent \$500 \p{on} the repairs. (\psst{Theme})
    \ex They charged/asked/paid/owed \$500 \choices{\p{for}\\\p{to} make} the repairs. (\rf{Theme}{Purpose})
    \ex \$500 \choices{\p{for}\\\p{to} make} the repairs was excessive. (\rf{Theme}{Purpose})
  \end{xlist}
\end{exe}
See \psst{Purpose} for additional discussion.
Contrast \cref{ex:goods} under \psst{Possession}.

\paragraph{\emph{Between} and \emph{among}.}
When two symmetric undergoers are collected in a single NP 
functioning as a set, it is marked as a \psst{Whole} construal:
\begin{exe}
  \ex There was a collision in mid-air \p{between} two light aircraft. (\rf{Theme}{Whole})
  \ex Links \p{between} science and industry are important. (\rf{Locus}{Whole})
\end{exe}

\paragraph{Secondary themes.}
Often, multiple similarly situated entities meet the criteria for \psst{Theme},
in which case both are labeled \psst{Theme} for the scene role.\footnote{As with \psst{Agent},
the scene role does not distinguish syntactically more prominent vs.~more oblique positions.}
For example, this can occur in concrete scenes of contact, separation, attachment,
combination, and substitution of two similar entities.

\begin{exe}
  \ex\begin{xlist}
    \ex\label{ex:repl-with} They replaced my old tires \p{with} new ones. [replacement]
    \ex\label{ex:subst-for} They substituted new tires \p{for} my old ones. [replacee]
  \end{xlist}
  \ex\begin{xlist}
    \ex His bicycle collided \p{with} hers. (\rf{Theme}{Ancillary})
    \ex Combine butter \p{with} vanilla. (\rf{Theme}{Ancillary})
    \end{xlist}
  \ex\begin{xlist}
    \ex The boys were separated \p{from} the girls. (\rf{Theme}{Source})
    \ex Keep the dogs \p{from} the cats. (\rf{Theme}{Source})
    \ex The shin bone is connected \p{to} the knee bone. (\rf{Theme}{Goal})
  \end{xlist}
\end{exe}
More abstract examples where a secondary \psst{Theme} PP cooccurs with a \psst{Theme}
direct object include \cref{ex:confuse-associate} and \cref{ex:save-prevent}.

By contrast, for similar scenes where the oblique argument is a ground-like entity 
(larger, less dynamic, more locational, etc.\ than the \psst{Theme}), 
that entity is typically a \psst{Locus}, \psst{Source}, or \psst{Goal}:
\begin{exe}
  \ex Dynamic:\begin{xlist}
    \ex Add vanilla \p{to} the mixture. (\psst{Goal})
    \ex Stir vanilla \p{into} the mixture. (\psst{Goal})
    \ex Detach the cable \p{from} the wall. (\psst{Source})
  \end{xlist}
  \ex Static:\begin{xlist}
    \ex The cable \choices{is attached\\connects} \p{to} the wall. (\rf{Locus}{Goal})
    \ex Protesters were \choices{kept\\missing} \p{from} the area. (\rf{Locus}{Source}) [repeated: \cref{ex:protesters}]
  \end{xlist}
\end{exe}

For creation or transformation of a whole entity (or a group of entities, such as ingredients) 
into another entity, \psst{Source} applies to the initial entity and \psst{Goal} to the result.

Multiple \psst{Theme}s can also be licensed by 2-argument adjectives:
\begin{exe}
  \ex\label{ex:readyFOR} We are ready/eligible/responsible/due \p{for} an upgrade. (\psst{Theme})
\end{exe}

See also: \psst{ComparisonRef}, \psst{Agent}, \psst{Beneficiary}

\begin{history}
  In v1, following many thematic role inventories,
  \sst{Patient} was a distinct label for undergoers that were
  affected (undergoing an internal change of state).
  It was merged into \psst{Theme} for v2 because the affectedness criterion can be subtle
  and difficult to apply.

  In v1 and v2.0--2.4, \psst{Theme} had a subtype \sst{Co-Theme} for core secondary themes.
  In v1, \sst{Co-Patient} was distinguished, in parallel with the \psst{Theme} vs.~\sst{Patient}
  distinction.
  \sst{Co-Patient} was merged with \sst{Co-Theme} in v2.0,
  and \sst{Co-Theme} was merged with \psst{Theme} in v2.5.
\end{history}

\hierCdef{Topic}

\shortdef{Subject matter in communication or cognition, 
or the matter something pertains to.}

This label describes what something is \emph{about} or \emph{in regard to}.
Contrast \psst{Content}, which applies to a message or thought itself.\footnote{For example, \pex{his claim \uline{\p{about} the moon landing}} is \psst{Topic}, whereas \pex{his claim \uline{that the moon landing was faked}} is \psst{Content}. 
}

A variety of prepositions---including the vast majority of occurrences of \p{about}---can 
mark a \psst{Topic}. The following subclasses warrant \psst{Topic} as the scene role:

\begin{itemize}
  \item \textbf{Communication} scenes: the content or subject matter of 
  speech, writing, art, performance, etc.
  \begin{xexe}
    \ex I \choices{gave a presentation\\spoke} \choices{\p{about}/\p{on}} politics.
    \ex They wouldn't stop arguing \p{over} the plan.
    \ex I was accused \p{of} treason.
    \ex a picture \p{of} Whistler's mother
    \ex two counts \p{of} making false statements 
    \ex three \choices{copies\\versions} \p{of} the test
    \ex\rf{Topic}{Identity}---see discussion at \psst{Identity}:
      \begin{xlist}
        \ex the topic/issue/question \p{of} semantics
        \ex the idea \p{of} raising money
      \end{xlist}
    \ex The \choices{ratings\\reviews} \p{for} this film are atrocious.
    \ex I did not hazard a guess \p{as\_to} the cause.
  \end{xexe}
  \item \textbf{Cognition} scenes: the content or subject matter of thought and knowledge---belief, opinion,
  decision, learning, study, interest, expertise, skill, etc.
  \begin{exe}
    \ex\begin{xlist}
      \ex Try not to think \p{about} it.
      \ex We took a minute to \choices{think\\ponder} \p{over} the situation.
      \ex I plan \p{on} going again.
      \ex I am focused \p{on} the task at hand.
      \ex There is not enough research \p{on} the effects of global warming.
      \ex She was dumbfounded \p{as\_to} why the police had done that.
      \ex Think \p{of} all the possibilities!
      \ex I have no memory \p{of} the incident.
      \ex I am aware \p{of} the problem.
      \ex You can have your choice \p{of} chicken or fish.
      \ex I disagree \p{with} that statement.
      \ex I am familiar \p{with} this topic.
      \ex Are you interested \p{in} \choices{politics?\\going to the party?}
      \ex I'm confident \p{in} your abilities.
    \end{xlist}
  \ex\label{ex:Activity}\begin{xlist}
    \ex My daughter excels \choices{\p{in}/\p{at}} sports.
    \ex\label{ex:cookieExpert} I'm \choices{an expert\\talented\\good} \p{at} baking cookies.
    \ex\label{ex:in-activity-Topic} 
      I wouldn't hesitate \p{in} seeing a doctor.\\{} 
      [but see \cref{ex:in-activity-Circumstance} under \psst{Circumstance}, which is syntactically parallel]
    \end{xlist}
  \end{exe}
  \item Relations of \textbf{regard}: the entity, issue, or aspect that the governing 
  predicate pertains to. The relation to the governor may be somewhat loose, 
  skirting the boundary between semantics and information structure.
  \begin{xexe}
    \ex Be reasonable \p{with} your expectations!
    \ex They are transparent \p{with} their fee.
    \ex The discount should apply \p{with} other restaurants too.
    \ex I approached the manager \p{about} the poor service. [implied communication]
    \ex I am a big baby \p{about} needles. [implied cognition]
    \ex The owner wouldn't budge \p{on} the price.
    \ex They came through \p{on} all of their promises.
    \ex She did not do the right thing \p{for} an item that was marked incorrectly.
    \ex I'm fast \p{at} baking cookies. [cf.~\cref{ex:cookieExpert}]
    \ex They have almost anything you could want \choices{\p{when\_it\_comes\_to}\\\p{in\_terms\_of}} spy and surveillance equipment .
  \end{xexe}
\end{itemize}

A few specific governors merit further discussion:

\paragraph{\pex{\lex[agree]{agree}}.}
\begin{xexe}
  \ex Let us agree \p{on} the deal. (\psst{Topic})
  \ex Let us agree \p{to} the deal. (\rf{Topic}{Goal})
  \ex I agree \p{with} the plan. (\psst{Topic})
  \ex I agree \p{with} you. (\rf{Experiencer}{Ancillary})
\end{xexe}

\paragraph{\pex{\lex[answer]{answer}}, \pex{\lex[respond]{respond}}, etc.}
\begin{exe}
  \ex\rf{Topic}{Goal}:\begin{xlist}
    \ex the answer \p{to} the question
    \ex my response \p{to} your question
  \end{xlist}
\end{exe}

For \w{\lex[respond]{respond} \p{with}} and similar, it depends whether the object is an action, 
a device facilitating communication, or some aspect of transferred information:
\begin{xexe}
  \ex He responded to my kick \p{with} a punch. (\psst{Means})
  \ex He responded to my accusation \p{with} a lawsuit. (\psst{Means})
  \ex He responded to my accusation \p{with} dishonest emails. (\psst{Instrument})
  \ex He responded to my accusation \p{with} falsehoods. (\psst{Topic})
\end{xexe}

\paragraph{\pex{\lex[problem with]{problem \p{with}}}, \pex{\lex[experience with]{experience \p{with}}}, etc.} 
These are simply \psst{Topic}:
\begin{xexe}
  \ex \choices{There was\\We had} a problem \p{with} mice in the basement.
  \ex I have limited experience \p{with} numerical methods.
  \ex \choices{I had a bad experience\\my bad experience} \p{with} a vampire.
\end{xexe}

See also: \psst{Stimulus}

Counterpart: \psst{Experiencer}

\begin{history}
  In v1, \lbl{Activity} covered usages such as in \cref{ex:Activity}, 
  but such usages were found to be infrequent and 
  \lbl{Activity} was deemed too narrow.
\end{history}

\hierCdef{Content}

\shortdef{Information that is thought or communicated.}

This label is for communicative or mental information content---i.e., a message or contemplated proposition---expressed more directly than a \psst{Topic}.

In English, this is usually conveyed not with an adposition, but with a \emph{content clause}\index{content clause} possibly marked by a complementizer:

\begin{exe}
  \ex I said/believed/discovered \uline{that it was time to leave}.
  \ex the news \uline{that Nixon had resigned}
\end{exe}

Outside of English, \psst{Content} may apply to adpositional or case marking of quoted speech (e.g., quotative markers).

Contrast \psst{Topic}, which is for the \emph{subject matter} of information, rather than the information itself.

\begin{history}
  In v2.6, \psst{Content} was introduced
  chiefly to accommodate languages (such as Korean) with adpositional quotative marking. 
  It was decided that \psst{Content} should be separate from \psst{Topic}.
  For comparison, FrameNet distinguishes \sst{Content} vs.~\sst{Topic} in the \href{https://framenet2.icsi.berkeley.edu/fnReports/data/frame/Mental_activity.xml}{Mental\_activity frame}, and \sst{Message} vs.~\sst{Topic} in the \href{https://framenet2.icsi.berkeley.edu/fnReports/data/frame/Communication.xml}{Communication frame}. The \psst{Content} supersense encompasses FrameNet's \sst{Message} as well as \sst{Content}.
\end{history}

\hierBdef{Ancillary}

\shortdef{A surplus participant in relation to an event (or state/situation).}

An \psst{Ancillary} participant accompanies another participant in the context of the event.
The \psst{Ancillary}'s participation is presented as similar to\slash in accordance with---but of secondary importance to---that of the other participant.

Sometimes called \emph{comitative}\index{comitative}.

Prototypical prepositions are \p{with}, \p{without}, \p{along\_with},
\p{together\_with}, and \p{together}: 

\begin{xexe}
  \ex Could you walk \choices{\p{with}\\\p{along\_with}\\\p{together\_with}} me to the store?
  \ex Can you go to the store \p{without} me?
  \ex Can we go to the store \p{together}?
\end{xexe}

A participant may be considered surplus/secondary for just the function or also at the scene level.
\psst{Ancillary} is the \emph{function} for adpositions like \p{with} that signal asymmetric togetherness
or co-participation. More specific spatial and configurational
(possession, part-whole, membership, etc.)\ relations take precedence at the \emph{scene} level:

\begin{exe}
  \ex\begin{xlist}
    \ex\label{ex:nextToMom} The girl is standing \p{next\_to} her mother. (\psst{Locus})
    \ex\label{ex:withMom} The girl is standing \p{with} her mother. (\rf{Locus}{Ancillary})
  \end{xlist}
  \ex\begin{xlist}
    \ex The girl is \p{by} the pigeon. (\psst{Locus})
    \ex The girl is \p{with} the pigeon. (presumably, close to and interacting with it or paying special attention to it) (\rf{Locus}{Ancillary})
    \ex Put the fork \p{with} the knives. (\rf{Goal}{Ancillary})
  \end{xlist}
  \ex\begin{xlist}
    \ex I work \p{with} Steve. (\rf{SocialRel}{Ancillary})
    \ex I am \p{with} Grunnings. (\rf{Org}{Ancillary})
    \ex people \p{with} Grunnings (= Grunnings employees) (\rf{Org}{Ancillary})
  \end{xlist}
\end{exe}

Some predicates have a role of primary semantic importance expressed via a \p{with}-PP.\footnote{These can be called semantically core roles, though making a core\slash non-core distinction is in general problematic.}
In such cases, \psst{Ancillary} should be the function only.
However, for many predicates it may be difficult to decide whether
\psst{Ancillary} should also be the scene role.
As a diagnostic, we test whether \p{together\_with} can be used---if not,
there is another role of primary importance to the scene.\footnote{For
the preposition \p{without}, the test is whether \p{together\_with} expresses its negation.}

\paragraph{Pure \psst{Ancillary} (scene and function).}
These license \emph{together}-insertion:

\begin{exe}
  \ex\begin{xlist}
    \ex I am admiring the paintings (together) \p{with} my friend\\{}
    [we probably infer that ``friend'' is paired with ``I'', and thus also admiring or at least viewing the paintings, but this requires pragmatics]
    \ex I am admiring the paintings (together) \p{with} the statues (= I am admiring the paintings, and the statues as well)\\{}
    [we infer that ``statues'' is paired with ``paintings'', and thus also being admired, but this requires pragmatics]
  \end{xlist}
  \ex (Together) \p{with} the president, the prime minister signed the declaration [explicit: president is together with somebody in the context of signing; inferred: president is together with the prime minister, and they probably both signed]
  \ex\begin{xlist}
    \ex I was traveling (together) \p{with} my friend/infant
    \ex my travels (together) \p{with} my friend
  \end{xlist}
  \ex I fought (together) \p{with} her to cure cancer. (= we fought on the same side)
\end{exe}

\paragraph{\psst{Ancillary} function only, predicate-licensed scene role.}
These resist \emph{together}-insertion:

\begin{exe}
  \ex\rf{Agent}{Ancillary}:\begin{xlist}
    \ex Why don't you talk (*together) \p{with} your friend?
    \ex I fought (*together) \p{with} her for a week. (= we argued opposite sides)
    \ex Please trade places (*together) \p{with} John.
  \end{xlist}
  \ex I'll have to check (*together) \p{with} my supervisor. (\rf{Recipient}{Ancillary})
  \ex He was not ready to share a house (*together) \p{with} her. (\rf{Possessor}{Ancillary})
  \ex I agree (*together) \p{with} John. (= we share the same opinion) (\rf{Experiencer}{Ancillary})
  \ex Don't compare me (*together) \p{with} my sister! (\rf{ComparisonRef}{Ancillary})
  \ex Why do people associate bats (*together) \p{with} death? (\rf{Theme}{Ancillary})
\end{exe}
See further examples at \psst{Theme}.

\paragraph{Item in one's possession.}
If the object denotes an item that the governor has on hand in their possession,
then the construal \rf{Possession}{Ancillary} is used:
\begin{exe}
  \ex I walked in \p{with} an umbrella. (\rf{Possession}{Ancillary})
\end{exe}

\paragraph{X\textsubscript{\emph{i}} \emph{bring}/\emph{take}/\dots~Y \p{with} PRON\textsubscript{\emph{i}}.}
This construction involves a \p{with}-PP that is coreferent with the subject.
The most basic meanings of these argument structures 
bundle motion, possession, location, and accompaniment. 
In such cases, the \p{with} is analyzed as \rf{Locus}{Ancillary}:\footnote{\label{fn:bring-with}\w{Bring} and similar verbs (\w{take}, \w{carry}, etc.)\ specify
motion-with-possession in their most literal sense (e.g., bringing a backpack).
If applying supersenses also to subjects and objects (\citealp{shalev-19}; see also \cref{fn:orig-subj-obj,fn:recip-subj-obj})
we would use \psst{Possession}/\psst{Possessor} as the scene roles of the subject/object respectively.
But if the object is volitional (e.g., bringing a friend),
the possession is bleached away,
so just \psst{Agent}/\psst{Theme} would apply to the subject/object.
In either case, the \p{with}-PP emphasizes that the other entity is located with the bringer,
so it receives \rf{Locus}{Ancillary}.}
\begin{xexe}
  \ex I brought my backpack/friend \p{with} me. (\rf{Locus}{Ancillary})\\{}
  [emphasizes that the backpack\slash friend is located with the speaker]
  \ex I brought my backpack/friend.
\end{xexe}

When the verb in this construction bears an extended meaning of stative or abstract accompaniment, \p{with} may be more appropriately analyzed as \rf{Ensemble}{Ancillary}:
\begin{exe}
    \ex\label{ex:bring-with-stative} The new year brings \p{with} it many challenges. (\rf{Ensemble}{Ancillary}) [cf.~\cref{ex:go-with-stative} at \psst{Ensemble}] 
\end{exe}

\paragraph{\p*{Together}{together}.}
The word \p{together}, when not followed by \p{with},
can denote reciprocal accompaniment and is analyzed like \w{\p{with} each other}:

\begin{xexe}
  \ex We were sitting/eating/working \p{together}.
  \ex The duck and the chick are \p{together}. (\rf{Locus}{Ancillary})
  \ex John and Mary are \p{together} (= a couple). (\rf{SocialRel}{Ancillary})
\end{xexe}

\paragraph{Versus \psst{Ensemble}.}
\psst{Ancillary} descibes a relation of an entity to an event/situation,
whereas \psst{Ensemble} is used for a relation directly between entities.

See also: \psst{Instrument}, \psst{Manner}

\begin{history}
  Prior to v2.5, a single label, \sst{Accompanier} (under \psst{Configuration}),
  covered both entity--entity and event--entity relations.
  In v2.5, \sst{Accompanier} was split into \psst{Ancillary} and \psst{Ensemble}.
\end{history}

\hierBdef{Stimulus}%
\shortdef{That which is perceived or experienced (bodily, perceptually, or emotionally).}

\psst{Stimulus} does not seem to have any prototypical adposition 
in the languages we have looked at. In English, it can be construed in several ways:
\begin{exe}
  \ex my affection \p{for} you (\rf{Stimulus}{Beneficiary})
  \ex scared \p{by} the bear (\rf{Stimulus}{Force})
  \ex You should \choices{listen\\pay attention} \p{to} the music. (\rf{Stimulus}{Goal})
  \ex \rf{Stimulus}{Direction}:
    \begin{xlist}
      \ex We were looking \p{at} the photo.
      \ex\label{ex:AngryAt} I was angry \p{at} him. [cf.~\cref{ex:AngryWith}]
      \ex I startled \p{at} the noise.
    \end{xlist}
  \ex \rf{Stimulus}{Topic} is assigned to cases where the PP describes the topic or content of one's emotion: 
    \begin{xlist}
      \ex I care \p{about} you.
      \ex That's what I love \p{about} the show.
      \ex I took\_pride \p{in} the results.
      \ex I was \choices{proud \p{of}\\happy \p{with}} the results.
      \ex\label{ex:AngryWith} I was angry \p{with} him. [cf.~\cref{ex:AngryAt}]
      \ex\label{ex:InLoveWith} I was in$_{\text{\rf{Characteristic}{Locus}}}$ love \p{with} him. [cf.~\cref{ex:InLove}]
      \ex They bored me \p{with} their incessant talk about cats.
    \end{xlist}
  \ex\label{ex:StimBen} \rf{Stimulus}{Beneficiary}:
    \begin{xlist} 
      \ex Her disdain \p{for} customers was apparent.
      \ex He has/feels compassion \choices{\p{towards}\\\p{for}} animals. 
    \end{xlist}
  \ex I am \choices{thankful\\grateful} \p{for} your help. (\rf{Stimulus}{Explanation})
\end{exe}
See also: \psst{Topic}, \psst{Beneficiary}

Counterpart: \psst{Experiencer}

\hierBdef{Experiencer}

\shortdef{Animate who is aware of a bodily sensation, perception, emotion, or mental state.}

\psst{Experiencer} does not seem to have any prototypical adposition 
in the languages we have looked at. In English, it can be construed in several ways:
\begin{exe}
  \ex\begin{xlist}
    \ex the anger \p{of} the students (\rf{Experiencer}{Gestalt})
    \ex the student\p{s'} anger over$_{\psst{Topic}}$ the decision (\rf{Experiencer}{Gestalt})
  \end{xlist}
  \ex\begin{xlist} 
    \ex Running is enjoyable \p{for} me. (\rf{Experiencer}{Beneficiary})
    \ex The pizza was (too) salty \p{for} me. (\rf{Experiencer}{Beneficiary})
  \end{xlist}
  \ex\begin{xlist}
    \ex It feels hot \p{to} me. (\rf{Experiencer}{Goal})
    \ex That was astounding \p{to} me. (\rf{Experiencer}{Goal})
    \ex This is \p{my} favorite movie. (\rf{Experiencer}{Gestalt})
  \end{xlist}
  \ex\begin{xlist}
    \ex The answer is known \p{by} me. (\rf{Experiencer}{Agent})
    \ex The answer is known \p{to} me. (\rf{Experiencer}{Goal})
    \ex That is \p{my} opinion. (\rf{Experiencer}{Gestalt})
    \ex That was \p{my} experience. (\rf{Experiencer}{Gestalt})
  \end{xlist}
\end{exe}

Bodily events with an \psst{Experiencer} are limited to \textbf{perceptions} like seeing and hearing, and \textbf{sensations} such as pain and hunger.
The undergoer of an involuntary bodily event like sneezing, bleeding, falling asleep, breaking a limb, or dying would instead be a \psst{Theme}, as these events are primarily physical in nature.

One whose \textbf{mental state} (including events of knowledge, memory, belief, desire, intention) or \textbf{emotion} is described is an \psst{Experiencer}. 
However, the individual is an \psst{Agent} if exhibiting or acting on their thoughts\slash emotions, even internally (e.g.~making a decision).

Less canonically, \psst{Experiencer} applies to semi-pragmatic usages meaning `from the perspective of':\footnote{Interestingly, 
many uses of \p{for} carry an information structural association of delimiting the scope of an assertion. 
\pex{\p*{For}{for} John, the party was not fun at all} makes no commitment regarding how fun the party was to others. 
\pex{This food is good \p{for}$_{\psst{Purpose}}$ dinner\slash \p{for}$_{\psst{Beneficiary}}$ folks with dietary constraints} 
and \pex{He is short \p{for}$_{\psst{ComparisonRef}}$ a basketball player} also have this property. 
As the present scheme targets semantic relations, it is not equipped to formalize pragmatic aspects of the meaning.}
\begin{exe}
  \ex \begin{xlist}
    \ex \p*{For}{for} John, the party was not fun at all. (\rf{Experiencer}{Beneficiary})
    \ex \p*{For}{for} John, there was no reason to attend. (\rf{Experiencer}{Beneficiary})
  \end{xlist}
\end{exe}

Elsewhere, the term \emph{cognizer} is sometimes used for one whose 
mental state is described.

Counterpart: \psst{Stimulus} or \psst{Topic}

\hierBdef{Originator}

\shortdef{Animate who is the initial possessor or creator/producer of something,
including the speaker/communicator of information. 
Excludes events where transfer/communication is not framed as unidirectional.}

A ``source'' in the broadest sense of a starting point/condition. 
Contrasts with \psst{Recipient} if there is transfer/communication.

English construals:\footnote{\label{fn:orig-subj-obj}If we consider subject position as an \psst{Agent} construal
and direct object position as a \psst{Theme} construal (\citealp{shalev-19}; cf.~\cref{fn:bring-with,fn:recip-subj-obj}), then we can add examples like
\pex{\uline{She} talked to her editor} (\rf{Originator}{Agent}) and
\pex{They robbed \uline{her} of her life savings} (\rf{Originator}{Theme}).
\psst{Originator} does not apply to the subject of events like \pex{exchange} or \pex{talk/chat (with)},
which involve a back-and-forth between 
multiple \psst{Agent}s.}
\begin{exe}
  \ex \rf{Originator}{Agent} (passive-\p{by} or adnominal \p{by}):
  \begin{xlist}
    \ex\label{ex:worksBy} works \p{by} Shakespeare [cf.~\cref{ex:worksOf,ex:worksGen}]
    \ex The telephone was invented \p{by} Alexander Graham Bell.
    \ex The story was \choices{given\\told} to$_{\text{\rf{Recipient}{Goal}}}$ her \p{by} her editor.
  \end{xlist}
  \ex \rf{Originator}{Source}:
  \begin{xlist}
    \ex\label{ex:worksOf} works \p{of} Shakespeare [cf.~\cref{ex:worksBy,ex:worksGen}]
    \ex The story was obtained \p{from} an anonymous White House employee.
    \ex I bought it \p{from} this company.
    \ex I heard the news \p{from} Larry.
  \end{xlist}
  \ex \rf{Originator}{Gestalt}: \begin{xlist}
    \ex \label{ex:worksGen} Shakespeare\p{'s} works [cf.~\cref{ex:worksOf,ex:worksBy}]
    \ex \label{ex:rodin} Rodin\p{'s} sculptures
    \ex \label{ex:restFood} the restaurant\p{'s} food
    \ex John\p{'s} \choices{question\\speech}
  \end{xlist}
\end{exe}

\paragraph{\emph{learn from}.\lex*[learn from]{learn \p{from}}} If the source of learning is an individual 
(or group of individuals, organization, etc.)\ who provides information, 
\rf{Originator}{Source} applies. Otherwise, it is simply \psst{Source}:
\begin{exe}
  \ex We learned a lot \p{from} Miss Zarves. (\rf{Originator}{Source})
  \ex We learned a lot \p{from} that \choices{book\\experience}. (\psst{Source})
\end{exe}


\begin{history}
  \psst{Originator} merges v1 labels \sst{Donor/Speaker} and \sst{Creator}, 
  which were difficult to distinguish in the case of authorship.
  \sst{Donor/Speaker} was a subtype of \sst{InitialLocation}, which 
  inherited from \sst{Location} and \psst{Source}. 
  \sst{Creator} was a subtype of \psst{Agent}.
  Moving \psst{Originator} directly under \psst{Participant} 
  puts it in a neutral position with respect to its possible construals.
\end{history}

\hierBdef{Recipient}

\shortdef{The party (usually animate) that is the endpoint of (actual or intended) transfer of a thing or message, 
becoming the final \psst{Possessor} or \psst{Gestalt}.
Excludes events where transfer/communication is not framed as unidirectional.}
A ``goal'' in the broadest sense of an ending point/condition. 
Contrasts with \psst{Originator}.

English construals:\footnote{\label{fn:recip-subj-obj}
\Citet{shalev-19} propose generalizing SNACS supersenses to include subjects and objects (see also \cref{fn:bring-with,fn:orig-subj-obj}).
If subject position is viewed as an \psst{Agent} construal,
then an active subject of a transfer verb like \pex{get} or \pex{receive} is \rf{Recipient}{Agent}.
If direct object position is viewed as a \psst{Theme} construal,
then \pex{She informed \uline{her editor}} is \rf{Recipient}{Theme}.}
\begin{exe}
  \ex She \choices{gave the story\\spoke} \p{to} her editor. (\rf{Recipient}{Goal})
  \ex What title did you give \p{to} your essay? [inanimate] (\rf{Recipient}{Goal})
  \ex news \p{for} our readers (\rf{Recipient}{Direction})
  \ex He is yelling \p{at} me to get ready! (\rf{Recipient}{Direction}\footnote{While \emph{yell \p{at}} 
often has a connotation of shouting criticism towards somebody, 
and criticism would suggest \psst{Beneficiary},
the \psst{Recipient} aspect of the meaning is more explicit and essential:
yelling from a distance at someone does not imply criticism, 
and criticism about someone who is absent is not yelling at them.})
  \ex The news was not well received \p{by} the White House. (\rf{Recipient}{Agent})
  \ex Timmy\p{'s} piano lesson (\rf{Recipient}{Gestalt})
  \ex I'll have to check \p{with} my supervisor. (\rf{Recipient}{Ancillary})
\end{exe}

\psst{Recipient} does not apply to events like \pex{exchange/talk/chat (\p{with})}, 
which involve a back-and-forth between 
multiple \psst{Agent}s:
\begin{exe}
  \ex She \choices{swapped stories\\chatted} \p{with} her friends. (\rf{Agent}{Ancillary})
\end{exe}

See also: \psst{Beneficiary}

\begin{history}
  In v1, \psst{Recipient} was the counterpart to \sst{Donor/Speaker}:
  \psst{Recipient} was a subtype of \sst{Destination}, which 
  inherited from \sst{Location} and \psst{Goal}. 
  Moving \psst{Recipient} directly under \psst{Participant} 
  puts it in a neutral position with respect to its possible construals.
\end{history}

\hierBdef{Cost}

\shortdef{An amount (typically of money) that is linked to an item or service 
that it pays for\slash could pay for, or given as the amount earned or owed.} 

The governor may be an explicit commercial scenario:
\begin{exe}
  \ex I \choices{bought\\sold} the book \p{for} \$10.
  \ex I got a refund \p{of} \$10.
  \ex\rf{Cost}{Locus}: \begin{xlist}
    \ex The book is \choices{priced\\valued} \p{at} \$10.
    \ex I bought it \p{at} a great price/rate.
  \end{xlist}
\end{exe}
Or the \psst{Cost} may be specified as an adjunct with a non-commerical governor:
\begin{exe}
  \ex You can ride the bus \p{for} \choices{free\\\$1}.
\end{exe}
\psst{Cost} is specifically about payment requested in exchange for goods or services (including income and revenue). 
If an amount of money is simply treated as property given, acquired, or possessed, then \psst{Possession} is appropriate:
\begin{exe}
  \ex I bestowed the winner \p{with} \choices{a bicycle\\\$100}. (\psst{Possession}) [repeated at \psst{Possession}]
\end{exe}
See discussion of transfer, goods, and services at \psst{Possession} and \psst{Theme}.

\begin{history}
  This category was not present in v1, which had the broader category \sst{Value}. 
  VerbNet \citep{verbnet,palmer-17} has a similar category called \sst{Asset}; we chose the name 
  \psst{Cost} to emphasize that it describes a relation rather than an entity type 
  (it does not apply to money with a verb like \pex{possess} or \pex{transfer}, 
  for instance).
\end{history}

\hierBdef{Beneficiary}

\shortdef{Animate or personified undergoer that is (potentially) 
advantaged or disadvantaged by the event or state.}

This label does not distinguish the polarity of the relation 
(helping or hurting, which is sometimes termed \emph{maleficiary}).

\begin{exe}
  \ex Vote \choices{\p{for}\\\p{against}} Pedro!
  \ex Junk food is bad \p{for} your health.
  \ex My parrot died \p{on} me.
  \ex\begin{xlist} 
    \ex These are clothes \p{for} children.
    \ex These are children\p{'s} clothes. (\rf{Beneficiary}{Possessor})
  \end{xlist}
  \ex Fortunately \p{for} the turkey('s future), he received a presidential pardon.
\end{exe}

\noindent Specific subclasses include:
\begin{itemize}
\item	Animate who will potentially experience a benefit or harm as a result of something 
but is not an experiencer or recipient of the main predicate itself. 
(May be an experiencer or recipient of the result.)
\item	Animate target of emotion or behavior, discussed below. 
\item	Animate who someone supports or opposes (e.g., \emph{vote \p{for}}, 
\emph{cheer \p{for}}, \emph{Hooray \p{for}}). 
\item Intended user/usee: 
  \begin{exe}
    \ex (We sell) clothes \p{for} children
    \ex a gallows \p{for} criminals
    \ex This is the car \p{for} you! [advertising idiom]
  \end{exe}
\item	Something characterized as good/appropriate (or not) for some kind of 
\textbf{animate user or usee}, delimiting the applicability of a descriptor 
to that kind of individual: 
  \begin{exe}
    \ex \begin{xlist}
      \ex This place is great \p{for} young children.
      \ex This is a great place \p{for} young children.
    \end{xlist}
  \end{exe}
\end{itemize}
The first and last items above have analogues with \psst{Purpose}. 
The key difference is that \psst{Beneficiary} applies to an animate participant, 
whereas \psst{Purpose} applies to an intended consequence or one of its inanimate participants.

\paragraph{Targets of behavior versus emotion.}
A preposition can mark an individual in the context of evaluating how someone else is treating them, 
with a noun or adjective governor. 
If behavior is more salient than emotion, then \psst{Beneficiary} is the scene role. 
If emotion is highly salient, then \psst{Stimulus} is the scene role.
\begin{exe}
  \ex Behavior-focused:
    \begin{xlist}
      \ex She exhibits rudeness \p{towards} customers. (\rf{Beneficiary}{Direction})
      \ex He is \choices{rude\\condescending} \p{to} women. (\rf{Beneficiary}{Goal})
      \ex He is gentle and compassionate \p{with} animals. (\rf{Beneficiary}{Theme})
    \end{xlist}
  \ex Emotion-focused, repeated from \cref{ex:StimBen}:
    \begin{xlist} 
      \ex Her disdain \p{for} customers was apparent. (\rf{Stimulus}{Beneficiary})
      \ex He has/feels compassion \choices{\p{towards}\\\p{for}} animals. (\rf{Stimulus}{Beneficiary})
    \end{xlist}
\end{exe}
Note that the emotion-focused examples can describe private emotional states directly, 
while the behavior-focused examples are behavior-based judgments or inferences about emotional states.

An obligation directed at somebody is analyzed like targeted behavior:
\begin{exe}
  \ex We have a solemn responsibility \p{to} our armed forces. (\rf{Beneficiary}{Goal})
\end{exe}

Similar to the behavior-focused examples, inanimate causes can have the potential to positively or negatively
affect somebody. Ability and permission modalities are included here:
\begin{exe}
    \ex\begin{xlist}
      \ex The strategy is \choices{beneficial \\ risky \\ an option} \p{for} investors. (\psst{Beneficiary})
      \ex The strategy \choices{is helpful \\ poses a risk \\ is available} \p{to} investors. (\rf{Beneficiary}{Goal})
    \end{xlist}
\end{exe}

\paragraph{Versus \psst{Recipient}.}
\psst{Beneficiary} applies to the classic English benefactive construction 
where it is ambiguous between assistance and intended-transfer:
\begin{exe}
  \ex John baked a cake \p{for} Mary. [to help Mary out, and/or with the intention of giving her the cake]
\end{exe}
However, if transfer (or communication) is the main semantics of the scene 
and benefit or harm is no more than an inference, then the scene role is \psst{Recipient}:
\begin{exe}
  \ex a \choices{message\\gift} \p{for} my mother (\rf{Recipient}{Direction})
  \ex a \choices{package} \p{for} the front office (\rf{Recipient}{Direction})
\end{exe}

See also: \psst{Experiencer}, \psst{Org}

\hierBdef{Instrument}

\shortdef{An entity that facilitates an action by applying intermediate causal force.}

Prototypically, an \psst{Agent} intentionally applies the \psst{Instrument} 
with the purpose of achieving a result:
\begin{exe}\ex\begin{xlist}
  \ex I broke the window \p{with} a hammer.
  \ex I destroyed the argument \p{with} my words.
\end{xlist}\end{exe}
Less prototypically, the action could be unintentional:
\begin{exe}
  \ex I accidentally poked myself in the eye \p{with} a stick.
\end{exe}
The key is that the \psst{Instrument} is not sufficiently ``independently causal'' 
to instigate the event.

However, to downplay the agency of the individual operating the instrument, 
the instrument can be placed in a passive \p{by}-phrase, 
which construes it as the instigator:
\begin{exe}\ex\label{ex:passiveInstrument}\begin{xlist}
  \ex The window was broken \p{by} the hammer. (\rf{Instrument}{Force})
  \ex My headache was alleviated \p{by} aspirin. (\rf{Instrument}{Force})
\end{xlist}\end{exe}
Note that the examples in \cref{ex:passiveInstrument} can be rephrased 
in active voice with the \psst{Instrument} as the subject.

A device serving as a mode of transportation or medium of communication 
counts as an \psst{Instrument}, but is often construed as a \psst{Locus} or \psst{Path}:
\begin{exe}
  \ex Communicate \p{by} \choices{phone\\email}. (\psst{Instrument})
  \ex Talk \p{on} the phone. (\rf{Instrument}{Locus})
  \ex Send it \choices{\p{over}\\\p{via}} email. (\rf{Instrument}{Path})
  \ex Travel \p{by} train. (\psst{Instrument})
  \ex Escape \p{with} a getaway car. (\psst{Instrument})
  \ex Escape \p{in} the getaway car. (\rf{Instrument}{Locus})
\end{exe}
This includes some expressions which incorporate the \psst{Instrument} 
in a noun:
\begin{exe}
  \ex ride \p{on} horseback (\rf{Instrument}{Locus})
  \ex hold \p{at} knifepoint (\rf{Instrument}{Locus})
\end{exe}
Other non-prototypical instruments that can be construed as paths 
include waypoints from \psst{Source} to \psst{Goal}, 
and people\slash organizations serving as intermediaries:
\begin{exe}
  \ex We flew to London \p{via} Paris. (\rf{Instrument}{Path})
  \ex I found out the news \p{via} Sharon. (\rf{Instrument}{Path})
  \ex Joan bought her house \p{through} a real estate agent. (\rf{SocialRel}{Instrument})
  \ex For my Honda I always got replacement parts \p{through} the dealership. (\rf{Org}{Instrument})
\end{exe}

Conversely, roadways count as \psst{Path}s but can be construed as \psst{Instrument}s:
\begin{exe}
  \ex Escape \p{through} the tunnel. (\psst{Path})
  \ex Escape \p{by} tunnel. (\rf{Path}{Instrument})
\end{exe}

Compare \psst{Means}, which is used for facilitative events rather than entities.
See also \psst{Topic}.

\hierAdef{Configuration}

\shortdef{Thing, usually an entity or property, that is involved 
in a static relationship to some other entity.}

Not used directly---see subtypes.

\hierBdef{Identity}

\shortdef{A category being ascribed to something, 
or something belonging to the category denoted by the governor.}

Prototypical prepositions are \p{of} (where the governor is the category) 
and \p{as} (where the object is the category):
\begin{exe}
  \ex\label{ex:stateof} the state \p{of} Washington [as opposed to the city\footnote{\Citet[p.~448]{cgel} use the term \emph{appositive oblique} when \p{of} marks a proper name in relation to a category such as \w{month}, \w{city}, or \w{state}.}]
  \ex The liberal state \p{of} Washington has not been receptive to Trump's message.
  \ex \p*{As}{as} a liberal state, Washington has not been receptive to Trump's message.
  \ex\label{ex:ascolleague} I like Bob \p{as} a colleague. [but not as a friend]
  \ex\label{ex:as-a-child2} I played the piano \p{as} a child. (\rf{Time}{Identity}) [also~\cref{ex:as-a-child}]
  \ex What a gem \p{of} a restaurant! [exclamative idiom: both NPs are indefinite]
  \ex the \choices{problem/task/hassle} \p{of} raising money
  \ex the age \p{of} eight
  \ex a \choices{height/distance} \p{of} 10m 
  \ex They did a great job \p{of} cleaning my windows.
  \ex\label{ex:shell} \rf{Topic}{Identity}, with a governing noun in the domain of communication or cognition:
    \begin{xlist}
      \ex the \choices{topic/issue/question} \p{of} semantics
      \ex the idea \p{of} raising money
    \end{xlist}
\end{exe}
Something may be specified with a category in order to disambiguate it \cref{ex:stateof}, 
or to provide an interpretation or frame of reference with which that entity is to be considered.
In some cases, like \cref{ex:shell}, the category is a \emph{shell noun} \citep{schmid-00} 
requiring further specification.

Categorizations may be situational rather than permanent/definitional:
\begin{exe}\ex\label{ex:assituational}\begin{xlist}
  \ex She appears \p{as} Ophelia in \emph{Hamlet}.
  \ex He is usually a bartender, but today he is working \p{as} a waiter.
\end{xlist}\end{exe}

Paraphrase test: ``(thing) IS (category) [in the context of the event]'': 
``Washington is a liberal state'', ``opening a new business is a hassle'', 
``She is Ophelia'', etc. Note that \p{as}+category may attach syntactically 
to a verb, as in \cref{ex:ascolleague} and  \cref{ex:assituational}, 
rather than being governed by the item it describes.

If the object of the preposition is a property (as opposed to a category), 
the scene role is \psst{Characteristic}:
\begin{exe} 
  \ex Adnominal: \rf{Characteristic}{Identity}\begin{xlist}
    \ex a car \p{of} high quality
    \ex a man \p{of} honor
    \ex a business \p{of} that sort [contrast with \psst{Species}, \cref{sec:Species}]
  \end{xlist}
  \ex Secondary predicate adjective: \rf{Characteristic}{Identity}\begin{xlist}
    \ex She described him \p{as} sad.
    \ex He strikes me \p{as} sad.
  \end{xlist}
\end{exe}

See also: \psst{ComparisonRef}

\begin{history}
  Generalized from v1, where it was called \sst{Instance} and restricted 
  to the ``(category) \p{of} (thing)'' formulation. 
  The relevant usages of \p{as} were labeled \sst{Attribute}.
\end{history}

\hierBdef{Species}

\shortdef{A category qualified by \w{sort}, \w{type}, \w{kind}, \w{species}, \w{breed}, etc. 
Includes \w{variety}, \w{selection}, \w{range}, \w{assortment}, etc.\ 
meaning `many different kinds'.}

\begin{exe}\ex\begin{xlist}
  \ex that sort \p{of} business
  \ex A good type \p{of} ant to keep is the red ant .
  \ex certain strains \p{of} \emph{Escherichia coli}
  \ex Modern breeds \p{of} these homing pigeons return reliably
  \ex Some poor sap applied the wrong brand \p{of} paint
  \ex This store offers a wide selection \p{of} footstools
\end{xlist}
  \ex \choices{an example\\the epitome} \p{of} Italian Renaissance architecture
\end{exe}

\psst{Species} is \emph{not} used if the sort/variety noun 
is the object rather than the governor:
\begin{exe}
  \ex a business \p{of} that sort (\psst{Characteristic})
\end{exe}

\hierBdef{Gestalt}

\shortdef{Generalized notion of ``whole'' understood with reference to 
a component part, possession, set member, or characteristic. 
See \psst{Characteristic}.}

\psst{Gestalt}---the supercategory of \psst{Whole}, \psst{Possessor}, \psst{Org}, and \psst{QuantityItem}---applies
directly for entities and eventualities which can loosely be conceptualized as 
containing or possessing something else, but for which 
neither \psst{Whole} nor \psst{Possessor} is a good fit.

\paragraph{Properties.}
The holder of a property if the property is the governor:
\begin{exe}
    \ex {\setlength\multicolsep{0pt}%
    \begin{multicols}{2}
      \begin{xlist} 
        \ex the blueness \p{of} the sky
        \ex the size \p{of} the crowd
        \ex the price \p{of} the tea
        \ex the start time \p{of} the party
        \ex the power \p{of} the president
        
        \sn the sky\p{'s} blueness
        \sn the crowd\p{'s} size
        \sn the tea\p{'s} price
        \sn the party\p{'s} start time
        \sn the president\p{'s} power
      \end{xlist}
    \end{multicols}}
    \ex It was the chairman\p{'s} fault/responsibility/right.
    \ex\label{ex:poss-menu} the restaurant\p{'s} extensive menu\footnote{The word \emph{menu} (literally an information artifact) can be read as standing metonymically for the dishes available for order, to be created and served by the restaurant. Semantically, the restaurant frame defines various specific actions\slash relationships involving the menu and items on it. \psst{Gestalt} is highly general and therefore an appropriate selection when none of these particular actions or relationships are in focus.} [cf.~\cref{ex:with-menu}]
    \ex a new way \p{of} thinking
    \ex\label{ex:amountGestalt} the amount \p{of} time allowed [but see \cref{ex:QuantityGestalt}]
    \ex the food/service \p{at} this restaurant (\rf{Gestalt}{Locus})
\end{exe}
By extension, a possessor of an item in relation to an attribute of that item
is \psst{Gestalt}:
\begin{exe}
  \ex\label{ex:odometer} \p{my} odometer number (= my car's odometer's number) [contrast \cref{ex:windshield}]
\end{exe}
The property may be a fact or detail conventionally associated with someone in life for which there is no more specific role:
\begin{exe}
    \ex \p{my} address; \p{my} price range; \p{my} home team; \p{my} career; \p{my} record of accomplishments
\end{exe}

\paragraph{Containers.}
The construal \rf{Locus}{Gestalt} is used for a container marked by the adposition:
\begin{exe}
\ex the room\p{'s} 2 beds (\rf{Locus}{Gestalt})
\end{exe}

\paragraph{Discourse-associated item.}
A referent temporarily associated with another referent in the discourse 
and used to help identify it: 
\begin{exe}
  \ex Sam\p{'s} dog (= the dog that Sam mentioned seeing earlier in the conversation)
\end{exe}

\paragraph{Other possessive constructions.}
\psst{Gestalt} is the construal for many uses of possessive syntax
where the semantic criteria for \psst{Possessor} are not met. 
For instance, s-genitive marking of participant roles (\psst{Agent}, \psst{Experiencer}, 
etc.)\ are analyzed with \psst{Gestalt} as the function. 
Moreover, the s-genitive construction, 
unlike \p{of}, is never analyzed with \psst{Whole} as the function, 
so \rf{Whole}{Gestalt} is used. 
See \cref{sec:genitives} for discussion of possessive constructions.

\hierCdef{Possessor}

\shortdef{Animate party that has a piece of \textbf{property} (something potentially with monetary value: the \psst{Possession}) on a permanent or temporary basis. The \psst{Possession} must be \emph{alienable}, i.e.~not a part or attribute of the \psst{Possessor}.} 

Prototypically expressed with the \emph{s-genitive} (\cref{sec:genitives}: \p{'s} and possessive pronouns), 
and \p{of} (the \emph{of-genitive}):

\begin{exe}
{\setlength\multicolsep{0pt}%
\begin{multicols}{2}
  \ex\begin{xlist}
  \ex the house \p{of} the Smith family
  \ex\label{ex:corgis} the corgis \p{of} Queen Elizabeth
  \sn the Smith family\p{'s} house
  \sn Queen Elizabeth\p{'s} corgis
\end{xlist}
\end{multicols}}
\end{exe}
\psst{Possessor} is not limited to cases of \emph{ownership}, but also includes temporary forms of possession, 
such when something is on loan to or under the control of the possessor. 
The \psst{Possessor} may be \textit{borrowing}, \textit{renting}, \textit{wearing}, or \textit{holding} the property:
\begin{exe}
  \ex John\p{'s} hotel room [the room John is staying in as a guest]
  \ex Mary\p{'s} delivery truck [the company truck that Mary drives as an employee]
\end{exe}

There may be an implicit piece of property of which the stated item is a part:
\begin{exe}
  \ex\label{ex:windshield} \p*{My}{my} windshield (= the windshield of my vehicle) is foggy. [contrast \cref{ex:odometer}]
\end{exe}

A wearer of attire may be construed in multiple ways:
\begin{exe}
{\setlength\multicolsep{0pt}%
\begin{multicols}{2}
  \ex the cloak \p{of} He-Who-Must-Not-Be-Named
  \sn He-Who-Must-Not-Be-Named\p{'s} cloak
\end{multicols}}
\ex the cloak \p{on} He-Who-Must-Not-Be-Named (\rf{Possessor}{Locus})
\end{exe}

\textbf{Pets}, by default, are treated as property rather than family members \cref{ex:corgis}.

\paragraph{Abstract possession.} 
For \psst{Possessor}\slash \psst{Possession} to apply to an abstract piece of property, the property must be a commodity in the financial/commercial domain, or information stored externally to the \psst{Possessor} in physical or electronic media.
\begin{exe}
  \ex Commodity: \begin{xlist}
    \ex \p{my} shares of stock
    \ex \p{my} insurance (= insurance that I ``own'')
    \end{xlist}
  \ex Stored information: \begin{xlist}
    \ex \p{my} computer file
    \ex \p{my} website
    \ex the campaign\p{'s} email list
    \ex \p{my} contract
    \end{xlist}
\end{exe}
This excludes other abstract notions that can be metaphorically possessed or transferred:
\begin{exe}
  \ex\begin{xlist}
    \ex the president\p{'s} power (\psst{Gestalt})
    \ex \p{my} memories of childhood (\rf{Experiencer}{Gestalt})
    \end{xlist}
\end{exe}

\paragraph{Creation or transfer of an item.}
\psst{Originator} should be preferred as the scene role wherever 
it is clear that the party in question created the item:
\begin{exe}
  \ex Rodin\p{'s} sculptures (\rf{Originator}{Gestalt}) [\cref{ex:rodin}]
  \ex the restaurant\p{'s} food (\rf{Originator}{Gestalt}) [\cref{ex:restFood}]
\end{exe}
In cases of \emph{explicit} transfer, the initial possessor of something 
is labeled \psst{Originator}, and the final possessor of something transferred is the \psst{Recipient}.
However, when there is a possessed item whose transfer is merely assumed from context or world knowledge 
(and the party in question is not the creator), default to \psst{Possessor}:
\begin{exe}
  \ex The \choices{shopkeeper\p{'s}\\store\p{'s}} merchandise is fantastic. [item to be sold]
  \ex Waiter, is \p{my} food ready? [item identified or ordered for purchase]
  \ex Use \p{my} money (the money that I gave you) wisely! [item that was transferred]
\end{exe}
Possessed nouns like \emph{gift} and \emph{contribution} that refer to an entity
but lexically imply a previous transfer event should be \psst{Possessor}
unless another argument of the noun disambiguates \psst{Originator} vs.~\psst{Recipient}
by process of elimination:
\begin{exe}
  \ex\begin{xlist}
    \ex That was \p{my} gift. (\psst{Possessor})
    \ex That was \p{my} gift to$_{\text{\rf{Recipient}{Goal}}}$ John. (\rf{Originator}{Gestalt})
    \ex That was \p{my} gift from$_{\text{\rf{Originator}{Source}}}$ Mary. (\rf{Recipient}{Gestalt})
  \end{xlist}
\end{exe}

\paragraph{Communication.} 
Communicative acts are treated as transfer of information, and thus \psst{Originator} 
and \psst{Recipient} apply to the communicator and addressee, respectively.

See also \psst{Ancillary}, \psst{Beneficiary}, \psst{Org}, \psst{OrgMember}, and \fullref{sec:genitives}.

\hierCdef{Whole}

\shortdef{Something described with respect to its part, portion, subevent, subset, 
or set element. See \psst{PartPortion}.}

\begin{exe}
  
  \ex {\setlength\multicolsep{0pt}%
  \begin{multicols}{2} 
   \begin{xlist}
    \sn \psst{Whole}
    \ex	the new engine \p{of} the car
    \ex	the flaxen hair \p{of} the girl
    \ex the body \p{of} Lord Voldemort
    \ex\label{ex:layers}	the 3 layers \p{of} the cake
    \ex\label{ex:prongs}	the 3 prongs \p{of} the strategy
    \ex the tastiest bit \p{of} the cake
    \ex the southern tip \p{of} the island
    \ex the interior \p{of} the shopping bag
    \ex the end \p{of} the journey
    \ex the 14~episodes \p{of} a TV series
    
    \sn \rf{Whole}{Gestalt}
    \sn the car\p{'s} new engine
    \sn the girl\p{'s} flaxen hair
    \sn Lord Voldemort\p{'s} body
    \sn the cake\p{'s} 3 layers
    \sn the strategy\p{'s} 3 prongs
    \sn the cake\p{'s} tastiest bit
    \sn the island\p{'s} southern tip
    \sn the shopping bag\p{'s} interior
    \sn the journey\p{'s} end
    \sn a TV series\p{'s} 14~episodes
  \end{xlist}
  \end{multicols}}
  \ex	the south \p{of} France
  \ex\label{ex:rest} The \choices{remainder\\rest} \p{of} the cake 
  \ex \rf{Whole}{Locus}: \begin{xlist}
    \ex	the 14~episodes \p{in} a TV series
    \ex	the new engine \p{in} the car
    \ex the escape key \p{on} the keyboard
    \ex the flaxen hair \p{on} the girl
  \end{xlist}
  \ex\label{ex:in-pile-adnominal}	the clothes \p{in} that pile are dirty (\rf{Whole}{Locus}) 
  \ex There are several options to choose \p{from}. (\rf{Whole}{Source})
  \ex\label{ex:sets} Sets and ratios:
    \begin{xlist}
      \ex This is one \p{of} the \choices{worst\\better} retaurants in town. (\psst{Whole})
      \ex 2 \p{in} 10 American children are redheads. (\rf{Whole}{Locus})
      \ex 2 \p{out\_of} 10 American children are redheads. (\rf{Whole}{Source}) 
      \ex \p*{Out\_of}{out\_of} the 10 children in the class, only Mary is a redhead. (\rf{Whole}{Source})
      \ex\label{ex:amongSet} \p*{Among}{among} the 10 children in the class, only Mary is a redhead. (\psst{Whole})
    \end{xlist}
\end{exe}

If the governor narrows the reference to a certain amount of the \psst{Whole}, 
the construal \rf{QuantityItem}{Whole} is used---see \cref{ex:QuantityWhole}.
Note that this only applies if the governor is a measure term; 
it does not apply to distinctive parts like ``layers'' \cref{ex:layers} 
and ``prongs'' \cref{ex:prongs}, even if a count is specified.

Used to construe geographic and temporal ``containers'':
\begin{exe}
  \ex	Famous castles \p{of} the valley (\rf{Locus}{Whole})
  \ex \begin{xlist}
    \ex the \choices{15th\\Ides} \p{of} March (\rf{Time}{Whole})
    \ex March \p{of} 44~BC (\rf{Time}{Whole})
  \end{xlist}
\end{exe}

The prepositions \p{between} and \p{among} can impose \psst{Whole} construals 
by combining two or more items in the object NP (contrast with \cref{ex:amongSet}):
\begin{exe}
  \ex\label{ex:betweenParties}  The negotiations \choices{\p{between}\\\p{among}} the parties went well. (\rf{Agent}{Whole})
  \exp{ex:betweenParties} The negotiations \p{by} the parties went well. (\psst{Agent})
\end{exe}

\begin{history}
  In v1, \sst{Superset} was distinguished as a subtype of \psst{Whole} 
  for examples such as \cref{ex:sets}, but the distinction was dropped for v2 
  (as was \sst{Elements}: see \psst{PartPortion}).
\end{history}

\hierCdef{Org}

\shortdef{An organization\slash institution when mentioned in relation to an \psst{OrgMember},
i.e., an individual who has a stable affiliation with that organization,
such as membership or a business relationship.}

We define \textbf{organization} as an established group of people with some social or societal purpose/function; this includes families, performing arts groups, schools, businesses, and governmental units:
\begin{exe}
  \ex\begin{xlist}
    \ex employees \p{of} the company (\rf{Org}{Gestalt})
    \ex I work \p{for} the United Nations (\rf{Org}{Beneficiary})
    \ex actors \p{in} the troupe (\rf{Org}{Locus})
  \end{xlist}
\end{exe}

Stative scenes include someone belonging to, being an employee of,
or being in a business relationship with an \psst{Org};
dynamic scenes include someone joining or leaving\slash beginning or ending
an engagement with an \psst{Org}.
\begin{exe}
  \ex I was hired \p{by} Microsoft. (\rf{Org}{Agent})
\end{exe}

Relations between organizations such as partnerships, subsidiaries,
and super-organization\slash sub-organization are \emph{not} covered by \psst{Org}, nor are general mentions of organizations:

\begin{exe}
  \ex My business was destroyed \p{by} Microsoft. (\psst{Agent})
  \ex Microsoft\p{'s} cloud services division (\psst{Whole})
\end{exe}

\psst{Org} lacks any prototypical adpositions in English, but participates in numerous construals:

\begin{exe}
  \ex \rf{Org}{Gestalt} with the institution as possessor:
  {\setlength\multicolsep{0pt}%
  \begin{multicols}{2}
    \begin{xlist}
      \ex the chairman \p{of} the board
      \ex the president \p{of} France
      \ex \choices{employees\\customers} \p{of} Grunnings

      \sn the board\p{'s} chairman
      \sn France\p{'s} president
      \sn Grunnings\p{'s} \choices{employees\\customers}
    \end{xlist}
  \end{multicols}}
  \ex\begin{xlist}
    \ex Mr. Dursley works \p{for} Grunnings. (\rf{Org}{Beneficiary})
    \ex Mr. Dursley works \p{at} Grunnings. (\rf{Org}{Locus})
    \ex Mr. Dursley is \p{from} Grunnings. (\rf{Org}{Source})
    \ex Mr. Dursley is \p{with} Grunnings. (\rf{Org}{Ancillary})
    \ex Mr. Dursley is employed \p{by} Grunnings. (\rf{Org}{Agent}) 
  \end{xlist}
  \ex\rf{Org}{Ancillary}:\begin{xlist}
    \ex I always do business \p{with} this company.
    \ex I bank \p{with} TSB.
    \ex my phone service \p{with} Verizon
  \end{xlist}
  \ex For my Honda I always got replacement parts \p{through} the dealership. [intermediary business] (\rf{Org}{Instrument})
  \ex I serve \p{on} the committee. (\rf{Org}{Locus})
\end{exe}

A family counts as an institution
construed as a \psst{Whole} (set of its members)
or as a \psst{Locus}:
\begin{exe}
  \ex I am the baby \p{of} the family. (\rf{Org}{Whole})
  \ex people \p{in} my family (\rf{Org}{Locus})
\end{exe}

For a relation between a unit and a larger institution,
use \psst{Whole}:
\begin{exe}
  \ex the Principals Committee \p{of} the National Security Council (\psst{Whole})
\end{exe}

See also: \psst{Stuff}

\begin{history}
  In v1, \sst{ProfessionalAspect} marked relations between an employee
  and an employer, supervisor, or coworker.
  In v2.0, this was revised to \sst{OrgRole}, for relations between an individual and an organization,
  and a supertype \psst{SocialRel} covering all established social relationships.
  This was further refined in v2.5, when \sst{OrgRole} was split into two supersenses,
  \psst{Org} and \psst{OrgMember}, so the directionality of the (asymmetric) relation would be specified.
  Instead of being under \psst{SocialRel}, these are under \psst{Gestalt} and \psst{Characteristic}, respectively.

  The separation of individual--organization and individual--individual relations
  follows the precedent of the Abstract Meaning Representation \citep[AMR;][]{amr,amr-guidelines},
  where \texttt{have-org-role-91} captures relations between
  an individual and an institution (such as an organization or family),
  and \texttt{have-rel-role-91} is used for relations between two individuals.
\end{history}

\hierCdef{QuantityItem}

\shortdef{Something measured by a quantity denoted by the governor (the \psst{QuantityValue}).}

The governor may be a precise or vague count/measurement.
This includes nouns like ``lack'', ``dearth'', ``shortage'', ``excess'', or ``surplus''
(meaning a too-small or too-large amount).

Question test: the governor answers ``How much/many of (object)?''

The main preposition is \p{of}.

\begin{itemize}
\item Simple \psst{QuantityItem}:
\begin{exe}
  \ex\label{ex:bottleQuantity}	Pour me a bottle('s worth) \p{of} beer. [but see \cref{ex:bottleStuff}]
  \ex	I have 2 years \p{of} training.
  \ex	\begin{xlist}
    \ex I ate \choices{6 ounces\\a piece} \p{of} cake.
    \ex	An ounce \p{of} compassion
  \end{xlist}
  \ex	There's a dearth \p{of} cake in the house.
  \ex	This cake has thousands \p{of} sprinkles.
  \ex They number in the tens \p{of} thousands.
  \ex	\begin{xlist}
    \ex\label{ex:anumber} I have a \choices{number\\handful} \p{of} students.
    \ex	I have a lot \p{of} students.
    \ex	We did a lot \p{of} traveling.
    \ex	There is a lot \p{of} wet sand on the beach.
  \end{xlist}
  \ex	A pair \p{of} shoes
\end{exe}

\item If the measure includes a word like ``amount'', ``quantity'', or ``number'',\footnote{Excluding
the expression ``a number'' meaning `several', as in \cref{ex:anumber}.}
the construal \rf{QuantityItem}{Gestalt} is used
(because the amount of something can be viewed as an attribute):
\begin{exe}
  \ex\label{ex:QuantityGestalt} \rf{QuantityItem}{Gestalt}:
  \begin{xlist}
    \ex	A generous amount \p{of} time
    \ex A large number \p{of} students
  \end{xlist}
\end{exe}
But if ``amount'', ``quantity'', etc. is used without a measure as its modifier,
it is simply \psst{Gestalt}: see \cref{ex:amountGestalt}.

\item If the governor is a \textbf{collective noun} not denoting an organization,
the construal \rf{QuantityItem}{Stuff} is used
(note that a ``consisting of'' paraphrase is possible):
\begin{exe}
  \ex\label{ex:QuantityStuff} \rf{QuantityItem}{Stuff}:
  \begin{xlist}
    \ex Can you outrun a herd \p{of} wildebeest?
    \ex Put 3 bales \p{of} hay on the truck.
    \ex	\choices{A group\\2 groups\\A throng} \p{of} vacationers just arrived.
  \end{xlist}
\end{exe}
For organizational collectives, see \psst{OrgMember}.

\item Otherwise, if the object refers to \textbf{a specific item or set},
and the quantity measures a portion of that item
(whether a quantifier, absolute measure, or fractional measure),
the construal \rf{QuantityItem}{Whole} is used:
\begin{exe}
  \ex\label{ex:QuantityWhole} \rf{QuantityItem}{Whole}:
  \begin{xlist}
    \ex	I ate 6 ounces \p{of} the cake in the refrigerator.
    \ex	I ate \choices{half\\50\%} \p{of} the cake.
    \ex	\choices{All/many/lots/a lot/\\some/few/both/none} \p{of} the town's residents
    are students.
    \ex	I have seen all \p{of} the city. (= the whole city)
    \ex	A lot \p{of} the sand on the beach is wet.
    \ex	2 \p{of} the children are redheads.
    \ex 2 \p{of} the 10 children in the class are redheads.
  \end{xlist}
\end{exe}
However, simple \psst{Whole} is used if the portion is specified as
``the rest'', ``the remainder'', etc., as in \cref{ex:rest}. 
\end{itemize}

\begin{history}
  Prior to v2.5, this was called \sst{Quantity}, which was inconsistent with
  other supersense names as the quantity itself is denoted by the governor
  rather than the object of the preposition.
\end{history}

\hierBdef{Characteristic}

\shortdef{Generalized notion of a part, 
feature\slash property\slash attribute name or value, 
qualitative state\slash condition, possession, 
or the contents or composition of something, 
understood with respect to that thing (the \psst{Gestalt}).}

Labels \psst{Possession}, \psst{PartPortion} and its subtype \psst{Stuff},
\psst{OrgMember}, and \psst{QuantityValue} and its subtype \psst{Approximator}
are defined for some important subclasses.

\psst{Characteristic} applies directly to:
\begin{itemize}
\item	A property value: 
\begin{exe} 
  \ex Adnominal: \rf{Characteristic}{Identity}\begin{xlist}
    \ex a car \p{of} high quality
    \ex a man \p{of} honor
    \ex a business \p{of} that sort [contrast with \psst{Species}, \cref{sec:Species}]
  \end{xlist}
  \ex Secondary predicate adjective: \rf{Characteristic}{Identity}\begin{xlist}
    \ex She described him \p{as} sad.
    \ex He strikes me \p{as} sad.
  \end{xlist}
\end{exe}
\item	Role of a complex framal \psst{Gestalt} that has no obvious decomposition into parts: 
\begin{exe}\ex \begin{xlist}
  \ex\label{ex:with-menu} the restaurant \p{with} \choices{a convenient location\\an extensive menu} [cf.~\cref{ex:poss-menu}]
  \ex a party \p{with} great music
\end{xlist}\end{exe}
\item	That which is located in a container denoted by the governor: 
\begin{exe}
  \ex a room \p{with} 2 beds [beds are among the things in the room]
  \ex\rf{Characteristic}{Stuff} where the object of the preposition is construed as describing the contents in their entirety:\begin{xlist}
    \ex a shelf \p{of} rare books
    \ex a cardboard box \p{of} snacks
  \end{xlist}
\end{exe}
\item With a transitive verb like \emph{search}, \emph{examine}, or \emph{test}, 
the attribute of the \psst{Theme} that is being examined:
\begin{exe}
  \ex He examined the vase \p{for} damage.
  \ex\label{ex:search-obj-for} He searched the room \p{for} his laser pistol. [contrast intransitive \psst{Theme}, \cref{ex:search-for}]
  \ex He was tested \p{for} low blood sugar.
\end{exe}
\item The scale or dimension by which items are compared:
\begin{exe}
  \ex The children are \choices{sorted\\screened} \p{by} height
  \ex\begin{xlist}
    \ex She exceeds him \p{in} height
    \ex There is no difference \p{in} height
  \end{xlist}
\end{exe}
\item The \textbf{form or shape} that an entity takes, or in which elements are arranged. This includes language of communication within an information source, and unit of measure (only the unit, not a full measurement) in relation to the attribute measured:
  \begin{exe}
    \ex\label{ex:in-shape} \rf{Characteristic}{Locus}:\begin{xlist}
      \ex The ribbon is (tied) \p{in} a bow.
      \ex The sand is \p{in} a pyramid shape.
      \ex\label{ex:inarow} I skipped lunch for$_{\psst{Frequency}}$ \choices{three days\\the third day} \p{in} a row. [see \cref{ex:freqrow}]
      \ex\label{ex:bookinfrench} The book is \p{in} French. [contrast \cref{ex:wroteinfrench}]
      \ex music \p{in} C major
      \ex the desk's height (measured) \p{in} inches
    \end{xlist}
  \end{exe}
\item An adverbial \textbf{depictive} characterizing a participant of an event:
  \begin{exe}
    \ex She entered the room \choices{\p{in} a stupor\\drunk}. (= she was in a stupor when she entered) (\rf{Characteristic}{Locus}) [repeated at \psst{Manner} for contrast]
    \end{exe}
\item	Anything that is borderline between the \psst{Possession} and \mbox{\psst{PartPortion}} subcategories.
  \item The \textbf{state or condition} that an entity is in. 
  The PP or intransitive preposition is used (especially predicatively) 
  to describe a qualitative state or condition of an entity 
  that is not simply a relation of location, time, possession, quantity, causation, etc.\ 
  between governor and object.
  For example:
  
  \begin{itemize}
    \item With the noun \pex{state}, \pex{condition}, etc.:
    
    \begin{exe}
      \ex\rf{Characteristic}{Locus}:\begin{xlist}
        \ex The chairs are \p{in} excellent shape.
        \ex I'm \p{in} no condition to go outside.
      \end{xlist}
    \end{exe}

    \item Bodily/medical conditions presented as applying to the governor:
    
    \begin{exe}
      \ex John is \choices{\p{on} his back\\\p{on} antibiotics\\\p{on} the ventilator\\\p{in} pain\\\p{in} a coma}. (\rf{Characteristic}{Locus})
    \end{exe}
    
    \item Miscellaneous qualitative senses of specific prepositions used statively:
    
    \begin{exe}
      \ex John is \choices{\p{for}\\\p{against}} the war. [opinion] (\rf{Characteristic}{Beneficiary})
      \ex John is \p{into} sports. [hobbies/interests] (\rf{Characteristic}{Goal})
    \end{exe}

    \item Idiomatic PPs expressing states, for example:\footnote{Often the object of the preposition 
    is determinerless (\pex{\p{in} business}) \citep{baldwin-06} 
    or has a fixed determiner (\pex{\p{in} a hurry}).}
    \begin{exe}
    \ex \pex{\p{on} fire} (contrast \pex{\p{in} the fire}), 
      \pex{\p{on} time} (contrast \pex{\p{at} the time}), 
      \pex{\p{in} trouble}, \pex{\p{in} love}, \pex{\p{in} tune}, \pex{\p{in} a hurry}, 
      \pex{\p{at} odds}, \pex{\p{out\_of} business}, \pex{\p{out\_of} control} (\rf{Characteristic}{Locus})
    \end{exe}
    
    \item Intransitive prepositions expressing a qualitative state (not location, time, etc.):
    \begin{exe}
      \ex\rf{Characteristic}{Locus}:\begin{xlist}
        \ex The lights are \choices{\p{on}\\\p{off}\\\p{out}}. [also at \psst{Source} for contrast]
        \ex Political TV shows are \p{in}. [in fashion]
      \end{xlist}
    \end{exe}
    Contrast intransitive predicative prepositions describing an \emph{event}:
    \begin{exe}
      \ex The party tomorrow is \p{on}. (\rf{Temporal}{Locus}) [see: \psst{Temporal}]
    \end{exe}
  \end{itemize}
  
  A few observations about these state PPs are in order.
  
  \begin{enumerate}
  \item In a reversal of the usual asymmetry between governor and adpositional object, 
  semantically, the PP defines the kind of scene that the governor participates in. 
  To an extent, this may be true of all predicative PPs, but the state PPs are often such 
  that the object of the preposition is neither an event nor a referential entity. 
  I.e., \pex{John is \p{in} a hurry} does not exactly express a relation 
  between the entities \pex{John} and \pex{a hurry}; rather, it expresses something 
  qualitative about the entity \pex{John}'s condition.
  
  \item The most idiomatic of the state PPs seem to resist questions of the form 
  \emph{What?}+NP-supercategory with a stranded preposition:
  \begin{exe}
    \ex More productive prepositional usages:\begin{xlist}
      \ex The party is \p{in} January.~$\rightarrow$ What month is the party \p{in}? [Or: When is the party?] (\psst{Time})
      \ex John is \p{on} aspirin.~$\rightarrow$ What medication is John \p{on}?\footnote{Or, colloquially, with a suspected mind-altering substance: \pex{What is John \emph{\p{on}}?!}} (\rf{Characteristic}{Locus})
    \end{xlist}
    \ex\label{ex:idiomPP} Less productive/more idiomatic preposition + NP combinations:\begin{xlist}
      \ex John is \p{in} \choices{a hurry\\a coma}.~$\nrightarrow$ What \_ is John \p{in}?\footnote{\pex{What condition/state is John \p{in}?} does work, but is quite vague.}  (\rf{Characteristic}{Locus})
      \ex John is \p{on} fire.~$\nrightarrow$ What \_ is John \p{on}? (\rf{Characteristic}{Locus})
    \end{xlist}
  \end{exe}

  \item Typically these states are binary: something is either \pex{\p{on} fire}\slash \pex{\p{on} time}, or not. 
  For some, the negation may be expressed by substituting a contrasting preposition: 
  an orchestra that is not \pex{\p{in} tune} is \pex{\p{out\_of} tune}.
  \end{enumerate}

  \paragraph{State PPs with complements.}
  The \rf{Characteristic}{Locus} construal is also used when there is effectively a preposition+NP+preposition  
  combination that links two arguments:
  \begin{exe}
    \ex\rf{Characteristic}{Locus}:\begin{xlist}
      \ex\label{ex:InLove} John is \p{in} love (with$_{\text{\rf{Stimulus}{Topic}}}$ Mary). [cf.~\cref{ex:InLoveWith}]
      \ex That is \p{at} odds with$_{\text{\rf{ComparisonRef}{Topic}}}$ our agreement.
    \end{xlist}
  \end{exe}

  \paragraph{Change-of-state PPs.} 
  Note that \psst{Characteristic} does not apply to an initial or result state, where \psst{Source} and \psst{Goal} are the respective scene roles 
  (collapsing the usual state/location distinction):
  \begin{exe}
    \ex John came \p{out\_of} a coma. (\psst{Source})
    \ex John slipped \p{into} a coma. (\psst{Goal})
    \ex \rf{Goal}{Locus}: \begin{xlist} 
        \ex The drugs put John \p{in} a coma.
        \ex They chopped the wood \p{in} pieces.
        \ex The devices may become \p{out\_of} sync.
    \end{xlist}
  \end{exe}
\end{itemize}

For some usages but not all, one of ``\psst{Gestalt} \{HAS, CONTAINS\} \psst{Characteristic}'' is entailed. 
This does not help to distinguish subtypes.

\paragraph{Versus \psst{Circumstance}.}
State PPs like \pex{\p{at} odds} and \pex{\p{on} medication}, which receive the 
construal \rf{Characteristic}{Locus}, are similar to situating events like 
\pex{\p{at} the party} and \pex{\p{on} vacation}, which are analyzed as 
\rf{Circumstance}{Locus}. What matters for the scene role is whether the object 
of the preposition is an event or not.

\paragraph{Versus \psst{Manner}.}
If a property pertains to an entity---whether that entity is the syntactic governor or not---then \psst{Characteristic}. \psst{Manner} is limited to descriptors of events.

\begin{history}
  The v1 label \sst{Attribute} was intended to apply to features of something, 
  but was vaguely defined. With the overhaul of the \psst{Configuration} 
  subhierarchy, \sst{Attribute} has primarily been replaced by 
  \psst{Characteristic} and its subtypes and \psst{Identity}.
\end{history}

\hierCdef{Possession}


\shortdef{Piece of \textbf{property} (something potentially with monetary value) that an animate party (the \psst{Possessor}) has on a permanent or temporary basis, or that is transferred between parties. The \psst{Possession} must be \emph{alienable}, i.e.~not a part or attribute of the \psst{Possessor}.} 

The \psst{Possessor} may \textit{own} or may be \textit{borrowing}, \textit{renting}, \textit{wearing}, or \textit{holding} the property.

Prototypical prepositions are \p{with} and \p{without}:
\begin{exe}
  \ex	People \choices{\p{with}\\\p{without}} money
\end{exe}
There is also a (negated) possession sense of \p{out}/\p{out\_of}:
\begin{exe}
  \ex\begin{xlist} 
    \ex We are \p{out\_of} toilet paper.
    \ex Toilet paper? We are \p{out}.
  \end{xlist}
\end{exe}

Attire may be construed in multiple ways:
\begin{exe}
  \ex the kid \p{with} \choices{a vest\\makeup} (on)
  \ex the kid \p{in} a vest (\rf{Possession}{Locus})
\end{exe}

Immediate concrete possession uses an \psst{Ancillary} construal:
\begin{exe}
  \ex Hagrid exited the shop \p{with} (= carrying) a snowy owl. (\rf{Possession}{Ancillary})
\end{exe}

\paragraph{Transfer, goods, and services.}
In a commercial scene, goods, services, and money are distinguished. 
\psst{Possession} is used as the scene role 
for goods for sale. 
\psst{Possession} also applies to a piece of property transferred between parties, lost, acquired, or carried, even if no money changes hands.
\psst{Theme} is the scene role for commercial services. 
\psst{Cost} applies to the money asked, paid, or owed.

The construal \rf{Possession}{Theme} is used for goods marked by \p{on}, \p{for}, etc., 
whereas \p{with} can be simple \psst{Possession}:
\begin{exe}
  \ex Simple change of possession and transfer: \begin{xlist}
    \ex I bestowed the winner \p{with} \choices{a bicycle\\\$100}. (\psst{Possession}) [repeated at  \psst{Cost}]
    \ex They robbed her \p{of} her life savings. (\rf{Possession}{Theme})
      \end{xlist}
  \ex\label{ex:goods} Goods: \begin{xlist}
    \ex They spent \$500 \p{on} the bicycle. (\rf{Possession}{Theme}) [also at \psst{Theme}]
    \ex They charged/asked/paid/owed \$500 \p{for} the bicycle. (\rf{Possession}{Theme})
    \ex \$500 \p{for} the bicycle was excessive. (\rf{Possession}{Theme})
  \end{xlist}
\end{exe}
Contrast \cref{ex:services} under \psst{Theme}.

Paraphrase test: ``\psst{Possessor} POSSESSES \psst{Possession}'', 
``\psst{Possessor} is IN POSSESSION OF \psst{Possession}'', or 
``\psst{Possessor} HAS ON \psst{Possession}'' for stative possession; 
``\psst{Recipient} ACQUIRES \psst{Possession}'' or ``\psst{Originator} LOSES POSSESSION OF \psst{Possession}'' for change of possession. 
``IN POSSESSION OF'' is especially appropriate for immediate concrete stative possession.

See further discussion at \psst{Possessor}.

\hierCdef{PartPortion}

\shortdef{A part, portion, subevent, subset, or set element (e.g., an example or exception) 
of some \psst{Whole}.}

Anything directly labeled with \psst{PartPortion} 
is understood to be \textbf{incomplete} relative to the \psst{Whole}.
This includes body parts and partial food ingredients.

Prototypical prepositions include \p{with}, \p{without};
\p{such\_as}, \p{like} for exemplification; 
and \p{but}, \p{except}, \p{except\_for} for exceptions:
\begin{exe}
  \ex \begin{xlist}
    \ex	a car \p{with} a new engine
    \ex	a strategy \p{with} 3 prongs
    \ex	the girl \p{with} flaxen hair
    \ex	a man \p{with} a wooden leg named Smith
    \ex	a valley \p{with} a castle
    \ex	a quintet \p{with} 2 cellos
    \ex	a performance \p{with} a guitar solo
    \ex	a cake \p{with} 3 layers
    \ex	a sandwich \p{with} wheat bread
    \ex	soup \p{with} carrots (in it)
    \ex	a chicken sandwich \p{with} ketchup (on it)
  \end{xlist}
  \ex	Bread \p{without} gluten
\end{exe}
Some can be paraphrased with INCLUDES, but this is not determinative.

\paragraph{Elements and Exceptions.} 
\psst{PartPortion} is used for adpositions marking a member or non-member of a set:
\begin{exe}
  \ex\label{ex:suchAs}	strategies \p{such\_as} divide-and-conquer
  \ex Everyone \choices{\p{except}/\p{but}} Bob plays trombone.
  \ex\label{ex:no-openings-except} There are no openings \p{except} on$_{\psst{Time}}$ Friday.\footnote{\Citet[pp.~641--643]{cgel} discuss how \p{except} and other prepositions of inclusion/exclusion may occur with \emph{matrix-licensed complements}. \Cref{ex:no-openings-except} is an example: it implies that there are no openings in general, but there are openings on Friday---that is, \pex{on Friday} is a time adjunct licensed by the clause rather than \p{except} itself.}
\end{exe}
Set-membership can be construed as comparison:
\begin{exe}
  \ex strategies \p{like} divide-and-conquer [same reading as \cref{ex:suchAs}]\\ (\rf{PartPortion}{ComparisonRef})
\end{exe}
The set may be an organizational collective:
\begin{exe}
  \ex	A piano quintet is a chamber group \p{with} a piano (in it)\\ (\rf{OrgMember}{PartPortion}) [repeated at \psst{OrgMember}]
\end{exe}

\paragraph{Diverse Examples.}
In describing a set or whole, a sort of scanning with \p{from}\dots\p{to} can be used indicate diversity or coverage of 
the items/parts:
\begin{exe}
  \ex\label{ex:diverserange} Everyone \p{from}$_{\text{\rf{PartPortion}{Source}}}$ the peasants 
  \p{to}$_{\text{\rf{PartPortion}{Goal}}}$ the lord and lady gathered for the feast.
\end{exe}

\paragraph{\pex{Start \p{with}}\lex*[start with]{start \p{with}}, \pex{end \p{with}}, etc.}
Along similar lines as \cref{ex:diverserange}, \p{with} can be used with 
an aspectual verb to indicate an item in a sequence: 
\pex{start \p{with}}, \pex{continue \p{with}}, \pex{end \p{with}}, and similar.
Here the scene role \psst{PartPortion} applies 
(though note that it is a part with respect to another argument of the verb, 
not the verb itself):
\begin{exe}
  \ex\rf{PartPortion}{Means}:\begin{xlist} 
    \ex My teacher started the lesson \p{with} a quiz.
    \ex The lesson started \p{with} a quiz.
  \end{xlist}
  \ex The meal started \p{with} an appetizer. (\rf{PartPortion}{Instrument})
\end{exe}

\begin{history}
  In v1, instead of this category, there were separate categories 
  \sst{Elements} for set members, \sst{Comparison/Contrast} for exemplification,
  and \sst{Attribute} for other parts (grouped with properties, which are now \psst{Gestalt}).
  (\sst{Superset} was removed along with \sst{Elements}: see \psst{Whole}.)
\end{history}

\hierDdef{Stuff}

\shortdef{The members comprising a group/ensemble, 
or the material comprising some unit of substance. 
\psst{Stuff} is distinguished from other instances of \psst{PartPortion}
in fully covering (or ``summarizing'') the aggregate whole.}

Paraphrase test: ``\psst{Whole} CONSISTS OF \psst{Stuff}''

\begin{exe}
  \ex	\begin{xlist}    
    \ex	A clump \p{of} sand
    \ex	A piece \p{of} wood
    \ex	An evening \p{of} Brahms
    \ex	A meal \p{of} salmon
  \end{xlist}
  \ex	A salad \choices{\p{of}\\\p{with}} mixed greens
  \ex\label{ex:bottleStuff} This bottle is \p{of} beer (and that one is of wine). (\rf{Characteristic}{Stuff}) [but see \cref{ex:bottleQuantity}]
  \ex A group/throng \p{of} vacationers (\rf{QuantityItem}{Stuff}) [governor is collective noun not denoting an organization; more at \psst{QuantityItem}]
  \ex \rf{OrgMember}{Stuff}:
  	\begin{xlist}
      \ex An order \p{of} nuns [repeated at \psst{OrgMember}]
      \ex	A chamber group \choices{\p{of}\\\p{with}} 5 players [repeated at \psst{OrgMember}]
    \end{xlist}
\end{exe}

\psst{Stuff} has no specific counterpart under \psst{Whole}.

\hierCdef{OrgMember}

\shortdef{Individual(s) who are members of an organization, when mentioned in relation to the organization (\psst{Org}).}

\begin{exe}
\ex \rf{OrgMember}{Gestalt} with possessive marking on the individual:
  \begin{xlist}
    \ex \p{my} school/gym [that I attend]
    \ex \p{my} work [the place where I work]
    \ex \p{my} landscaping company [that I hired]
    \ex the family \p{of} Miss Zarves
    \ex Miss Zarves\p{'s} family
  \end{xlist}
\ex \rf{OrgMember}{Possessor} if the individual is understood to possess authority
  within or as a representative of the institution:
  \begin{xlist}
    \ex \p{my} small business [that I own or operate]
    \ex the president\p{'s} administration
  \end{xlist}
\ex\rf{OrgMember}{PartPortion} if the governor is an organizational collective noun
  and the object of the preposition denotes a subset of members:
    \begin{xlist}
      \ex	A piano quintet is a chamber group \p{with} a piano (in it) [repeated at \psst{PartPortion}]
    \end{xlist}
\ex\rf{OrgMember}{Stuff} if the governor is an organizational collective noun
and the object of the preposition describes the full membership:
  \begin{xlist}
    \ex An order \p{of} nuns [repeated at \psst{Stuff}]
    \ex A chamber group \choices{\p{of}\\\p{with}} 5 players [repeated at \psst{Stuff}]
  \end{xlist}
\end{exe}

\psst{OrgMember}s are generally people; sub-organizations within a larger organization are excluded.

Note that relations between persons, even in the context of an organization, fall under \psst{SocialRel} (e.g.~\pex{\p{my} boss}).

A member of an organization is only labeled \psst{OrgMember} when the relationship to the \psst{Org} is in focus: e.g., an activity done as part of a job with an organization will not trigger \psst{OrgMember}:

\begin{exe}
  \ex At the restaurant we were seated \p{by} the host. (\psst{Agent})
\end{exe}

\begin{history}
  See history note at \psst{Org}.
\end{history}

\hierCdef{QuantityValue}

\shortdef{The amount marked as the quantity of something (the \psst{QuantityItem}).}

Except for its subtype \psst{Approximator},
there are no prototypical adpositions for this category in English:
wherever quantity--quantified relations are marked by a preposition,
the preposition is on the quantified thing, and is thus labeled \psst{QuantityItem}.

\psst{QuantityValue} can, however, appear as the scene role with very specific predicates
like in \cref{ex:numbersIN}:

\begin{xexe}
  \ex\label{ex:numbersIN} The deaths numbered \p{in} the thousands. (\rf{QuantityValue}{Locus})
  \ex There were thousands \p{of} deaths. (\psst{QuantityItem})
\end{xexe}

\begin{history}
  Added in v2.5.
\end{history}

\hierDdef{Approximator}

\shortdef{An adposition that converts a point value (or pair of values)
into an approximate value or range,
without establishing a relation between two items/values to be compared.}

This includes:
\begin{itemize}
  \item \textit{Prepositions of approximation:} \p{about}, \p{around}, \p{in\_the\_vicinity\_of}
  \item \textit{Prepositions of scalar comparison:} \p{over}, \p{under}, \p{between}, \lex[at\_least]{\p{at}\_least}, \lex[at\_most]{\p{at}\_most}, \p{more\_than}, \p{less\_than}, \p{greater\_than}, and \p{fewer\_than}
\end{itemize}

For instance:
\begin{exe}
  \ex\label{ex:approxEggs}\begin{xlist}
    \ex\label{ex:aboutApproxEggs} We have \p{about} a dozen eggs left.
    \ex We have \p{in\_the\_vicinity\_of} a dozen eggs left.
    \ex We have \p{over} a dozen eggs left.
    \ex We have \p{between} 3 and 6 eggs left.
  \end{xlist}
  \ex\label{ex:approxWide} The lake is \p{around}/\lex[at\_least]{\p{at}\_least} a mile wide.
\end{exe}
\psst{Approximator} prepositions can also apply to expressions of spatial and temporal distance
(see \psst{Direction} and \psst{Interval}).
The syntactic analysis of these constructions is not obvious;
here the policy is simply to apply the label \psst{Approximator} while remaining
agnostic as to the precise syntax.%
\footnote{These constructions are
markedly different from most PPs; it is even questionable whether these usages
should count as prepositions. Without getting into the details here,
even if their syntactic status is in doubt,
we deem it practical to assign them with a semantic label in our inventory because they
overlap lexically with ``true'' prepositions.}

Note, however, that a simple relative comparison of an unknown value against a point on a scale
qualifies as \rf{ComparisonRef}{Locus} (see discussion at \psst{ComparisonRef}): 
\begin{exe}
  \ex expenses \p{under} \$100 (\rf{ComparisonRef}{Locus})
  \ex\label{ex:copulaCompRef} Your score is \choices{\p{above}\\\p{over}\\\lex[at\_least]{\p{at}\_least}} 100. (\rf{ComparisonRef}{Locus})
\end{exe}
\cref{ex:copulaCompRef} features a copular sentence with a preposition at the beginning of a predicate complement.
In cases like this, is arguably ambiguous as to whether the preposition acts as a modifier of the quantity,
which would suggest \psst{Approximator}, or establishes a relation between subject
and predicate, which would suggest \rf{ComparisonRef}{Locus}.
In general we prefer the latter analysis.

\begin{history}
  Prior to v2.5, this was under \sst{Quantity}, which in v2.5 was replaced
  with the dichotomy of \psst{QuantityValue} and \psst{QuantityItem}.
  \psst{Approximator} was moved under \psst{QuantityValue} as it applies to
  prepositions that modify point values to produce approximate or range values.
\end{history}

\hierBdef{Ensemble}

\shortdef{Entity that another entity is grouped with.}

\psst{Ensemble} labels a connective that relates two things (typically entities) to indicate that they form a general semantic grouping of some kind. In this grouping, they are on roughly equal footing---neither is a part, member, possession, location, or containing event for the other---though one may be presented as slightly more prominent in the discourse.

Ensemble relations in English are prototypically expressed with conjunctions like \w{and},
but may also be construed via \psst{Ancillary} adpositions:

\begin{exe}
  \ex\rf{Ensemble}{Ancillary}:\begin{xlist}
    \ex rice \choices{\p{with}\\\p{without}} beans (= served or mixed together)
    \ex A smile \p{with} a nod was his answer.
    \ex\label{ex:go-with-stative} Do these shoes go \p{with} this outfit? [cf.~\cref{ex:bring-with-stative} at \psst{Ancillary}] 
  \end{xlist}
\end{exe}

\paragraph{Versus \psst{Ancillary}.}
\psst{Ensemble} is used for a relation directly between entities,
whereas \psst{Ancillary} describes a relation of an entity to an event/situation.

\begin{history}
    Prior to v2.5, a single label, \sst{Accompanier} (under \psst{Configuration}),
    covered both entity--entity and event--entity relations.
    In v2.5, \sst{Accompanier} was split into \psst{Ancillary} and \psst{Ensemble}.

    The term ``ensemble'' is borrowed from \citet[pp.~214]{lindstromberg-10},
    though it is applied more narrowly here.
\end{history}

\hierBdef{ComparisonRef}

\shortdef{The reference point in an explicit comparison (or contrast), i.e.,
an expression indicating that something is
\textbf{similar/analogous to}, \textbf{different from}, \textbf{the same as},
or an \textbf{alternative to} something else.}

The marker of the ``something else'' (the ground in the figure–ground relationship) 
is given the label \psst{ComparisonRef}:
\begin{exe}
  \ex \begin{xlist}
    \ex She is taller \p{than} me.
    \ex She is taller \p{than} I am.
    \ex She is taller \p{than} she is wide.
    \ex She is better at math \p{than} at drawing.
    \ex The shirt is more gray \p{than} black.
  \end{xlist}
  \ex\label{ex:comparAs} \begin{xlist}
    \ex She is as tall \p{as} I am.
    \ex Your face is (as$_{\text{\rf{Characteristic}{Extent}}}$) red \p{as} a rose. (more on \p{as}-\p{as} comparatives: \cref{sec:as-as})
    \ex Your surname is the\_same \p{as} mine.
  \end{xlist}
  \ex Harry had never met anyone quite \p{like} Luna.
  \ex It was \choices{\p{as\_if}\\\p{like}} he had insulted my mother.
\end{exe}

The comparison is often made with respect to some dimension or attribute, the \psst{Characteristic}, 
which may or may not be scalar. 
The comparison may be figurative, employing simile, hyperbole, or spatial metaphor 
(\pex{close to} in the sense of `similar to'). 
The \psst{ComparisonRef} may even be a desirable or hypothetical/irrealis 
event or state (\pex{It was \p{as} it should have been}).

Prototypical prepositions include \p{than}, \p{as} (including the second item 
in the \p{as}--\p{as} construction), \p{like}, \p{unlike}. 
Prominent construals are \p{to} (\psst{Goal} for similar-thing) 
and \p{from} (\psst{Source} for dissimilar-thing).

\paragraph{\psst{Locus} construal for relative locative position on scale.}
Prototypically-locative prepositions that are \emph{relative} (\p{above}, \p{below}, \p{between}, \p{under}, etc.---%
in contrast to the \emph{absolute} ones like \p{at}, \p{in}, and \p{on})---%
invite a comparison between two things.
Where the relation between governor and object exists mainly to compare two items
(or their values) on an abstract scale, and the preposition metaphorically expresses this
relation as a relative location, \rf{ComparisonRef}{Locus} applies.
Examples include:
\begin{exe}
  \ex\rf{ComparisonRef}{Locus}:\begin{xlist}
    \ex\textit{Scale of measurement:} Your heart rate is \p{above} \choices{100~bpm\\normal\\mine}.
    \ex\textit{Reference point for cost:} The price is \p{within} my budget.
    \ex\textit{Scale of progress:} My team is \p{ahead\_of} your team in the tournament.
    \ex\label{ex:preferOver} \textit{Relative preference:} I prefer this restaurant \p{over} that one.\footnote{This is closely related to the notion of an alternative as in \cref{ex:alternativeOver}.}\\{}
    [paraphrase: I like this restaurant better \p{than} that one.]
  \end{xlist}
\end{exe}
This excludes absolute prepositions, as in \cref{ex:absoluteScale},
as well as prepositional phrases conveying circumstantial information about a scene
(e.g.~place, time, manner) or cost \cref{ex:nonCompare}.
\begin{exe}
  \ex\label{ex:absoluteScale}\begin{xlist}
    \ex Your heart rate is \p{at} 70~bpm. (\psst{Locus})
    \ex That's \p{in} my price range. (\psst{Locus})
    \ex The book is priced \p{at} \$10. (\rf{Cost}{Locus})
  \end{xlist}
  \ex\label{ex:nonCompare}\begin{xlist}
    \ex restaurants \p{within} 10 miles [physical location] (\psst{Locus})
    \ex I will explain my argument \p{below}. [discourse location] (\psst{Locus})
    \ex The guests will arrive \p{after} 6:00. (\psst{Time})
    \ex The book is priced \p{below} \$10. (\rf{Cost}{Locus})
  \end{xlist}
\end{exe}
See also \psst{Approximator}.

\paragraph{\psst{Source} and \psst{Goal} construals.}
Relations of congruence (resemblance, equivalence, proportionality, etc.)\ may be expressed with \p{to}, 
while difference may be expressed with \p{from}:
\begin{exe}
  \ex\rf{ComparisonRef}{Goal}:\begin{xlist}
    \ex Shall I compare thee \p{to} a summer's day?
    \ex Her height is \choices{equal\\close} \p{to} mine.
    \ex The cost should be proportional \p{to} the number of attendees. 
  \end{xlist}
  \ex\rf{ComparisonRef}{Source}:\begin{xlist}
    \ex We need to distinguish what is achievable \p{from} what is desirable.
    \ex Her height is different \p{from} mine.\footnote{American English. Interestingly, 
    \emph{different \p{to}} occurs in British English.}
  \end{xlist}
\end{exe}

\paragraph{\psst{Ancillary} construal.}
\begin{exe}
  \ex Don't compare me \p{with} my sister! (\rf{ComparisonRef}{Ancillary})
\end{exe}

\paragraph{Category as standard.} 
An indirect comparison can be made by relating something to a category 
to which it may or may not belong. 
The category stands for its members or prototypes. For example, in:
\begin{exe}
  \ex\label{ex:catAsStandard} He is short \p{for} a basketball player. (\psst{ComparisonRef})
\end{exe}
the category \pex{basketball player} serves as the standard against which \pex{he} is deemed short.

\paragraph{Sufficiency and excess.}
Sufficiency and excess can be expressed with
adverbs (\pex{too}, \pex{enough}, \pex{insufficiently}, etc.)\ and
adjectives (\pex{insufficient}) that license a PP or infinitival expressing the consequence.\footnote{See the Degree-Consequence construction \citep{bonial-18}.}
For example:
\begin{exe}
  \ex\label{ex:BasketballCmp}\rf{ComparisonRef}{Purpose}:\begin{xlist}
    \ex\label{ex:tooShort} He is \choices{too short\\not tall enough} \choices{\p{for}\\\p{to} play} basketball.
    \ex\label{ex:insufficient} His height is insufficient \p{for} basketball.
  \end{xlist}
\end{exe}
Playing basketball is the desired outcome, but it is conditional
on some scalar property relative to an implicit point on the scale---in \cref{ex:BasketballCmp},
a minimum height associated with playing basketball.
As a consequence, the desired outcome may or may not be blocked.
Thus, the consequence phrase helps to establish a reference point of comparison.

As discussed under \psst{Purpose}, if the consequence phrase in such a construction
meets the criteria for purposes, it is labeled \rf{ComparisonRef}{Purpose}.
Otherwise, the non-purpose consequence is labeled \rf{ComparisonRef}{Goal}.

\paragraph{\rf{Manner}{ComparisonRef} construal.}
This applies to an analogy describing the \emph{how} of an event
(be it agentive or perceptual):
\begin{exe}
\ex \rf{Manner}{ComparisonRef}:\begin{xlist}
  \ex You eat \p{like} a pig (eats).
  \ex You smell \p{like} a pig.
\end{xlist}
\end{exe}
However, where an analogy is an external comment on an event 
rather than filling in a role of the event, it is simply \psst{ComparisonRef}. 
Contrast:
\begin{exe}
  \ex You ate a whole pie \p{like} my cousin did.
  \begin{xlist}
    \ex \emph{Role reading:} The way in which you ate a pie was similar. (\rf{Manner}{ComparisonRef})
    \ex \emph{External comment reading:} You ate a whole pie, and so did my cousin. (\psst{ComparisonRef})
  \end{xlist}
\end{exe}

\paragraph{Analogy and non-analogy readings of \p{like}.}
In descriptions, adverbial \p{like}, \p{as\_if}, etc.\  
can be ambiguous, especially in a scene of perception. 
For example:
\begin{exe}
  \ex This looks \p{like} a Van Gogh painting.
  \begin{xlist}
    \ex \emph{Analogy reading:} This looks similar to a Van Gogh painting. (\rf{Manner}{ComparisonRef})
    \ex \emph{Conclusion reading:} This looks to be a Van Gogh painting (it probably is one). (\rf{Theme}{ComparisonRef})
  \end{xlist}
  \ex It sounded \choices{\p{like}/\p{as\_if}}
  \begin{xlist}
    \ex \dots he had drunk a gallon of helium. (\rf{Manner}{ComparisonRef}: analogy reading more likely)
    \ex \dots they weren't taking me seriously. (\rf{Theme}{ComparisonRef}: conclusion reading more likely)
  \end{xlist}
\end{exe}
Similarly for \pex{seem \p{like}}, \pex{feel \p{like}}, etc.

Another ambiguity can arise when \p{like} occurs with \pex{what} as its extracted object. 
In the following sentences, the most likely interpretation is not one of analogy between 
two things, but rather an open-ended description. 
(\pex{Who does it look \p{like}?}, by contrast, implicates an analogy to an individual.) 
We therefore treat \pex{\p{like} what} as a PP idiom, 
and label it \rf{Manner}{ComparisonRef}:
\begin{exe}
  \ex\label{ex:whatlike}\rf{Manner}{ComparisonRef}:\begin{xlist}
    \ex I know what\_~~Steve looks~~\_\p{like}. (I know how Steve looks.)
    \ex What\_~~does her hair look~~\_\p{like}? (How does her hair look?)
    \ex What\_~~is the party~~\_\p{like}? (How is the party?)
  \end{xlist}
\end{exe}
A \pex{how}-paraphrase is generally possible, though \pex{how} may suggest 
a positive or negative evaluation is available, whereas \pex{what} is more neutral.

Constrast unaccusative perception verb + \p{of} combinations:
\begin{exe}
  \ex\label{ex:smellOfNotCmp} \choices{Your father smells\\The soup tastes} \p{of} elderberries. (\rf{Manner}{Stuff}) [also~\cref{ex:smellOf}]
\end{exe}

\paragraph{Category exemplars and set members.} 
When governed by an NP naming a category or set, \p{like} is ambiguous 
between exemplifying a member, as in \cref{ex:likeSetMember} and \cref{ex:likeCatMember}, 
and merely indicating similarity, as in \cref{ex:likeSetSimilar} and \cref{ex:likeCatSimilar}:
\begin{exe}
  \ex Colbert frequently promotes comedians \p{like} himself.
    \begin{xlist}
      \ex\label{ex:likeSetSimilar} [\emph{Exclusive/restrictive reading:} \emph{similar to} himself (but not including himself)]
        (\psst{ComparisonRef})
      \ex\label{ex:likeSetMember} [\emph{Inclusive/nonrestrictive reading:} \emph{such as}/\emph{including} himself (he promotes himself, among others)]
        (\rf{PartPortion}{ComparisonRef})
    \end{xlist}
  \ex 
    \begin{xlist}
      \ex\label{ex:likeCatSimilar} I don't know anyone else \p{like} her. [anyone else \emph{similar to} her]\\ (\psst{ComparisonRef})
      \ex\label{ex:likeCatMember} It must be great to have a wonderful doctor \p{like} \choices{her\\she is}.\\ {}
      [It must be great to have her because she is a wonderful doctor]\\ (\rf{Identity}{ComparisonRef})
    \end{xlist}
\end{exe}

\paragraph{Instead-of alternatives.}
\psst{ComparisonRef} also applies to a default or already established thing
for which something else stands in or is chosen as an alternative.
\begin{exe}
  \ex I ordered soup \choices{\p{instead\_of}\\\p{rather\_than}} salad.
  \ex \p*{Instead\_of}{instead\_of} ordering salad, I ordered soup.
  \ex The new shirts were gray \p{instead\_of} black.
\end{exe}
May be construed spatially:
\begin{exe}
  \ex\label{ex:alternativeOver} I chose soup \p{over} salad. (\rf{ComparisonRef}{Locus})
\end{exe}
This is similar to the static-preference use of \p{over} illustrated in \cref{ex:preferOver}.
See also \psst{Ancillary} and \psst{Theme}.

\begin{history}
  A separate category \sst{InsteadOf} was introduced in v2.0 for alternatives,
  but in v2.5 it was merged with \psst{ComparisonRef} after the distinction
  became fraught for some uses of \p{instead\_of} and \p{rather\_than}.
\end{history}

\hierBdef{SetIteration}

\shortdef{Set-wise relation such as the unit of measure in a rate expression.}

This applies to rates using \p{per} or \p{by} to specify a unit, or atemporal \p{for} + \w{each}/\w{every} indicating a regular correspondence:

\begin{exe} \ex \begin{xlist}
  \ex The cost is \$10 \p{per} item.
  \ex \p*{For}{for} every 10 you buy, you get 1 free.
  \ex A fuel efficiency of 40 miles \p{per} gallon (of gas)
  \ex Pizza is sold \p{by} the slice.
  \ex They charge \p{by} the hour.
\end{xlist}\end{exe}

It also applies to expressions that denote a generalization over a set:
\begin{exe} \ex \begin{xlist}
  \ex Pets \lex[in\_general]{\p{in}\_general} tend to trigger my allergies.
  \ex Supposedly, \lex[on\_average]{\p{on}\_average} we consume a gallon of milk every day.
\end{xlist}\end{exe}

Contrast \psst{Frequency}, which describes \emph{how often} an \emph{event} occurs.

\paragraph{NPN Construction.} \psst{SetIteration} applies to the preposition linking two nominals (often the same noun) when the combination carries a meaning of iteration or regular correspondence. This applies whether the nouns are temporal or atemporal:

\begin{exe} \ex\label{ex:npn} \begin{xlist}
  \ex day \p{to} day tasks
  \ex day \p{by} day
  \ex day \p{after} day
  \ex We interviewed candidate \p{after} disappointing candidate.
  \ex We'll match your contribution dollar \p{for} dollar.
  \ex They went house \p{to} house in search of the fugitive.
\end{xlist}\end{exe}

\Cref{ex:npn} exemplifies what is known as the \emph{NPN Construction}: it is syntactically idiosyncratic, as the preposition and second noun cannot necessarily be omitted (\pex{*We'll match your contribution dollar.})\ \citep{jackendoff-08}.\footnote{Only a few prepositions occur productively in this construction. In the interest of simplicity, we refrain from attempting to assign a separate function (such as \psst{Goal} for \p{to} or \psst{Time} for \p{after}) that may explain why certain prepositions participate in the construction with a certain meaning.}

Not all instances of the construction have an iterative meaning, however: 
\begin{exe} \ex\label{ex:npn2} \begin{xlist}
  \ex morning \p{to} night talks (\rf{Duration}{EndTime})
  \ex wall \p{to} wall carpeting (\rf{Extent}{Goal})
\end{xlist}\end{exe}

Note that in \cref{ex:npn,ex:npn2}, the scene role technically represents the semantic output of the whole construction. In \cref{ex:npn2}, the function is different from the scene role, focusing on the preposition as a marker of the second nominal.


\begin{history}
  In v1, this fell under \sst{Value}. 
  In v2.0, it was called \sst{RateUnit}.
  In v2.6, it was broadened and renamed from \sst{RateUnit} to \sst{SetIteration}.
\end{history}

\hierBdef{SocialRel}

\shortdef{Party (individual, group of persons, or institution) 
with which another party has a stable affiliation.}

Typically, \psst{SocialRel} applies directly to interpersonal relations 
(versus \psst{Org} and \psst{OrgMember} for relations involving an organization).
It does not have any prototypical adpositions. 
Construals include:
\begin{exe}
  \ex \begin{xlist}
      \ex\label{ex:workwithSR} I work \p{with} Michael. (\rf{SocialRel}{Ancillary})
      \ex Joan has a class \p{with} Miss Zarves. (\rf{SocialRel}{Ancillary})
    \end{xlist}
  \ex people \p{with} children (\rf{SocialRel}{Characteristic})
  \ex\rf{SocialRel}{Gestalt} 
  {\setlength\multicolsep{0pt}%
  \begin{multicols}{2}
    \begin{xlist}
      \ex Joan is the \choices{sister\\wife} \p{of} John.
      \ex Joan is a student \p{of} Miss Zarves.
      \ex the rivalry \p{of} the teams

      \sn Joan is John\p{'s} \choices{sister\\wife}.
      \sn Joan is Miss Zarves\p{'s} student.
      \sn the team\p{s'} rivalry
    \end{xlist}
  \end{multicols}}
  \ex the rivalry \p{between} the teams (\rf{SocialRel}{Whole}) [see \cref{ex:betweenParties}]
  \ex Joan is studying \p{under} Prof.~Smith. (\rf{SocialRel}{Locus})
  \ex Joan is married \p{to} John. (\rf{SocialRel}{Goal})
  \ex Joan is divorced \p{from} John. (\rf{SocialRel}{Source})
  \ex Joan bought her house \p{through} a real estate agent. [intermediary] (\rf{SocialRel}{Instrument})
\end{exe}

Note, however, that \emph{work \p{with}} is ambiguous between 
being in an established professional relationship \cref{ex:workwithSR}, 
and engaging temporarily in a joint productive activity:
\begin{exe}
  \ex\label{ex:workwithCA} I was working \p{with} Michael after lunch. (\rf{Agent}{Ancillary})
\end{exe}
It is up to annotators to decide from context which interpretation 
better fits the context.

\begin{history}
  Renamed from v1 label \sst{ProfessionalAspect}, which was borrowed from 
  \citet{srikumar-13,srikumar-13-inventory}.
  The name \psst{SocialRel} reflects
  a broader set of stative relations involving an individual 
  in a social context, including kinship and friendship.
  See also note under \psst{Org}.
\end{history}

\section{Constraints on Role and Function Combinations}\label{sec:constraints}

The present scheme emerged out of extensive descriptive work with corpus data. 
Given the abundance of rare preposition usages, this document does not claim 
to cover every possible role\slash function combination for English, 
let alone other languages. 
Below are the few categorical restrictions that seem warranted for English.

\subsection{Supersenses that are purely abstract}\label{sec:abstractlabels}

\psst{Participant} and \psst{Configuration} are intended only to 
organize subtrees of the hierarchy, and not to be used directly. 

\subsection{Supersenses that are never used in English}

\psst{Content} is not expected to apply to English prepositions or possessives as either role or function. It is included in the hierarchy for use by other languages.

\subsection{Supersenses that cannot serve as functions}

\textbf{For English prepositions and possessives, \psst{Experiencer}, \psst{Stimulus}, \psst{Originator}, \psst{Recipient}, \psst{SocialRel}, \psst{Org}, \psst{OrgMember}, \psst{Ensemble}, and \psst{QuantityValue}
can only serve as scene roles, not functions.} 
Though scenes of perception, transfer, and interpersonal/organizational relationships 
are fundamental in language, they always seem to exploit construals from other domains 
(motion, causation, possession, and so forth).
(They may be marked more canonically by other English constructions, or by adpositional and case constructions in other languages.)

For example, \cref{ex:RecGoal} is clearly \psst{Recipient} at the scene level---Sam 
acquires possession of the box---but also 
fits the criteria for \psst{Goal} because Sam is an endpoint of motion 
(and \p{to} frequently marks \psst{Goal}s that are not \psst{Recipient}s). 
\Cref{ex:RecAgent} and \cref{ex:RecPoss} reflect \rf{Recipient}{Agent} and 
\rf{Recipient}{Gestalt} construals, respectively. 
\begin{xexe}
  \ex\label{ex:RecGoal} Give the box \p{to} Sam. (\rf{Recipient}{Goal})
  \ex\label{ex:RecAgent} the box received \p{by} Sam (\rf{Recipient}{Agent})
  \ex\label{ex:RecPoss} Sam\p{'s} receipt of the box (\rf{Recipient}{Gestalt})
\end{xexe}
Though the \psst{Goal} construal is arguably the most canonical expression of \psst{Recipient},
there is no preposition with a primary meaning of \psst{Recipient} independent of one of these other domains.

\textbf{Additional constraints on functions arise in the context of specific 
constructions (\cref{sec:cxns}).} For instance,
\begin{itemize}
  \item the s-genitive requires either \psst{Possessor} or \psst{Gestalt} as its function (\cref{sec:genitives})
  \item passive \p{by} requires \psst{Agent}, \psst{Force}, or \psst{Causer} as its function (\cref{sec:passives})
\end{itemize}

\subsection{Supersenses that cannot serve as roles}

In the present scheme, there are no supersenses that are restricted to serving as functions.

\subsection{No temporal-locational construals}\label{sec:temploc}

\textbf{Temporal prepositions never occur with a function of \psst{Locus}, \psst{Path}, or \psst{Extent}.}

Languages routinely borrow from spatial language to describe time, 
and spatial cognition may underlie temporal cognition \citep[e.g.,][]{lakoff-80,nunez-06,casasanto-08}.
A liberal use of construal would treat \pex{arriving \p{in} the afternoon} as \rf{Time}{Locus}, 
\pex{sleeping \p{through} the night} as \rf{Duration}{Path}, 
\pex{running \p{for} 20~minutes} as \rf{Duration}{Extent}, and so forth.
However, for simplicity and practicality, we elect not to annotate \psst{Locus}, \psst{Path}, or \psst{Extent} 
construals on ordinary temporal adpositions. Thus:
\begin{xexe}
  \ex arriving \p{in} the afternoon (\psst{Time})
  \ex sleeping \p{through} the night (\psst{Duration})
  \ex running \p{for} 20~minutes (\psst{Duration})
\end{xexe}
\rf{Time}{Direction} is possible, however, as are other atemporal functions:
\begin{xexe}
  \ex Schedule the appointment \p{for} Monday. (\rf{Time}{Direction})
  \ex January \p{of} last year (\rf{Time}{Whole})
  \ex Will you attend Saturday\p{'s} class? (\rf{Time}{Gestalt})
  \ex It took a year\p{'s} work to finish the book. (\rf{Duration}{Gestalt})
\end{xexe}

Note that the above is qualified to `ordinary temporal adpositions'. 
\textbf{When the first argument of a comparative construction is marked with \p{as}, 
the function is always \psst{Extent}, even if the scene role is temporal.} 
See \cref{sec:as-as}.

\subsection{Construals where the function supersense is an ancestor or descendant of the role supersense}

Ordinarily, if a construal holds between two (distinct) supersenses, these are from different branches of the hierarchy.
In a few cases, however, one is the ancestor of the other.

\paragraph{Role is ancestor of function.}
\begin{itemize}
  \item Setting events or situations with a salient spatial metaphor are \rf{Circumstance}{Locus} or \rf{Circumstance}{Path}.
  \item Fictive motion (the extension of a normally dynamic preposition to a static spatial scene) 
  can warrant \rf{Locus}{Goal} or \rf{Locus}{Source}, as discussed under \psst{Locus}.
  \item Complete contents of containers are \rf{Characteristic}{Stuff}.
\end{itemize}

\paragraph{Function is ancestor of role.}
\begin{itemize}
  \item Some s-genitives are annotated as \rf{Whole}{Gestalt}: see \cref{sec:genitives}.
  \item When an organization is framed via a genitive construction in relation to its members, \rf{Org}{Gestalt} is used.
  \item For \w{amount}/\w{number}/etc.~+ \p{of} + ITEM, \rf{QuantityItem}{Gestalt} is used.
  \item When a locative PP is coerced to a goal, as with \emph{put}, \rf{Goal}{Locus} is used.
\end{itemize}

\section{Special Constructions}\label{sec:cxns}

This section discusses notable constructions that are not limited to a single supersense.

\subsection{Genitives/Possessives}\label{sec:genitives}

\Citet{blodgett-18} detail the application of this scheme to English possessive constructions:
the so-called \textbf{s-genitive}, as in \cref{ex:SGen}, and 
\textbf{of-genitive}, as in \cref{ex:OfGen}:
\begin{exe}
  \ex\label{ex:SGen} 
    \begin{xlist}
      \ex \choices{the Smith family\p{'s}\\\p{their}} house (\psst{Possessor})
      \ex \choices{the tea\p{'s}\\\p{its}} price (\psst{Gestalt})
    \end{xlist}
  \ex\label{ex:OfGen} 
    \begin{xlist}
      \ex the house \p{of} the Smith family (\psst{Possessor})
      \ex the price \p{of} the tea (\psst{Gestalt})
    \end{xlist}
\end{exe}
Note that the s-genitive is realized with case marking (clitic \p{'s} or possessive pronoun\footnote{For ease of indexing, 
\p{'s} or \p{s'} is preferred over possessive pronouns for s-genitive examples in this document.}) 
rather than a preposition, 
and the case-marked NP in the s-genitive alternates with the object of the preposition in the of-genitive.
(This may feel unintuitive: annotators looking at the s-genitive construction are often tempted to focus on 
the role occupied by the head noun rather than the case-marked noun.)

The s-genitive and of-genitive are particularly associated with 
\psst{Possessor} (which applies to a canonical form of possession) 
and the more general category \psst{Gestalt}; both supersenses are illustrated above \cref{ex:SGen,ex:OfGen}.
In addition, both genitive constructions can mark participant roles and other kinds of relations, 
including \psst{Whole} and \psst{SocialRel} relations. 
When the s-genitive is used, the \emph{function} is always either \psst{Gestalt} (most cases) 
or \psst{Possessor} (when the possession is sufficiently canonical).
While overlapping in scene roles with the s-genitive, 
\p{of} is considered compatible with some additional functions, 
including \psst{Whole}, \psst{Source}, and \psst{Theme}; thus of-genitives 
with such roles do not need to be construed as \psst{Gestalt} or \psst{Possessor}:
\begin{exe}
  \ex\rf{SocialRel}{Gestalt}:\begin{xlist}
    \ex the grandfather \p{of} Lord Voldemort
    \ex \choices{Lord Voldemort\p{'s}\\\p{his}} grandfather
  \end{xlist}
  \ex\begin{xlist}
    \ex the hood \p{of} the car (\psst{Whole})
    \ex the nose \p{of} He-Who-Must-Not-Be-Named (\psst{Whole})
    \ex \choices{the car\p{'s} hood\\\p{its}} (\rf{Whole}{Gestalt})
    \ex \choices{He-Who-Must-Not-Be-Named\p{'s} nose\\\p{his}} (\rf{Whole}{Gestalt})
  \end{xlist}
  \ex\begin{xlist}
    \ex the arrival \p{of} the queen (\psst{Theme})
    \ex \choices{the queen\p{'s} arrival\\\p{her}} (\rf{Theme}{Gestalt})
  \end{xlist}
  \ex \choices{Shakespeare\p{'s}\\\p{his}} works (\rf{Originator}{Gestalt})
  \ex These are children\p{'s} clothes.\footnote{Cannot readily be paraphrased with \p{their} because \w{children} is not referential, 
  but rather refers to a kind. This construction has been termed the \emph{descriptive genitive} \citep[pp.~322, 327--328]{quirk-85}.} [clothes intended for use and possession by children] (\rf{Beneficiary}{Possessor})
\end{exe}

The literature on the genitive alternation examines the factors that condition 
the choice of construction; important factors include the length and animacy of the possessed NP.
In addition, \p{of} participates in certain constructions that are not really possessives---%
e.g.~\pex{this sort \p{of} sweater} (\psst{Species}).

Some difficult cases are clarified below.

\paragraph{Person in relation to a place or travel.}
In relation to an act of travel, the person is treated as a (possibly non-volitional) 
participant in a motion event. 
Otherwise, a person in relation to an associated place is \psst{Gestalt}.
\begin{exe}
  \ex \p{my} \choices{destination\\journey\\travels} (\rf{Theme}{Gestalt})
  \ex \p{my} \choices{hometown\\birthplace} (\psst{Gestalt})
\end{exe}

\paragraph{Lexicalized expressions and idioms.}
If the genitive is lexicalized as part of a name or term, it is not separately annotated:
\begin{exe} \ex \begin{xlist}
  \ex a \w{master\_'s} degree (non-SNACS)
  \ex a \w{New\_Year\_'s} resolution; a resolution for \w{New\_Year\_'s} (non-SNACS)
  \end{xlist}
\end{exe}

Certain idioms require an s-genitive \textbf{argument} that does not participate in 
any transparent semantic relationship; for these, \backposs is used (\cref{sec:possidiom}).


\subsection{Passives}\label{sec:passives}

The construction for passive voice (in verbs and nominalizations thereof) 
involves an optional \p{by}-PP;
the object of \p{by} alternates with the subject in active voice. 
While a variety of scene roles can be expressed with this phrase, 
the \emph{functions} associated with passive \p{by} are limited to 
\psst{Agent} and \psst{Force} (for participants within the event) and \psst{Causer} (for a participant in a causing event):
\begin{xexe}
  \ex the decisive vote \p{by} the City Council (\psst{Agent})
  \ex the devastation wreaked \p{by} the fire (\psst{Force})
  \ex The horse was jumped over the fence \p{by} Claire. (\psst{Causer})
  \ex This story was told \p{by} my grandmother. (\rf{Originator}{Agent})
  \ex The news was not well received \p{by} the White House. (\rf{Recipient}{Agent})
  \ex Mr. Dursley is employed \p{by} Grunnings. (\rf{Org}{Agent})
  \ex The window was broken \p{by} the hammer. (\rf{Instrument}{Force})
  \ex scared \p{by} the bear (\rf{Stimulus}{Force})
\end{xexe}

\subsection{Comparatives and Superlatives}

Various constructions express a comparison between two arguments. 

\paragraph{\psst{ComparisonRef} for second argument.}
When the second argument (the point of reference) 
is adpositionally marked, \psst{ComparisonRef} is used, regardless of 
its complement's syntactic type: 
\begin{exe}
  \ex\label{ex:comparisonrefArg}\begin{xlist} 
        \ex Your face is as red \p{as} \choices{a rose\\mine is}. (\psst{ComparisonRef})
        \ex Your face is redder \p{than} \choices{a rose\\mine is}. (\psst{ComparisonRef})
    \end{xlist}
\end{exe}
See further examples at \psst{ComparisonRef}.

\subsubsection{\p*{As}{as}-\p{as} comparative construction}\label{sec:as-as}

\paragraph{\psst{Extent} argument.} 
In an \p{as}-\p{as} comparison, the scene role of the first argument 
(the object of the first \p{as}) is the role that would be operative 
if the construction were removed and only the first argument remained: 
e.g., \pex{I stayed as long as I could} $\rightarrow$ \pex{I stayed long}.
The function of the first \p{as} is always \psst{Extent} 
to reflect that it marks the degree on a scale:
\begin{exe}
  \ex\begin{xlist}
    \ex I helped \p{as} much as I could. (\psst{Extent})
    \ex Your face is \p{as} red as a rose. (\rf{Characteristic}{Extent})
    \ex I helped \p{as} carefully as I could. (\rf{Manner}{Extent})
    \ex I stayed \p{as} long as I could. (\rf{Duration}{Extent})
    \ex I helped \p{as} often as I could. (\rf{Frequency}{Extent})
    \ex I've eaten (twice) \p{as} much (food) as you. [amount of something]\\ 
    (\rf{Approximator}{Extent})
  \end{xlist}
\end{exe}

\paragraph{Second argument: \psst{ComparisonRef}.} 
See \cref{ex:comparisonrefArg} above.

\subsubsection{Superlatives}\label{sec:superlative}

\psst{Whole} is used for the superset or gestalt licensed by a superlative:
\begin{exe}
  \ex the youngest \p{of} the children (\psst{Whole})
\end{exe}
See more at \psst{Whole}.

\subsection{Infinitive Clauses}\label{sec:inf}

In its function as infinitive marker, \p{to} is not generally considered to be a preposition. 
Nevertheless, we consider all uses of \p{to} for adposition supersense annotation 
because infinitive clauses (infinitivals) can express similar semantic relations 
as prepositional phrases. 

\subsubsection{Infinitival varieties of \psst{Purpose}}

Most notably, infinitival purpose adjuncts alternate 
with \p{for}-PP purpose adjuncts:
\begin{exe}
  \ex\label{ex:infPurpose}\psst{Purpose}:\begin{xlist}
    \ex\begin{xlist}
      \ex Open the door \p{to} let in some air.
      \ex Open the door \p{for} some air.
    \end{xlist}
    \ex\begin{xlist}
      \ex I flew to headquarters \p{to} meet with the principals.
      \ex I flew to headquarters \p{for} a meeting with the principals.
    \end{xlist}
  \end{xlist}
\end{exe}
Thus, from a practical point of view, we might as well treat infinitival \p{to} 
as capable of marking a \psst{Purpose}.

The following list summarizes semantic analyses that we consider for infinitivals,
which are detailed under \psst{Purpose}:
\begin{itemize}
  \item \textbf{Purpose adjuncts}, whether are adverbial or adnominal.
  These are labeled \psst{Purpose}. Some can be paraphrased with \p{in\_order\_to}.

  \item In a \textbf{commercial scene}, a service to performed in exchange for
  payment; labeled \rf{Theme}{Purpose}.
  Repeated from the discussion under \psst{Theme}:
  \begin{exe}
    \ex\begin{xlist}
      \ex They asked \$500 \p{to} make the repairs. (\rf{Theme}{Purpose})
      \ex \$500 \p{to} make the repairs was excessive. (\rf{Theme}{Purpose})
    \end{xlist}
  \end{exe}

  \item \textbf{Result} infinitives, such as those in \cref{ex:result},
  are labeled \psst{Goal}.

  \item Constructions of \textbf{sufficiency and excess}---\pex{too short \p{to} ride},
  \pex{not tall enough \p{to} ride}, etc., where the assertion of sufficiency or excess
  licenses an infinitival---are labeled \rf{ComparisonRef}{Purpose} or \rf{ComparisonRef}{Goal}.
  See discussions at \psst{ComparisonRef} and \psst{Purpose}.
\end{itemize}
The non-semantic label \backi applies to all other uses of the infinitive.

\subsubsection{Infinitivals with \p{for}-subject}
In \cref{ex:infPurpose}, the infinitive clause has no local subject---rather, 
an argument of the matrix clause doubles as the subject of the infinitive clause 
(control). However, a separate subject can be introduced with \p{for}, 
in which case \p{for}+NP is treated as a dependent of the infinitive verb 
and labeled \backi:
\begin{exe}
  \ex\begin{xlist}
    \ex I opened the door [\p{for}$_{\text{\backi}}$ Steve \p{to}$_{\psst{Purpose}}$ take out the trash].
    \ex It cost \$500 [\p{for}$_{\text{\backi}}$ the mechanic \p{to}$_{\text{\rf{Theme}{Purpose}}}$ make the repairs].
  \end{xlist}
\end{exe}

\subsubsection{\p*{For\_to}{for\_to} infinitives}
These occur in some English dialects:
\href{https://ygdp.yale.edu/phenomena/for-to-infinitives}{\emph{for to} infinitives}

\subsubsection{Other infinitivals}
Examples of infinitival tokens that do not receive a semantic label appear in \fullref{sec:specialinf}.

%

\subsection{PP Idioms}

Many PPs exhibit some amount of lexicalization or idiomaticity.
This is especially true of PPs that tend to be used predicatively.
In general it is extremely difficult to establish tests to distinguish idiomatic PPs 
from fully productive combinations. 
However, the usual criteria apply for the supersense analysis.

For example, if the PP answers a \emph{Where?}\ question, 
it qualifies as \psst{Locus}; 
whereas qualitative states usually have \psst{Characteristic} as the scene role:
\begin{exe}
  \ex He is \p{out\_of} town. (\psst{Locus})
  \ex The company is \p{out\_of} business. (\rf{Characteristic}{Locus})
\end{exe}
See further discussion at \psst{Characteristic}.

\subsubsection{PP Idioms vs.~Multiword Prepositions}\label{sec:ppmwp}

A PP idiom is a fixed or semi-fixed expression consisting of an adposition plus its complement 
(usually an NP, AdjP, or AdvP), which must be a complete phrase. 
In some of these expressions the complement may take variable modifiers (e.g., \pex{\p{on}\_ ONE's \_own}: see \cref{sec:possidiom}). 
The PP idiom as a whole does not take a complement (is intransitive). 
A fixed expression ending in a transitive preposition like \p{of} or \p{as} 
(\p{in\_search\_of}, \p{as\_long\_as}) requires a complement, and thus is not a PP idiom.\footnote{Infinitive marker 
\p{to} counts as a transitive preposition for purposes of this definition.}

\subsubsection{Reflexive PP Idioms}\label{sec:refl}

Certain idiomatic constructions involve a preposition that requires a reflexive 
direct object.

\paragraph{PERFORM-ACTIVITY \emph{for} oneself.\lex*[for oneself]{\p{for} oneself}}
\begin{itemize}
  \item When something is done for one's own benefit rather than someone else's:
    \begin{exe}
      \ex I took a vacation \p{for} myself (\psst{Beneficiary})
    \end{exe}
  \item When something is done in a way that affords direct rather than second-hand information:
    \begin{exe}
      \ex You should try out the restaurant \p{for} yourself! (\rf{Agent}{Beneficiary})
    \end{exe}
\end{itemize}
\paragraph{PERFORM-ACTIVITY \emph{by} oneself.\lex*[by oneself]{\p{by} oneself}}
\begin{itemize}
  \item When something is done without accompaniment (the negation would be \emph{\p{with} others}):
    \begin{exe}
      \ex I had lunch (all) \p{by} myself [`alone'] (\psst{Ancillary}\footnote{Though \emph{myself}
      is not literally accompanying \emph{I}, the PP as a whole describes the nature of accompaniment (or lack thereof).}) 
    \end{exe}
  \item When something is accomplished independently, without assistance:
    \begin{exe}
      \ex I made the decision (all) \p{by} myself. (\psst{Manner})
      \ex The computer rebooted all \p{by} itself. (\psst{Manner})
    \end{exe}
\end{itemize}
\paragraph{BE \emph{by} oneself.\lex*[by oneself]{\p{by} oneself}}
Alone; unaccompanied:
\begin{exe}
  \ex I am \p{by} myself right now. (\psst{Ancillary})
\end{exe}

\subsection{Ages}\label{sec:age}

An individual's age is a temporal property, licensing both \psst{Time} and \psst{Characteristic} prepositions:
\begin{exe}
  \ex\begin{xlist} 
    \ex a child \p{of} (age) 5 (years) (\psst{Characteristic})
    \ex Martha was already reading \choices{\p{at}/\p{by}/\p{before}} (the age of$_{\text{\psst{Identity}}}$) 5 (years). (\psst{Time})
  \end{xlist}
\end{exe}

\subsection{Fixed expressions considered non-adpositional}\label{sec:fixed}

\begin{itemize}
\item \textbf{Named entities}, including multiword names (e.g., \emph{Out of Africa},
\emph{The Taming of the Shrew}), are treated as single lexemes
and should receive a nominal\slash entity type, not a SNACS supersense.

\item \textbf{Grammaticalized multiword expressions} like the hedges in \cref{ex:hedges} and the
semi-auxiliaries in \cref{sec:specialinf} should not receive a SNACS supersense.

\begin{exe}
  \ex\label{ex:hedges} I'm \choices{\lex[kind\_of]{\w{kind\_of}}\\\lex[sort\_of]{\w{sort\_of}}} hungry. (non-SNACS)
\end{exe}

\item \textbf{Verb-particle combinations} where the particle is not adding a compositional
spatial meaning are treated as verbal multiword expressions
and do not receive a SNACS supersense.

\begin{exe}
  \ex Compositional spatial meanings:\begin{xlist}
    \ex The leaves blew \p{up}. (= blew into the air) (\psst{Direction})
    \ex The fan blew the leaves \p{up}. (\psst{Direction})
  \end{xlist}
  \ex Idiomatic/non-spatial meanings:\begin{xlist}
    \ex\begin{xlist}
      \ex I \w{blew\_up} the balloon. (= inflated) (non-SNACS)
      \ex I \w{blew\_} the balloon \w{\_up}. (= inflated) (non-SNACS)
    \end{xlist}
    \ex The bomb \w{blew\_up}. (= literally exploded) (non-SNACS)
    \ex My friend \w{blew\_up} at$_{\text{\rf{Beneficiary}{Direction}}}$ me. (= exploded in anger) (non-SNACS)
  \end{xlist}
\end{exe}

\item Where a verb or other content word absolutely requires a transitive preposition
to receive the correct meaning, as in \cref{ex:come-across},\footnote{These can be called \textbf{integral prepositions}\index{integral preposition}. Verbs with integral prepositions consitute a subtype of \textbf{prepositional verbs}\index{prepositional verb}, i.e.~verbs that select for a particular preposition.
Both \cref{ex:wait-for} and \cref{ex:come-across} can be considered prepositional verbs.} it is treated as a content multiword expression
and does not receive a SNACS supersense.

\begin{exe}
  \ex\label{ex:wait-for} I decided to wait \p{for} someone. (How long did you wait?) (\psst{Theme})
  \ex\label{ex:come-across} At the library I \w{came\_across} an interesting book. (\#When did you come?) (non-SNACS)
\end{exe}
\end{itemize}

\section{Special Labels}\label{sec:special}

For annotating data, there needs to be a way to indicate that \emph{none} 
of the adposition supersenses apply to a particular token. 

\subsection{DISCOURSE (\backd)}\label{sec:discourse}

Discourse connectives and other markers that transition between ideas 
or convey speaker attitude/hedging/emphasis/attribution but do not belong 
to propositional content. Examples include:

\begin{exe}
\ex \p{according\_to}; \lex[after\_all]{\p{after}\_all}, \lex[of\_course]{\p{of}\_course}, \lex[by\_the\_way]{\p{by}\_the\_way}; 
\lex[for\_chrissake]{\p{for}\_chrissake} (interjection); 
\lex[above\_all]{\p{above}\_all}, \lex[to\_boot]{\p{to}\_boot}; 
\lex[in\_other\_words]{\p{in}\_other\_words}, \lex[on\_the\_other\_hand]{\p{on}\_the\_other\_hand}; 
\lex[in\_my\_experience]{\p{in} my experience}, \lex[in\_my\_opinion]{\p{in}\_my\_opinion}
\end{exe}

This label also covers ``additive focusing markers'' 
\citep[p.~592]{cgel} with a meaning similar to `also' or `too',
where an item is added to something already established in the discourse:
\begin{exe}
  \ex\begin{xlist}
    \ex I shot the sheriff \lex[as\_well]{\p{as}\_well}.
    \ex They serve coffee, and tea \lex[as\_well]{\p{as}\_well}.
  \end{xlist}
\end{exe}
It also covers topicalization markers:
\begin{exe}
  \ex \p*{As\_for}{as\_for} the sheriff, well, I shot 'im.
\end{exe}
Finally, \backd applies to adpositions relating a metalinguistic mention of 
a speech act to the speech content itself---whether the adposition 
introduces this speech act mention, as in \cref{ex:toSumItUp},
or links the discourse expression to a subordinate statement, as in \cref{ex:sumItUpWith}.
\begin{exe}
  \ex\begin{xlist}
    \ex\label{ex:toSumItUp} \p*{To}{to} sum it up: It was a terrible experience.
    \ex\label{ex:sumItUpWith} I will sum it up \p{with}: It was a terrible experience.
  \end{xlist}
\end{exe}

\subsection{COORDINATOR (\backc)}\label{sec:coord}

Coordinating conjunctions and similar expressions where 
the two elements in the relation are semantically on an equal footing, 
rather than in a figure/ground relationship: 
\begin{exe}
  \ex They serve coffee \p{as\_well\_as} tea. [`They serve coffee and also tea']
  \ex Make sure to separate the dark \p{versus} light clothes. 
\end{exe}

\subsection{OTHER INFINITIVE (\backi)}\label{sec:specialinf}

As described in \cref{sec:inf}, infinitive clauses are analyzed with a supersense 
if and only if they serve as a purpose adjunct, or in certain purpose-related constructions 
(result; complement of entity-referring indefinite pronoun;
commercial service;
that which something is good or bad for, or sufficient or excessive for).
The special label \backi is reserved for all other uses of infinitival \p{to}, 
as well as \p{for} whenever it introduces the subject of an infinitive clause.\footnote{Essentially, 
our position is that these uses of infinitivals are more like syntactically core elements 
(subject, object) than obliques, and thus should be excluded from semantic annotation 
under the present scheme.}

Infinitivals warranting \backi include:
\begin{exe}\ex\begin{xlist}
  \ex I want \p{to} meet you. [complement of control verb]
  \ex I would\_like \p{to} try the fish. [\pex{would\_like} is a polite alternative to \pex{want}]
  \ex It seems \p{to} be broken. [complement of raising verb]
  \ex You have an opportunity \p{to} succeed. [complement of noun]
  \ex I'm ready \p{to} leave. [complement of adjective]
  \ex I'm glad \p{to} hear you're engaged! [complement of emotion adjective]
  \ex These new keys are expensive \p{to} copy. [tough-movement]
  \ex My plan is \p{to} eat at noon. [infinitival as NP]
  \ex It's impossible \p{to} get an appointment. [infinitival as NP, with cleft]
  \ex I know how \p{to} lead. [complement of wh-word]
  \ex I have something \p{to} do. [complement of indefinite pronoun that doesn't refer to an entity]
\end{xlist}\end{exe}

Multiword auxiliaries---such as quasi-modals \lex[have\_to]{\pex{have\_to}} `must', \lex[ought\_to]{\pex{ought\_to}} `should', etc., 
as well as \lex[have\_yet\_to]{\pex{have\_yet\_to}}---subsume the infinitival \p{to}, so no label on \p{to} is required:
\begin{exe}
  \ex You have\_to choose a date.
\end{exe}

Whenever \p{for} introduces a subject of an infinitival clause, the \p{for} token is labeled 
\backi (regardless of whether \p{to} receives a semantic label; see \cref{sec:inf}):
\begin{exe}\ex\begin{xlist}
  \ex I need [\p{for}$_{\text{\backi}}$ you \p{to}$_{\text{\backi}}$ help me].
  \ex I opened the door [\p{for}$_{\text{\backi}}$ Steve \p{to}$_{\psst{Purpose}}$ take out the trash].
\end{xlist}\end{exe}

\subsection{OPAQUE POSSESSIVE SLOT IN IDIOM (\backposs)}\label{sec:possidiom}

Semantic supersenses are used where possible for genitive\slash possessive 
constructions, as discussed in \cref{sec:genitives}. 
However, there are a few idioms which require a possessive pronoun 
that does not participate transparently in any semantic relation; 
these are designated with the special label \backposs:
\begin{exe}\ex\begin{xlist}
  \ex I am eating \lex[on\_my\_own]{on\_~~\p{my}~~\_own} today.
  \ex She tried \p{her} best.
  \ex He's not \p{your} average baseball player.
  \ex Billy knows \p{his} ABCs!
\end{xlist}\end{exe}
It is also used for the possessive in the \emph{way} construction: 
contrast
\begin{exe}
  \ex I like \p{her} way of eating---it is very polite. (\psst{Gestalt})
  \ex She will arrive soon: she is on\_~~\p{her}~~\_way. (\backposs) [see \cref{ex:on-way}]
  \ex I don't want to drive there because it is out\_of\_~~\p{my}~~\_way. (\backposs)
\end{exe}

Note that in the above examples, the possessive referent is variable, 
as reflected in the form of the pronoun.
For fully lexicalized phrases with genitive marking, see \cref{sec:genitives}.

\section{Changelog}

\subsection{Changes from earlier versions of this document}

\begin{itemize}
  \item \emph{Version 2.6 (June 18, 2022):} 
  \begin{itemize}
    \item \textbf{Rename \sst{Causer}$\rightarrow$\psst{Force} and \sst{RateUnit}$\rightarrow$\psst{SetIteration}.} The latter is expanded in scope.
    \item \textbf{2 new categories, \psst{Causer} and \psst{Content}, necessary for other languages.}
    \item Miscellaneous new examples and clarifications (\href{https://github.com/carmls/snacs-guidelines/milestone/4?closed=1}{Milestone 2.6}).
  \end{itemize}
  \item \emph{Version 2.5 (April 1, 2020):}
  Several structural changes to the hierarchy for better internal consistency
  and clarity:
  \begin{itemize}
    \item \textbf{\sst{Co-Agent} has been merged with \psst{Agent}, and \sst{Co-Theme} with \psst{Theme}.}
    \item \textbf{\sst{Accompanier} has been split into \psst{Ancillary} and \psst{Ensemble}.}
    \item \textbf{\sst{InsteadOf} has been merged with \psst{ComparisonRef}} (GitHub issue \ghi{5}).
    \item \textbf{\sst{OrgRole}, which was under \sst{SocialRel}, has been split into \psst{Org} (under \psst{Gestalt}) and \psst{OrgMember} (under \psst{Characteristic}) in order to reflect the asymmetry of the relation.}
    \item \textbf{\sst{Quantity} has been renamed \psst{QuantityItem}, and its inverse \psst{QuantityValue} has been added.} \psst{Approximator} has been moved under \psst{QuantityValue}.
    \item Revised the definition of \psst{Approximator}, and established the use of \rf{ComparisonRef}{Locus} for some uses of relative locative prepositions (see \psst{ComparisonRef}: ``\psst{Locus} construal for relative locative position on scale'') (\ghi{59}, \ghi{16}, \ghi{20}).
    \item Added \fullref{sec:construal} (\ghi{64}).
    \item An example labeled \rf{Theme}{Purpose}, \cref{ex:readyFOR},
    has been relabeled to plain \psst{Theme} for consistency with current guidelines for \psst{Purpose};
    and \w{responsible} has been added as one of the governors in the example (\ghi{55}).
    \item Added example of unit of measure in relation to attribute (\ghi{58}).
  \end{itemize}
  \item \emph{Version 2.4 (January 2, 2020):}
  \begin{itemize}
    \item \textbf{Overhauled definition and criteria for \psst{Purpose} versus other infinitivals.}
    Updated \psst{ComparisonRef}, \fullref{sec:inf}, \fullref{sec:specialinf} accordingly.
    \item \textbf{New section: \fullref{sec:fixed}}
    \item Several more \textbf{possessive} clarifications under \psst{Gestalt}, \psst{Possessor}, \fullref{sec:genitives}, \fullref{sec:possidiom} (GitHub issues \ghi{32}, \ghi{33}, \ghi{34}, \ghi{35}, \ghi{37}, \ghi{38})
    \item Several more verbs that select for prepositions under \psst{Theme} (\ghi{4}, \ghi{15}, \ghi{22}, \ghi{24}, \ghi{28}, \ghi{42}) and \psst{Species} (\ghi{39})
    \item Added an example of an informational \psst{Locus} (\ghi{14}).
  \end{itemize}
  \item \emph{Version 2.3 (August 18, 2019):} 
  \begin{itemize}
    \item Added \fullref{sec:ppmwp}.
    \item \textbf{\psst{Possessor}, \psst{Possession}: Substantially revised  to clarify their scope.}
    \item Added some difficult possessive examples in \psst{Gestalt}, and elsewhere for \rf{Agent}{Gestalt}, \rf{Theme}{Gestalt}, \rf{SocialRel}{Gestalt}, and \rf{Experiencer}{Gestalt} construals
    \item \textbf{\psst{Possession}, \psst{Theme}: Added discussion of transfer, goods, and services. 
    Goods and other transferred items are now \psst{Possession}, while services remain as \psst{Theme}.}
    \item \psst{Cost}: Clarified the explanation of when money should be treated like any other possession.
    \item \textbf{Reclassified adverbial and predicative entity-descriptions (depictives, shapes, states) from \psst{Manner} to \psst{Characteristic}, and revised their definitions accordingly.}
    \item \textbf{Specified that \psst{Temporal} should be used directly for aspectual prepositions (previously it was an abstract category: \cref{sec:abstractlabels}).}
    \item Moved journey-type PPs from \psst{Manner} to \psst{Circumstance}.
    \item \psst{Means}: Revised the definition and clarified relationship to \psst{Manner}.
    \item \psst{SocialRel}: Reworded the definition and added a kinship example for \rf{SocialRel}{Characteristic}.
    \item Clarified that \psst{Experiencer} applies to bodily \emph{sensations} (not physical bodily changes), and added cognition examples.
    \item Changed the treatment of replacees from \sst{InsteadOf} to \sst{Co-Theme}.
    \item List PP idioms in the index.
    \item \psst{Path}: Clarified that motion events can be located as points (\psst{Locus}), and relocated \cref{ex:pathmanner} from \psst{Manner}.
    \item Rectified inconsistent treatment of a family member in relation to the family (now always \sst{OrgRole}).
    \item Organizational collective members: now \Rf{OrgRole}{PartPortion}, formerly \Rf{OrgRole}{Characteristic}.
    \item Added dotted version number to title and adjusted author list.
  \end{itemize}
  \item \emph{Version 2.2 (July 2, 2018):} 
  \begin{itemize}
    \item Policy changes reflected in STREUSLE~4.0: 
    \begin{itemize}
      \item Rewrote \fullref{sec:genitives} and updated corresponding examples 
      to reflect a clarified policy on possessive construals. 
      Moved wearer from \psst{Gestalt} to \psst{Possessor}
      and attire from \psst{Characteristic} to \psst{Possession}.
      \item Added \fullref{sec:passives} and updated corresponding examples.
    \end{itemize}
    \item Policy changes that are reflected in STREUSLE~4.1: 
    \begin{itemize}
      \item In \cref{sec:as-as}, changed the function of the first \p{as} in the \p{as}-\p{as} construction 
      to \psst{Extent} (was \psst{Identity}).
      \item Changed the function of \psst{Originator} possessives to \psst{Gestalt} (was \psst{Possessor}).
      \item Expanded documentation and removed inconsistencies around containers and collective nouns 
      (see \psst{Stuff}, \sst{Quantity}, \psst{Characteristic}, \sst{OrgRole}).
      \item Specified \rf{Manner}{ComparisonRef} for certain adverbial uses of \p{like}.
      \item Revised the definition of \psst{Recipient} to relax the requirement of animacy.
      \item Mentioned conditions as a subclass of \psst{Circumstance}.
      \item Renamed \sst{Part/Portion} to \psst{PartPortion} to avoid technical complications 
      of the slash.
    \end{itemize}
    \item Added \fullref{sec:constraints}.
    \item Added \fullref{sec:age}.
    \item A few additional examples and fixes.
    \item Added an index of construals by function.
    \item Changes from v1 had neglected to mention the removal of \sst{Affector}, \sst{Undergoer}, \sst{Place}, 
    \sst{Elements}, and \sst{Superset} (thanks to Ken Litkowski for pointing this out).
  \end{itemize}
  \item \emph{Version 2.1 (January 16, 2018):} 
  \begin{itemize}
    \item Broadened and clarified \sst{DeicticTime}, moved it up a level in the hierarchy, 
      and renamed it to \psst{Interval}. Clarified the distinction between \psst{Interval} and \psst{Duration}.
    \item Clarified \psst{Locus}, \psst{Source}, \psst{Goal}, \psst{Path}, and 
      \psst{Direction}, especially with regard to (i)~intransitive prepositions, 
      (ii)~distance measurements, and (iii)~inherent parts.
    \item Significantly expanded the scope of \psst{Manner} to cover states of entities and depictives.
    \item Clarified \p{like} as \psst{ComparisonRef} with regard to categories and sets, 
    and \psst{PartPortion} with regard to elements and exceptions.
    \item Clarified \p{with} in regard to \psst{Topic} and \psst{Stimulus}.
    \item Added discussion of the ambiguity of temporal \p{over}: \psst{Duration} versus \rf{Time}{Duration}.
    \item Extensively clarified \psst{Purpose} and \psst{Beneficiary}, 
    and their relationship to \psst{ComparisonRef}, \psst{Recipient}, \psst{Experiencer}, and \psst{Stimulus}.
    \item Clarified that goods and services are \psst{Theme}; expanded on \sst{Co-Theme} examples.
    \item \psst{Frequency} used for an iteration.
    \item Various selectional verbs and miscellaneous constructions.
    \item Added examples of \p{'s} possessive/genitive marking.
    \item Added section for special syntactic constructions (\cref{sec:cxns}).
    \item Added special labels (\cref{sec:special}).
    \item Added an index of adpositions and supersenses, and an index of construals.
    \item Revised the title, abstract, and introductory material.
  \end{itemize}
  \item \emph{Version 2.0 (April 7, 2017): See \cref{sec:v1v2}}
\end{itemize}

\subsection{Major changes from v1}\label{sec:v1v2}

Changes that affect only a single label are explained below the relevant 
v2 labels.

\begin{itemize}
  \item \textbf{Removed multiple inheritance.} 
  The v1 network was quite tangled. The structure is greatly simplified 
  by analyzing some tokens as \emph{construals} \citep[\cref{sec:construal};][]{hwang-17}.
  \item \textbf{Revised and expanded the \psst{Configuration} subhierarchy.}
  \item \textbf{Removed the locative concreteness distinction.}
  In v1, labels \sst{Location}, \sst{InitialLocation}, and \sst{Destination} 
  were reserved for concrete locations, and the respective supertypes 
  \psst{Locus}, \psst{Source}, and \psst{Goal} used to cover abstract locations.
  This distinction was found to be difficult and without apparent
  relevance to preposition system of English or the other languages considered. 
  The concrete labels were thus removed.
  \item \textbf{Removed the location/state/value distinction.}
  The v1 scheme attempted to make an elaborate distinction between 
  values, states, and other kinds of abstract locations. 
  However, the English preposition system does not seem particularly 
  sensitive to these distinctions. (We are not aware of any prepositions 
  that mark primarily values or primarily states; rather, productive 
  metaphors allow locative prepositions to be extended to cover these, 
  and there are cases where teasing apart abstract location vs.~state vs.~value 
  is difficult.) Therefore, \sst{State}, \sst{StartState}, \sst{EndState}, 
  \sst{Value}, and \sst{ValueComparison} were removed. 
  \item \textbf{Revised the treatment of comparison and related notions.} 
  Removed \sst{Comparison/Contrast}, \sst{Scalar/Rank}, \sst{ValueComparison}; 
  moved \psst{Approximator} under \sst{Quantity}.
  \item \textbf{Greatly simplified the \psst{Path} subhierarchy.} See \cref{sec:Path}.
  \item \textbf{Simplified the \psst{Temporal} subhierarchy.} See \cref{sec:Temporal}.
  \item \textbf{Removed} \sst{Activity} (mostly replaced with \psst{Circumstance} and \psst{Topic}), 
  \sst{Reciprocation} (mostly merged with \psst{Explanation}), and \sst{Material} (merged with \psst{Source}).
  \item \textbf{Removed abstract labels} \sst{Affector}, \sst{Undergoer}, and \sst{Place}.
  \item \textbf{Removed the theme/patient distinction.} \psst{Theme} now includes patients.
  \item \textbf{Removed the primary/secondary participant distinction.} Whereas v1 had \sst{Co-\{Agent,Patient,Theme\}}, v2.5 allows multiple \psst{Agent} arguments
  or multiple \psst{Theme} arguments of the same predicate.
\end{itemize}

\bibliographystyle{plainnat}
\bibliography{psst2.bib}


\printindex
\printindex[construals]
\printindex[revconstruals]

\end{document}